%% file: arxiv_main.tex
\documentclass[11pt]{article}

\usepackage[T1]{fontenc}
\usepackage[margin=1in]{geometry}
\usepackage{amsmath,amssymb}
\usepackage{booktabs,longtable}
\usepackage{graphicx}
\usepackage{microtype}
\usepackage[round]{natbib}
\usepackage[hidelinks]{hyperref}

\newcommand{\E}{\mathbb{E}}
\newcommand{\Prob}{\mathbb{P}}
\newcommand{\QK}{\mathcal{Q}_K}
\newcommand{\Law}{\operatorname{Law}}
\newcommand{\TATE}{\operatorname{TATE}}
\newcommand{\TCATE}{\operatorname{TCATE}}

\begin{document}
\raggedbottom

\title{Wasserstein Causal Forests for Distribution-Valued Outcomes}
\author{Hugo Gobato Souto\\ University of São Paulo\\ \texttt{hgsouto@usp.br}\\ Permanent contact: \texttt{hugogobatosouto@gmail.com}}
\date{}
\maketitle

\begin{abstract}
This paper proposes Wasserstein Causal Forests (WCF) for settings in which each unit's outcome is itself a probability distribution. This study also defines finite-grid transformed average and conditional average treatment effects, including a reference-distance contrast that asks whether treatment moves unit-level distributions toward a prespecified benchmark. Simulations cover null effects, location and shape changes, limited overlap, equal-mean but different laws, heterogeneous effects, multimodality, and structural zeros. WCF is most accurate on the conditional-law metric in most reported designs and sharply improves reference-effect estimation in the principal location-and-shape settings, but it is less accurate than the forest baselines for multimodal settings. WCF is applied to the famous Project STAR \citep{word1990state}, revealing that small classes alter more than the mean: they raise within-grade mathematics achievement by $0.161$ standard deviations on average (SE $0.028$); but the gain is not a uniform location shift, it is larger in the upper part of the classroom score distribution ($+0.179$ at the ninetieth percentile versus $+0.091$ at the tenth) and, in descriptive stratum estimates, largest in the schools serving the most economically disadvantaged students (highest free-lunch quartile, $+0.262$, versus $+0.092$ to $+0.164$ elsewhere), while overall dispersion is essentially unchanged.
\end{abstract}

\section{Introduction}\label{sec:intro}

Average treatment effects reduce an intervention to a difference in means, albeit many scientific questions concern changes in dispersion, tails, asymmetry, participation, or multimodality. Conditional distributional treatment effects address such questions through conditional mean embeddings, distributional forests, and related nonparametric tools \citep{Counterfactualmeanembeddings,CATEGeneralization,GRFDistributionalCausalEffects,martineztaboada2023efficient,DRF-paper,naf2026causaldrf}. A distinct but complementary setting is studied in this paper: the observed outcome for each unit is itself a distribution, represented by a quantile function on a common grid. Examples include classroom score distributions, village consumption distributions, and regional wage distributions.

Regressing one summary at a time requires fixing the scientific question before estimation and can miss effects that leave the selected summary unchanged. Two conditional laws can have the same mean while differing in spread, tails, shape, or the frequency of a degenerate component. Estimating the conditional law of the quantile vector supplies one coherent object from which several prespecified functionals can be evaluated. It also keeps separate two operations (and their respective causal treatment effect estimands) that need not commute: averaging a transformed distribution-valued outcome and transforming its average \citep{souto_diamantis_tcda_2026}. Nonetheless, it is worth remembering that the richer target does not remove the need to choose a grid, control representation error, or justify causal identification, but makes those choices explicit.

This research proposes Wasserstein Causal Forests (WCF). For each treatment arm, WCF represents the conditional law of a unit-level quantile vector by a small cloud of monotone particles. The particles are learned with an energy-score objective, shared treatment-arm tree partitions, and shrinkage of arm-specific updates. A second layer uses cross-fitted augmented inverse-propensity weighting (AIPW) to calibrate marginal and moderator-specific effects for declared scalar functionals. The law estimator and calibration layer have different roles: calibration changes reported functional contrasts, not the fitted particle law.

The paper contributes 1. a conditional-law learner adapted to the Wasserstein geometry of one-dimensional distributions, 2. a finite-grid specialization of the Topological Average Treament Effect (TATE) and Topological Conditional Average Treament Effect (TCATE) estimants \citep{souto_diamantis_tcda_2026}, including a reference-distance effect, and 3. an optional two-part representation for an atom at an entirely degenerate unit-level distribution. The proposed model is evaluated against existing benchmark models in the literature, namely Distributional Random Forests (DRF) \citep{DRF-paper} and Causal-DRF \citep{naf2026causaldrf}, report heterogeneous-effect designs explicitly, retain a difficult multimodal setting in which WCF does not lead, and finally apply WCF for three real world settings, namely Project STAR \citep{word1990state}, a randomized class-size experiment, America state-year minimum wages effect, and Kenyan cash transfers \citep{egger2022general,egger2024replication}.

\paragraph{Related work.}

Several literatures study distributional causal questions, but they differ in what constitutes one observational unit. Counterfactual distribution methods identify marginal distributions of a scalar potential outcome and their features \citep{Chernozhukov2013}. Conditional distributional treatment-effect methods instead compare $\Law(Y^1\mid X=x)$ and $\Law(Y^0\mid X=x)$ for scalar or multivariate $Y$. Kernel conditional mean embeddings permit flexible law comparisons and tests \citep{Counterfactualmeanembeddings,CATEGeneralization,martineztaboada2023efficient,jain2026conditional}; robust pseudo-outcome regressions target conditional kernel effects \citep{GRFDistributionalCausalEffects}. In WCF, by contrast, one observation $Y_i$ is already a distribution. The target $P_a^K(x)$ is therefore a law over unit-level distributions, not merely the conditional distribution of scalar observations pooled across units. This additional probability layer is what permits treatment effects on, for example, classroom-specific dispersion while retaining heterogeneity across classrooms.

Distributional Random Forests estimate multivariate conditional laws through forest weights and kernel-based splitting \citep{DRF-paper}; uncertainty procedures for those laws are developed by \citet{naf2023confidence}. Causal-DRF adapts this construction to conditional kernel treatment effects using a treatment-aware shared forest \citep{naf2026causaldrf}. WCF shares the goal of flexible conditional-law estimation but uses a different representation and optimization problem: a covariate-indexed particle law on the monotone quantile cone, fitted directly by a proper energy score. The shared partition and shrinkage of treatment-arm updates are motivated by the same finite-sample tension found in causal forests and Bayesian causal forests, where prognostic structure should be learned without allowing treatment-effect regularization to absorb confounding \citep{wager2018estimation,athey2019generalized,hahn2020bayesian}.

The functional layer connects WCF to semiparametric causal estimation. Standard AIPW scores can estimate marginal or stratum-specific contrasts of scalar summaries under the usual causal assumptions \citep{bang2005doubly,chernozhukov2018double}. WCF uses its out-of-fold particle means as outcome nuisances, but does not claim that doubly robust scalar calibration makes the complete law estimator doubly robust. This separation also clarifies the relation to TCDA \citep{souto_diamantis_tcda_2026}.

Finally, the structural-zero extension is related to two-part models that distinguish participation from intensity \citep{duan1983medical,mullahy1986modified}. Here the first component is an atom at an entirely degenerate distribution, not a zero observation inside an otherwise nondegenerate unit-level distribution. The distinction determines both the statistical model and the meaning of the estimated component probability.

\section{Targets and Estimator}\label{sec:model}

\subsection{Distribution-valued outcomes and causal targets}\label{sec:targets}

For independent units, observe $O_i=(X_i,A_i,Q_i)$, where $X_i\in\mathbb R^d$ contains pretreatment covariates, $A_i\in\{0,1\}$ is treatment, and $Y_i\in\mathcal P_2(\mathbb R)$ is the unit-level outcome distribution. We observe its quantile function $Q_i=q_K(Y_i)$ at levels $0<u_1<\cdots<u_K<1$, where
\begin{align*}
 q_K(Y)&=\{Q_Y(u_1),\ldots,Q_Y(u_K)\},\\
 \QK&:=\{q\in\mathbb R^K:q_1\leq\cdots\leq q_K\}.
\end{align*}
With positive quadrature weights $w_k$ summing to one, define
\[
 d_{W,K}(q,q')=\left\{\sum_{k=1}^K w_k(q_k-q'_k)^2\right\}^{1/2}.
\]
This is a quadrature approximation to the one-dimensional $W_2$ distance and is exactly the Wasserstein distance between the corresponding weighted discrete distributions \citep{peyre2019optimaltransport}. The superscript $a$ denotes a potential outcome, so $Q^a=q_K(Y^a)$.

Assume consistency, conditional exchangeability $Y^a\perp A\mid X$, and positivity $0<e(X)<1$ almost surely, where $e(x)=\Prob(A=1\mid X=x)$ \citep{rosenbaum1983propensity}. These assumptions identify the across-unit conditional law
\begin{align*}
 P_a^K(x)&:=\Law(Q^a\mid X=x)\\
 &=\Law(Q\mid A=a,X=x).
\end{align*}
The distinction between $Y^a$ and $P_a^K(x)$ is important: $Y^a$ is one unit's within-unit distribution, whereas $P_a^K(x)$ describes heterogeneity in those distributions across comparable units.

Let $h:\QK\rightarrow\mathbb R$ be a measurable, integrable scalar summary of a quantile vector. Thus $h(Q)=h\{q_K(Y)\}$ evaluates that summary for the observed unit-level distribution, while $h\circ q_K$ denotes the same mapping written as a function of the original distribution: $(h\circ q_K)(Y)=h\{q_K(Y)\}$. For example, $h$ may return the grid mean, standard deviation, upper-tail mean, a selected quantile, skewness, or distance to a reference vector. The finite-grid outcome-level effects can be defined as:
\begin{align}
 \TATE_{h,K}
 &=\E\{h(Q^1)\}-\E\{h(Q^0)\},\label{eq:tate}\\
 \TCATE_{h,K}(x)
 &=\E\{h(Q^1)-h(Q^0)\mid X=x\}.\label{eq:tcate}
\end{align}
These agree with the TATE formulation of TCDA \citep{souto_diamantis_tcda_2026} after taking its outcome representation to be $h\circ q_K$; equation~\eqref{eq:tcate} is the corresponding conditional extension. No transformed moderator is needed. When results are summarized over a prespecified covariate stratum $B_j\subseteq\mathbb R^d$, one can report the group effect $\E\{h(Q^1)-h(Q^0)\mid X\in B_j\}$, which averages the TCATE over that stratum. 

For a prespecified reference quantile vector $q_\star^K\in\QK$ (i.e., quantile vector of an ideal/desired distribution for $Y$) representing a substantive benchmark, set $h_{\star,K}(q)=d_{W,K}(q,q_\star^K)$. This paper calls \eqref{eq:tate} and \eqref{eq:tcate} with $h=h_{\star,K}$ the reference TATE and reference TCATE. Although it may seem counterintuitive, a negative reference effect is desirable: it indicates that the treatment reduces the expected distance between the unit-level distributions and the "ideal" distribution. These effects compare expected unit-level distances; they neither compare barycenters nor require a joint coupling of $Q^0$ and $Q^1$.

\subsection{Conditional particle laws}\label{sec:particle}

WCF approximates $P_a^K(x)$ by
\begin{align}
 \widehat P_{a,M}^K(x)&=\frac{1}{M}\sum_{m=1}^M
 \delta_{p_{am}^K(x)},\label{eq:particle-law}\\[-2pt]
 &\hspace{2.1cm}p_{am}^K(x)\in\QK.\nonumber
\end{align}
where $K$ is the number of quantile levels, $M$ is the number of equally weighted particles per arm, and $\delta_q$ is the Dirac probability measure at $q$. For $p_{1:M}=(p_1,\ldots,p_M)$ and an observed vector $q$, the boosting loss is the collision-smoothed energy score
\begin{align}
 S_{\varepsilon,M}(p_{1:M};q)
 &=\frac1M\sum_m d_\varepsilon(p_m,q)\nonumber\\
 &\quad-\frac{1}{2M^2}\sum_{m,\ell}d_\varepsilon(p_m,p_\ell),\label{eq:energy}\\
 d_\varepsilon(p,q)&=
 \{d_{W,K}(p,q)^2+\varepsilon^2\}^{1/2}-\varepsilon.
\end{align}
The unsmoothed energy score is strictly proper under a finite first-moment condition \citep{gneiting2007proper}. The first term fits the observed quantile vectors, while the second prevents particle collapse. A shared regression tree maps covariates to arm-specific leaf updates in $\mathbb R^{MK}$. Each update is projected onto $\QK$ via weighted isotonic regression to ensure valid, monotonically increasing quantile vectors. Candidate splits rely on a pooled multi-output criterion and must retain observations from both treatment arms in each child node. Within each leaf, the treatment-arm update contrast is shrunk toward zero while preserving the count-weighted pooled update, mirroring regularization strategies in causal forests \citep{wager2018estimation}. Appendix~\ref{app:algorithm} details the optimization and tuning of the model as well as its asymptotics.

The two fitted arms represent separate marginal conditional laws. Because the particles in the treatment arm are learned independently of those in the control arm, individual particles are not coupled across treatment arms; consequently, differences between individual particles do not represent unit-level treatment effects.

\subsection{Functional calibration and structural mass}\label{sec:calibration}

For a declared functional $h$, define the WCF outcome-regression estimate as the mean of $h$ over the fitted particles,
\[
 \widehat m_{a,h}^{(-f)}(x)
 =\frac{1}{M}\sum_{m=1}^M h\{\widehat p_{am}^{K,(-f)}(x)\},
\]
where $(-f)$ indicates that the observation's fold was excluded when fitting WCF. Let $\widehat e_i^{(-f)}$ be the corresponding cross-fitted estimate of the propensity $e(X_i)$, bounded away from zero and one. With $H_i=h(Q_i)$, define
\begin{align}
 \widehat\phi_i(h)={}&\widehat m_{1,h}^{(-f)}(X_i)-\widehat m_{0,h}^{(-f)}(X_i)\nonumber\\
 &+\frac{A_i}{\widehat e_i^{(-f)}}\{H_i-\widehat m_{1,h}^{(-f)}(X_i)\}\nonumber\\
 &-\frac{1-A_i}{1-\widehat e_i^{(-f)}}\{H_i-\widehat m_{0,h}^{(-f)}(X_i)\}.\label{eq:aipw}
\end{align}
The sample average estimates $\TATE_{h,K}$. Its average within a prespecified stratum $X\in B_j$ estimates the corresponding group effect; a regression of the scores on $X$ can instead target the full TCATE curve. This score has the usual doubly robust mean-zero property when either the propensity model or both arm-specific functional regressions are consistently estimated, subject to positivity, moment, cross-fitting, and nuisance-rate conditions \citep{bang2005doubly,chernozhukov2018double}. This double-robustness property applies to the calibrated scalar functionals, whereas the underlying particle law is estimated as a plug-in conditional distribution. Nevertheless, this does not preclude the particle law from supporting causal estimations; the fitted conditional particles can still be used directly to evaluate TATE, TCATE, and reference-distance effects.

Finally, when some units have the entirely degenerate outcome $Q^a=\mathbf 0_K$, the optional two-part WCF model estimates $\pi_a(x)=\Prob(Q^a\neq\mathbf0_K\mid X=x)$ and the nondegenerate law $P_a^{+,K}(x)$, then assembles
\begin{equation}
 \widehat P_a^{\mathrm{2p},K}(x)=
 \{1-\widehat\pi_a(x)\}\delta_{\mathbf0_K}
 +\widehat\pi_a(x)\widehat P_{a,M}^{+,K}(x).\label{eq:two-part}
\end{equation}
This is a genuine two-part WCF estimator for the structural mass and assembled law. This two-part implementation does not apply the AIPW layer to scalar functionals of this mixture, so its scalar effects are plug-in rather than doubly robust. This limitation concerns calibration evidence, not the availability of the two-part law.

\section{Simulation Studies}\label{sec:simulations}

\subsection{Design and evaluation}\label{sec:sim-design}

The simulations use three complementary families. The location-shape family is a regular-regime benchmark: a placebo (LS0), location-and-shape heterogenous effects that vary with $X_4$ under moderate overlap (LS1), a location heterogenous effect that varies with $X_3$ together with a constant scale change (LS2), and the LS1 outcome surfaces under limited overlap (LS3). The second family is a mechanism-based stress-test suite: a location effect varying with $X_1$ and a scale effect varying with $X_2$ (S1), a stochastic null (S2), equal mean quantile curves with an across-unit spread contrast that varies with $X_4$ (S3), a multimodal outer law with a location effect proportional to $X_1$ (S4), a shape effect modified by $X_1$ (S5), and a weak heterogeneous effect under strong confounding (S6). Conditional effects are evaluated along each design's active moderator: $X_4$ for LS1, LS3, S3, Z1, and Z3; $X_3$ for LS2; $X_2$ for Z2; and $X_1$ for the remaining designs. Thus S1, S4, S5, and S6 directly test effects that vary across strata of $X_1$, while LS1 and LS3 test $X_4$-indexed heterogeneity and LS2 tests $X_3$-indexed heterogeneity; the placebo and null designs LS0, S2, and Z0 retain $X_1$ as stability checks. The structural-zero family contains a placebo (Z0), participation-only (Z1), positive-component-only (Z2), and joint participation and positive-component effects (Z3). Exact equations appear in Appendix~\ref{app:dgps}. 

WCF is compared with DRF and Causal-DRF \citep{DRF-paper,naf2026causaldrf}. The primary experiments use $n=1000$, mirroring the typical scale of the applied studies considered, $R=50$ paired Monte Carlo replications, $K=25$ quantile levels, $M=10$ WCF particles, and an independent test sample of 1000 units in every replication to avoid overfitting. The main text selects LS1, LS3, S4, and S5 because together they test ordinary shape changes, limited overlap, heterogeneous effects, and the principal particle-model failure. Appendix~\ref{app:simulation-results} reports every design in the fifty-replication $n=1000$ comparison. Additionally, sensitivity analyses reported in Appendix~\ref{app:sensitivity} vary the quantile grid and particle budget, the assignment mechanism, and the propensity model, add null companion designs, and include a sample-size study over $n\in\{125,250,500,1000\}$ with ten paired replications (given limited computational budget). Except for the sample-size study, these checks use the same fifty paired replications and confirmatory protocol as the primary comparison.

Concerning evaluation metrics, conditional-law error is squared maximum mean discrepancy (MMD$^2$) between the fitted and quadrature-approximated true conditional laws, averaged over arms and test units, using a common Gaussian kernel within each replication \citep{gretton2012kernel}. In the primary comparison and Appendix~\ref{app:simulation-results}, all functional-effect columns use one convention: Monte Carlo RMSE. For a scalar $h$, marginal RMSE is $[R^{-1}\sum_{r=1}^R\{\widehat\TATE_{h,K}^{(r)}-\TATE_{h,K}\}^2]^{1/2}$; conditional RMSE additionally averages the squared errors over four prespecified strata of the design's active moderator. Reference errors use $h_{\star,K}$. The aggregate functional panels also average squared errors over the grid mean, grid standard deviation, grid skewness, and upper-half mean before taking the square root. Figure bars and table parentheses are Monte Carlo standard errors over the fifty replications of the $n=1000$ comparison. The resolution, assignment, propensity, and null tables in Appendix~\ref{app:sensitivity} follow the same fifty-replication convention, except that native-grid law entries remain Monte Carlo means; the sample-size study uses ten replications and states its own convention. For the two-part designs, mass error evaluates $1-\pi_a(x)$ and mass-TCATE error evaluates its stratum contrast. Every reported criterion is an error, so smaller is better.

\subsection{Results}\label{sec:sim-results}

\begin{figure}[t]
\centering
\includegraphics[width=0.94\textwidth]{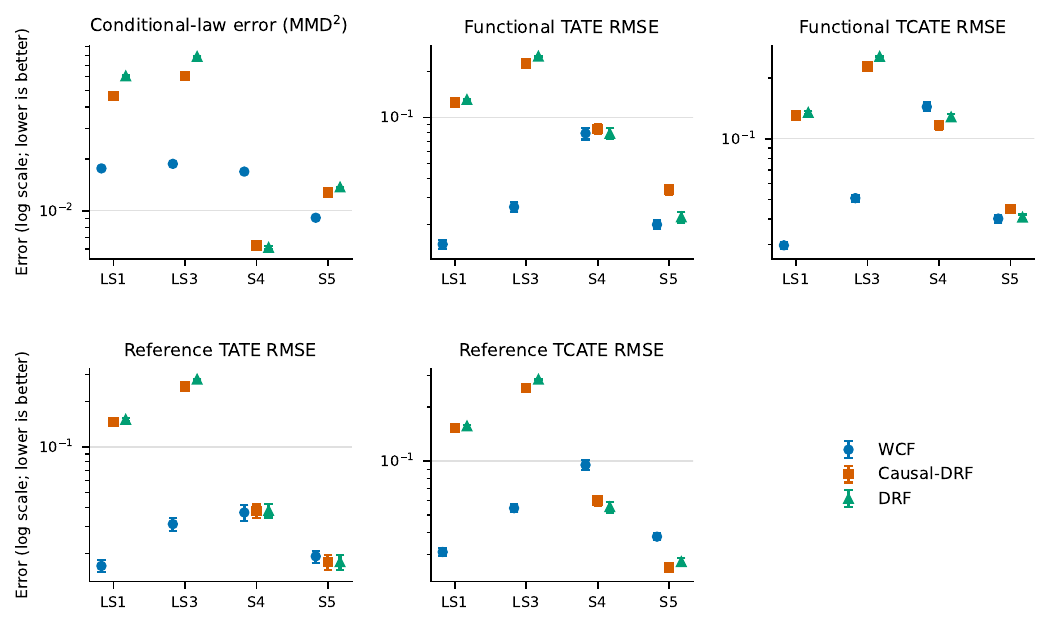}
\caption{Simulation accuracy on selected designs at $n=1000$ and $R=50$ paired replications. Law points are mean MMD$^2$; the other panels are Monte Carlo RMSEs. Bars show one Monte Carlo standard error. LS1 changes location and shape, LS3 combines the same effect with limited overlap, S4 has a multimodal outer law with a heterogeneous location effect, and S5 has heterogeneous shape effects. All vertical axes use a log scale.}
\label{fig:simulation-summary}
\end{figure}

Figure~\ref{fig:simulation-summary} shows all five principal criteria. In the smooth location-and-shape designs LS1 and LS3, WCF has the smallest error throughout. Its law errors are $0.0176$ and $0.0187$, compared with $0.0466$ and $0.0608$ for the stronger forest baseline, while its reference-TCATE RMSEs are $0.0311$ and $0.0546$, compared with $0.1526$ and $0.2551$. The advantage persists under LS3's deteriorating overlap, consistent with a shared covariate partition and a particle law fitted directly with the Wasserstein energy score. 

The S designs show that global law accuracy and accuracy for one nonlinear functional need not have the same ordering. In S4, the separated multimodal outer law is poorly matched to a small equally weighted particle cloud: WCF loses the law and reference-TCATE criteria, while it retains the lowest reference-TATE point estimate. In S5, WCF again has the best law error under heterogeneous shape effects, but it loses the reference-TCATE criterion and its reference-TATE comparison is within paired uncertainty. These reversals are not contradictions because MMD$^2$ assesses the whole fitted law, whereas the reference criteria assess one nonlinear functional after scalar calibration.

Taken together, WCF is most attractive when the scientific target is the conditional law or several functionals of a smooth, nonmultimodal law, especially when shared structure across treatment arms is valuable. The results do not show a general failure under heterogeneity: WCF leads in all LS designs with heterogeneous effects, leads in heterogeneous S1, and is competitive on the aggregate functionals in S5. They instead identify target-specific difficulty for multimodality. 

Briefly about the degenerated family found in Appendix~\ref{app:simulation-results}: the two-part WCF estimator has the lowest law, structural-mass, and mass-TCATE errors across Z0 to Z3. In Z3, its law error is $0.0187$ and its mass error is $0.0622$, compared with $0.0796$ and $0.1782$ for Causal-DRF and $0.2671$ and $0.3624$ for DRF. These results evaluate the assembled law and participation component, not AIPW-calibrated mixture effects.

Sensitivity checks vary $K\in\{5,25,49\}$, $M\in\{5,10,25\}$ and  $n\in\{125,250,500,1000\}$. Under the sensitivity protocol, native-grid law, mean-quantile, and end-to-end reference errors change by at most a few percent across $K$ and $M$, so $(K,M)=(25,10)$ remains the primary computational setting; because the native target changes with $K$, these comparisons do not separate quadrature resolution from tail coverage. Assignment experiments at both $n=500$ and $n=1000$ show that WCF has the lowest end-to-end reference-TCATE error in all four mechanisms at $n=500$ and in three of four at $n=1000$, where DRF is slightly more accurate under linear assignment. The reported errors are nearly invariant to the propensity factory because the confirmatory AIPW layer is fixed, while the estimator's internal calibration diagnostic flags the untuned gradient-boosting factory as three to four times less stable than the logistic, random-forest, or oracle alternatives. On the null companions, WCF's pooled false-effect error on the grid-mean functionals is roughly three times smaller than the forest baselines. Additionaly, the sample-size sensitivity study has convergent results to the main simulation studies, namely WCF has the lowest law error in all 16 LS cells, 20 of 24 S cells, and all 16 Z cells. Yet, it is worth mentioning that reference-TCATE error deteriorates faster as $n$ decreases: WCF remains best throughout LS, but in S it leads only in S1 and its small-sample disadvantage is largest for multimodal S4 and weak-effect, strong-confounding S6. Appendix~\ref{app:sensitivity} gives the complete numerical tables and trajectories, together with the null, assignment, and resolution checks.

\section{Project STAR Application}\label{sec:applied}

Project STAR randomized pupils within schools to small classes, regular classes, or regular classes with a full-time teacher aide \citep{word1990state,krueger1999experimental,chetty2011kindergarten}. We reshape 24,610 tested students into 1,333 classroom-grade units: 522 small-class units and 811 controls, of which 400 are regular classes and 411 have an aide. Each unit is represented by $K=25$ quantiles of within-grade standardized mathematics scores. Small-class units contain 7 to 20 tested students and control units 13 to 29, so their observed quantile vectors differ in measurement precision. The learner receives nine pretreatment or classroom-composition covariates: school free/reduced-lunch share, four teacher characteristics (sex, race, graduate degree, and years of experience), school urbanicity, and the classroom shares of female, Black, and free/reduced-lunch students. The four prespecified strata $B_j$ are quartiles of the school free/reduced-lunch share and are nearly balanced, with 333, 333, 334, and 333 classroom-grade units.

The reference distribution is the grade average of pooled control-student quantiles. Because outcomes are standardized within grade and the benchmark is constructed from controls, reference effects describe movement toward or away from a blended control distribution rather than toward an external welfare standard. Fitted propensities range from $0.232$ to $0.551$, their treated and control means are $0.394$ and $0.390$, and none lies outside $[0.05,0.95]$. Overlap is therefore not a binding concern.

\begin{figure}[t]
\centering
\includegraphics[width=0.92\linewidth]{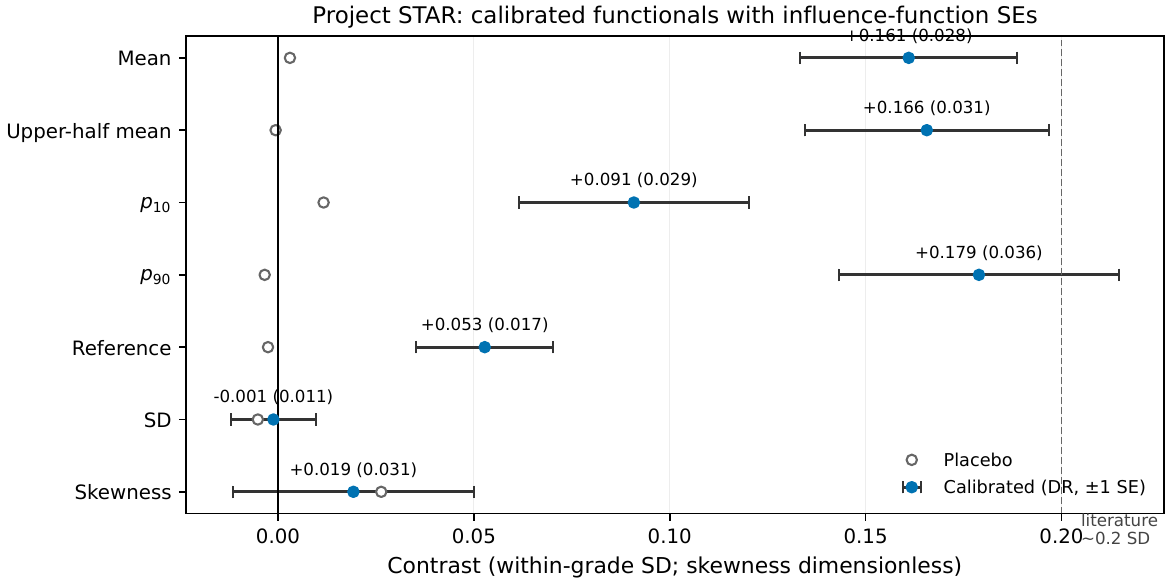}
\caption{Project STAR calibrated functional contrasts with influence-function standard errors (bars). Open circles report the permuted-treatment placebo, and the dashed line marks the experimental literature value of about one fifth of a within-grade standard deviation for the mean effect. Skewness is dimensionless; all other contrasts are in within-grade standard deviations.}
\label{fig:applied-star-estimates}
\end{figure}

\begin{figure}[t]
\centering
\includegraphics[width=0.96\textwidth]{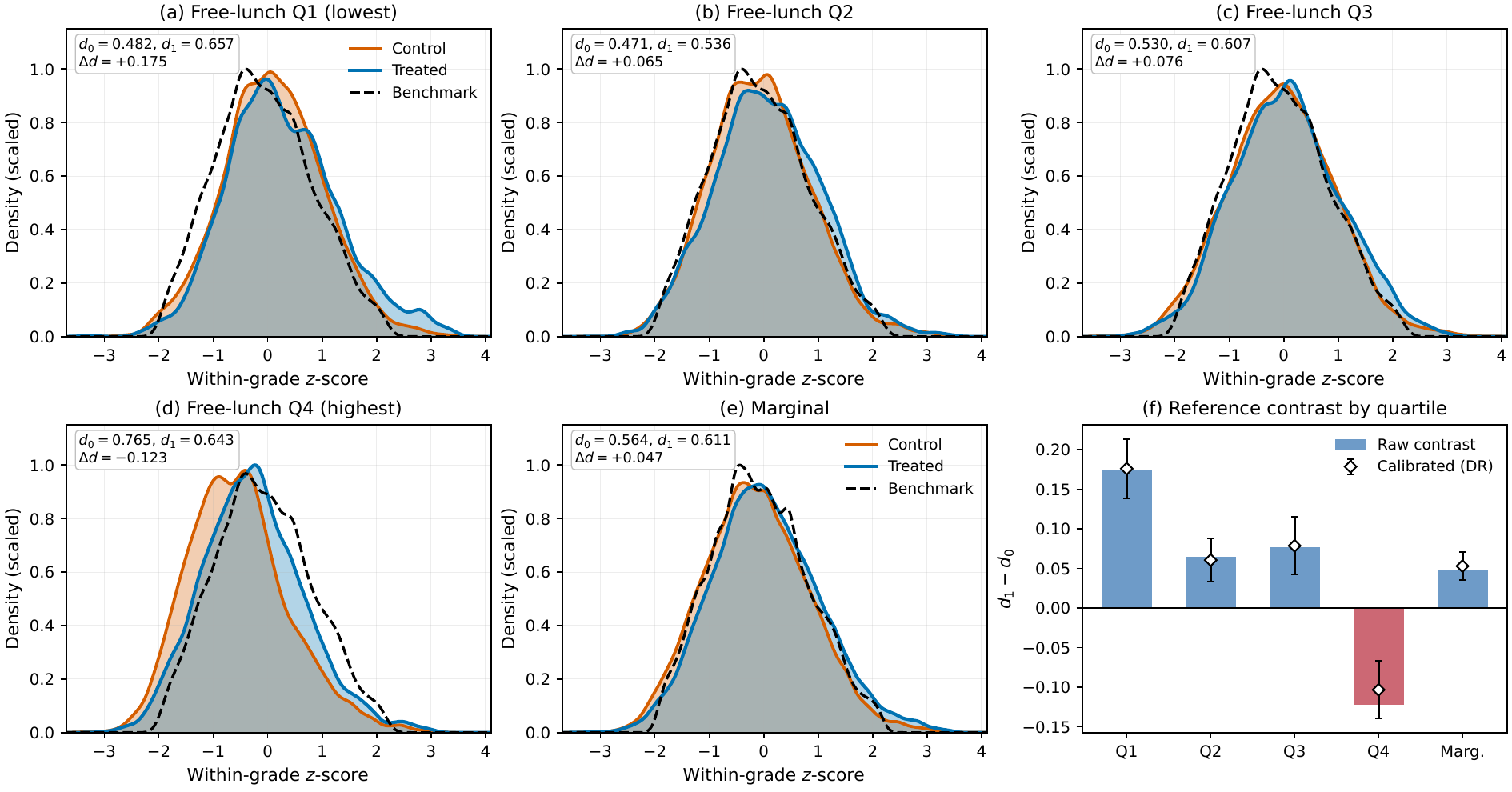}
\caption{Project STAR arm densities relative to the pooled control benchmark, by school free-lunch quartile. Panels Q1 to Q4 and the marginal panel show unadjusted equal-unit mixtures of classroom-grade laws. The final panel compares raw and calibrated reference contrasts. The treated distribution separates most clearly in Q4, where the control law lies below the blended benchmark and treatment moves it toward that benchmark.}
\label{fig:applied-star-densities}
\end{figure}

Figure~\ref{fig:applied-star-estimates} depicts the overall results. Marginally, the calibrated mean contrast is $+0.161$ (influence-function SE $0.028$), near the experimental literature's student-level estimate of approximately $0.2$ standard deviations and the WCF plug-in estimate of $+0.150$. The upper-half mean and ninetieth-percentile contrasts are $+0.166$ and $+0.179$, compared with $+0.091$ at the tenth percentile. Small classes therefore produce larger gains in the upper part of the classroom distribution rather than a uniform location shift. Dispersion is essentially unchanged ($-0.001$), and the reference TATE contrast is $+0.053$, indicating that the treatment shifts the distribution slightly farther from the pooled control benchmark. Combined with the positive mean contrast, this reveals that small classes shift the entire classroom score distribution to the right of the control reference; hence achieving a better performance.

The stratum summaries reveal additional structure. Mean effects are $+0.164$, $+0.092$, $+0.126$, and $+0.262$ from the lowest to highest free-lunch quartile; the upper-tail contrasts follow the same pattern. To understand how these heterogenous gains interact with the distributional baseline, we examine the reference-distance effects found in Figure~\ref{fig:applied-star-densities}. Across most schools, the treatment pushes the outcome distribution farther from the pooled control benchmark, except in the most vulnerable classrooms. In Q4, the control median is $-0.53$ while the blended benchmark median is $-0.11$, so the upward shift from small classes moves the outcome law closer to the benchmark and produces a reference contrast of $-0.103$. In Q1 to Q3 the control distributions are already at or above the blend, and similar upward shifts move them farther away. 

Besides this applied study, Appendix~\ref{app:applications} reports results for two additional settings: an analysis of Kenyan cash transfers \citep{egger2022general,egger2024replication}, which reveals lower-tail gains and movement toward an external consumption benchmark, and an evaluation of U.S. state-year minimum wage policies, which highlights compression and rightward shifts in regional wage distributions.

\section{Conclusion}\label{sec:conclusion}

Wasserstein Causal Forests provide a flexible framework for estimating treatment-arm conditional laws of distribution-valued outcomes and calibrating selected scalar functionals. The simulation studies demonstrate strong performance under location, shape, and overlap variations, including settings where mean effects are entirely absent, while also highlighting limitations, such as the inadequacy of small equal-weight particle clouds for complex multimodal laws. Beyond the primary simulations, empirical evaluations in Project STAR, Kenyan cash-transfer experiments, and state-level minimum wage analyses show how distributional estimation uncovers rich structural changes, ranging from upper-tail classroom gains and lower-tail consumption shifts to regional wage compression.

Nonetheless, reference effects must be interpreted with care: movement toward a benchmark is not inherently beneficial, nor is movement away inherently harmful, as success depends entirely on a scientifically defensible choice of target. Additional limitations include finite $K$ and $M$ approximations, the absence of formal uniform coverage guarantees, and quantile measurement error arising from finite lower-level sample sizes. Furthermore, clustered and design-based extensions remain necessary for dependent units. Within these boundaries, estimating the conditional distribution as a primary object offers a robust, reusable representation for causal questions that cannot be reduced to a single regression target.

\bibliography{references}

\clearpage
\appendix
\input{supplement}

\end{document}

%% file: supplement.tex
\section{Estimator Details and Asymptotic Theory}\label{app:algorithm}

\subsection{Observations, probability layers, and causal targets}

One observational unit contributes $O_i=(X_i,A_i,Q_i)$, where $X_i\in\mathbb R^d$ is a vector of pretreatment adjustment variables, $A_i\in\{0,1\}$ is treatment, and $Q_i$ represents that unit's outcome distribution. Specifically, if $Y_i$ is a probability distribution on $\mathbb R$, then $Q_i=q_K(Y_i)=(Q_{Y_i}(u_1),\ldots,Q_{Y_i}(u_K))$ at fixed levels $0<u_1<\cdots<u_K<1$. Thus $n$ counts units, $K$ counts coordinates of each unit's outcome, and $M$ below counts particles used to represent heterogeneity \emph{across} units. These three sample or resolution parameters have different roles. The ordered vectors belong to $\QK=\{q\in\mathbb R^K:q_1\leq\cdots\leq q_K\}$. For fixed $w_k>0$ with $\sum_k w_k=1$, write
\[
 \|v\|_w=\Big(\sum_{k=1}^K w_kv_k^2\Big)^{1/2},\qquad
 d_{W,K}(q,r)=\|q-r\|_w.
\]
The latter is the $W_2$ distance between $\sum_k w_k\delta_{q_k}$ and $\sum_k w_k\delta_{r_k}$, by the one-dimensional quantile formula \citep[Eq.~(2.36)]{peyre2019optimaltransport}. It approximates the distance between the original distributions only to the accuracy of their quantile-grid representation.

Write $Q^a=q_K(Y^a)$ for the potential quantile vector under arm $a$, $e(x)=\Prob(A=1\mid X=x)$, $e_1=e$, and $e_0=1-e$. Consistency, $Y^a\perp A\mid X$, and $0<e(X)<1$ identify
\[
 P_a(x):=P_a^K(x)=\Law(Q^a\mid X=x)=\Law(Q\mid A=a,X=x).
\]
Throughout this appendix $K$ and $w$ are fixed unless explicitly stated otherwise, so the superscript $K$ is usually suppressed. A draw from $P_a(x)$ is a whole ordered vector, not one scalar draw from $Y^a$. Conditional laws and their means are identified $P_X$-almost everywhere; a value at an individual point of a continuous covariate distribution requires a chosen regular version, for example one continuous near that point.

For a prespecified measurable summary $h:\QK\to\mathbb R$ with $\E|h(Q^a)|<\infty$ for each arm, put $H=h(Q)$ and
\[
 \mu_{a,h}(x)=\int h(q)\,dP_a(x)(q),\qquad
 \tau_h(x)=\mu_{1,h}(x)-\mu_{0,h}(x),\qquad
 \theta_h=\E\{\tau_h(X)\}.
\]
These are, respectively, the arm-specific conditional mean, $\TCATE_{h,K}(x)$, and $\TATE_{h,K}$. For a fixed measurable stratum $B$ with $p_B=\Prob(X\in B)>0$, define $\theta_{h,B}=\E\{\tau_h(X)\mid X\in B\}$. A stratum effect is not a pointwise TCATE. Conditioning on a moderator or a stratum also does not replace adjustment for the full $X$ in the arm regressions and propensity.

Examples are the grid mean $\sum_k w_kq_k$, the grid standard deviation $[\sum_k w_k(q_k-\sum_\ell w_\ell q_\ell)^2]^{1/2}$, and a specified coordinate $q_j$. For a fixed benchmark $q_\star\in\QK$, $h_\star(q)=\|q-q_\star\|_w$ gives the reference TATE and TCATE. A negative reference effect means a reduction in expected distance to this benchmark, rather than an ordering of welfare without further scientific justification. All these targets average a function of each unit's distribution. In general $\int h\,dP_a\neq h(\int q\,dP_a)$; in particular, distance to the mean quantile vector does not replace mean distance to the reference. Neither identification nor estimation of these contrasts requires a joint law of $(Q^0,Q^1)$.

\subsection{Particle representation and the complete fitting step}

For each $x$ and arm $a$, WCF returns
\[
 \widehat P_a(x)=M^{-1}\sum_{m=1}^M\delta_{p_{am}(x)},\qquad p_{am}(x)\in\QK.
\]
Each particle is a possible unit-level quantile vector; it is not a confidence draw or a matched potential outcome. Particle labels within an arm are arbitrary, and cross-arm differences $p_{1m}-p_{0m}$ have no individual causal interpretation. Functional predictions are $\widehat m_{a,h}(x)=M^{-1}\sum_m h\{p_{am}(x)\}$.

For $\varepsilon>0$, define $r_\varepsilon(p,q)=(\|p-q\|_w^2+\varepsilon^2)^{1/2}$ and $d_\varepsilon(p,q)=r_\varepsilon(p,q)-\varepsilon$. The observed-arm empirical objective is
\begin{equation}
 \widehat R_\varepsilon(p)=\frac1n\sum_{i=1}^n
 \left\{\frac1M\sum_m d_\varepsilon(p_{A_i m}(X_i),Q_i)
 -\frac1{2M^2}\sum_{m,\ell}d_\varepsilon(p_{A_i m}(X_i),p_{A_i\ell}(X_i))\right\}.
 \label{eq:app-empirical-risk}
\end{equation}
Write $S_{\varepsilon,M}(p_{1:M};q)$ for the expression in braces for one outcome $q$ and one particle cloud. The double sum includes diagonal pairs, whose contribution is zero. The first term attracts particles toward observed vectors and the second rewards dispersion. Smoothing makes the gradient well defined at coincident particles. No propensity weights enter this fitting objective: conditional on $X=x$, each observed arm supplies data from its own identified law. Overlap is needed to learn both laws over the target covariate population.

Both arms start from the same $M$ pooled observed vectors. The implementation orders the pooled vectors lexicographically and takes the indices $\lfloor(m-1/2)n/M\rfloor$, $m=1,\ldots,M$, using zero-based indexing. This is a deterministic initialization, not a multivariate quantile definition. At iteration $t$, abbreviate $p_m=p_{A_i m}^{(t)}(X_i)$. The negative preconditioned gradient for coordinate $k$ is
\begin{equation}
 G_{imk}^{(t)}=-\frac{M}{w_k}\frac{\partial S_{\varepsilon,M}(p_{1:M};Q_i)}{\partial p_{mk}}
 =-\frac{p_{mk}-Q_{ik}}{r_\varepsilon(p_m,Q_i)}
 +\frac1M\sum_{\ell=1}^M\frac{p_{mk}-p_{\ell k}}{r_\varepsilon(p_m,p_\ell)}.
 \label{eq:app-gradient}
\end{equation}
Flattening $(m,k)$ yields one $MK$-dimensional response $G_i^{(t)}$ per unit. A single tree uses pooled multi-output squared-error reduction to choose covariate splits. For a node $I$, its criterion is $\sum_{i\in I}\|G_i^{(t)}-\overline G_I^{(t)}\|_2^2$; a candidate split maximizes its reduction subject to the total and arm-specific minimum child counts. The partition is shared, but the terminal-node prediction is arm specific. This is a boosting tree construction, so asymptotic results for honest subsampled random forests do not apply merely because the estimator is named WCF.

In a terminal leaf with $n_0,n_1>0$, let $n_a$ count arm-$a$ observations and let $u_a\in\mathbb R^{MK}$ be their mean gradient target. For $\lambda\geq0$, define
\[
 \bar u=\frac{n_0u_0+n_1u_1}{n_0+n_1},\quad
 \Delta_u=u_1-u_0,\quad n_{\rm eff}=\frac{n_0n_1}{n_0+n_1},\quad
 c_\lambda=\frac{n_{\rm eff}}{n_{\rm eff}+\lambda}.
\]
The leaf predictions are
\[
 \widehat u_0=\bar u-\frac{n_1}{n_0+n_1}c_\lambda\Delta_u,\qquad
 \widehat u_1=\bar u+\frac{n_0}{n_0+n_1}c_\lambda\Delta_u.
\]
They minimize $\sum_a n_a\|v_a-u_a\|_2^2+\lambda\|v_1-v_0\|_2^2$ over $(v_0,v_1)$. Thus $\lambda=0$ retains the raw arm means, while larger $\lambda$ shrinks their contrast without changing their count-weighted mean. The leaf containing a new $x$ supplies the update block $\widehat u_{am}^{(t)}(x)\in\mathbb R^K$. WCF then sets
\[
 p_{am}^{(t+1)}(x)=\Pi_{\QK}^w\{p_{am}^{(t)}(x)+\eta_t\widehat u_{am}^{(t)}(x)\},\qquad
 \Pi_{\QK}^w(v)=\arg\min_{q\in\QK}\sum_k w_k(q_k-v_k)^2.
\]
This projection is unique because the objective is strictly convex and coercive on the nonempty closed convex cone $\QK$, and it is computed by weighted pool-adjacent-violators \citep[Section 2, Algorithm 1]{busing2022monotone}. It is nonexpansive: the two projection optimality inequalities give $\|\Pi_{\QK}^w(v)-\Pi_{\QK}^w(v')\|_w^2\leq\langle\Pi_{\QK}^w(v)-\Pi_{\QK}^w(v'),v-v'\rangle_w$, and weighted Cauchy--Schwarz completes the bound. Here $\langle u,v\rangle_w=\sum_k w_ku_kv_k$. Backtracking halves $\eta_t$ until the observed-arm objective does not increase; fitting stops if its backtracking budget finds no acceptable step. This guarantees accepted-step descent, not global optimization of the nonconvex particle objective.

The reported fits use at most 100 iterations, initial step size $0.12$, depth 4, minimum total leaf size 10, minimum arm-specific leaf size 5, $\varepsilon=10^{-3}$, and $\lambda\in\{0,50,500\}$ selected by held-out observed-arm energy risk. Outcome fitting uses three folds for LS and two for S and Z. Propensity estimation uses five-fold logistic regression, clipped to $[0.02,0.98]$. The selected learner is refitted to obtain the final law; out-of-fold predictions supply calibration. DRF uses 2,500 trees per arm and Causal-DRF uses 2,500 shared trees \citep{DRF-paper,naf2026causaldrf}. These are finite experimental budgets, not sequences already proved to satisfy the asymptotic conditions below.

\subsection{What the energy objective identifies}

For any probability laws $F,G$ on $\QK$ with finite first $\|\cdot\|_w$ moments, let
\[
 S_\varepsilon(F;q)=\E_{Z\sim F}d_\varepsilon(Z,q)
 -\tfrac12\E_{Z,Z'\sim F}d_\varepsilon(Z,Z'),\qquad
 D_\varepsilon(F,G)=2\E d_\varepsilon(Z,V)-\E d_\varepsilon(Z,Z')-\E d_\varepsilon(V,V'),
\]
where $Z,Z'$ are independent with law $F$, $V,V'$ are independent with law $G$, and all four draws are mutually independent. $D_\varepsilon$ is an energy divergence between \emph{outer} laws, whereas $d_{W,K}$ compares two unit-level vectors.

\paragraph{Proposition A.1 (strict propriety, including fixed smoothing).}
For each fixed $\varepsilon\geq0$,
\begin{equation}
 \E_{V\sim G}\{S_\varepsilon(F;V)-S_\varepsilon(G;V)\}
 =\tfrac12D_\varepsilon(F,G)\geq0,
 \qquad D_\varepsilon(F,G)=0\ \Longleftrightarrow\ F=G.
 \label{eq:app-proper}
\end{equation}
In particular, fixed positive smoothing does not change the population-optimal unrestricted law.

\paragraph{Proof.}
The equality follows by expansion. The loss has the kernel-score form of \citet[Section 5.1]{gneiting2007proper}, with the sign changed from a reward to a loss. We establish strictness for fixed smoothing directly. Transform $q$ to $z=(\sqrt{w_1}q_1,\ldots,\sqrt{w_K}q_K)$, an invertible linear map. For $r\geq0$,
\[
 \sqrt{r^2+\varepsilon^2}-\varepsilon
 =\frac1{2\sqrt\pi}\int_0^\infty(1-e^{-tr^2})e^{-\varepsilon^2t}t^{-3/2}\,dt.
\]
For $r>0$ this identity follows by differentiating in $r^2$ and using $\int_0^\infty e^{-at}t^{-1/2}dt=\sqrt{\pi/a}$, then integrating from zero; continuity gives the endpoint. Consequently $D_\varepsilon(F,G)$ is the integral, with this positive weight, of
$\E k_t(Z,Z')+\E k_t(V,V')-2\E k_t(Z,V)$, where $k_t(z,v)=e^{-t\|z-v\|_2^2}$ and the expectations now use transformed draws. If $\varphi_F,\varphi_G$ denote their characteristic functions, the Fourier transform of a Gaussian expresses this integrand as
\[
 (4\pi t)^{-K/2}\int_{\mathbb R^K}
 e^{-\|\omega\|_2^2/(4t)}|\varphi_F(\omega)-\varphi_G(\omega)|^2\,d\omega.
\]
The Gaussian weight is integrable and characteristic functions are bounded, so Fubini applies. The expression is nonnegative; if it is zero, continuity of the characteristic functions makes them equal everywhere, and uniqueness of characteristic functions gives $F=G$. Finite first moments justify integration of the constituent $1-k_t$ terms in the outer integral. Its weight is positive for every $t>0$, proving strictness, also when $\varepsilon=0$. This also explains why strict propriety does not imply that a restricted $M$-particle fit exactly reproduces an arbitrary law. $\square$

Let $R_\varepsilon(F)=\E S_\varepsilon(F_A(X);Q)$ for a pair $F=(F_0,F_1)$ of measurable conditional laws, and let $P=(P_0,P_1)$ denote the truth. For population risk comparisons assume
\[
 \E\|Q\|_w<\infty,\qquad
 \E_X\sum_a e_a(X)\int\|q\|_w\,dF_a(X)(q)<\infty
\]
for every candidate $F$ under consideration. These conditions make both risks finite, since $d_\varepsilon(q,r)\leq\|q-r\|_w$; finite conditional moments at almost every $x$ alone do not ensure this integrability across $X$. Conditioning on $(X,A)$ then gives the exact identity
\begin{equation}
 R_\varepsilon(F)-R_\varepsilon(P)
 =\tfrac12\E_X\sum_{a=0}^1 e_a(X)D_\varepsilon(F_a(X),P_a(X)).
 \label{eq:app-risk-id}
\end{equation}
Thus positivity identifies both conditional laws at the population optimum. This is not double robustness: the law fit has no propensity-based bias correction.

\subsection{Conditional-law consistency and its algorithmic conditions}

\paragraph{Theorem A.2 (excess-risk bound and consistency).}
Suppose the units are i.i.d., the causal assumptions and population-risk integrability conditions above hold, and $c\leq e(X)\leq1-c$ almost surely for some $c>0$. Fix $K\geq1$, positive weights $w$ summing to one, and $\varepsilon\geq0$. Let $\mathcal F_n$ be a nonempty deterministic class of pairs of measurable equal-weight $M_n$-particle laws, and suppose the fitted predictor $\widehat F\in\mathcal F_n$ is jointly measurable in the training data and a new covariate. Assume also that the supremum below is measurable, as holds for the finite and compact finite-dimensional particle classes considered here. Define
\[
 U_n=\sup_{F\in\mathcal F_n}|\widehat R_\varepsilon(F)-R_\varepsilon(F)|,\qquad
 b_n=\inf_{F\in\mathcal F_n}\{R_\varepsilon(F)-R_\varepsilon(P)\}.
\]
If $\widehat R_\varepsilon(\widehat F)\leq\inf_{F\in\mathcal F_n}\widehat R_\varepsilon(F)+o_n$ for a nonnegative optimization error $o_n$, then
\begin{equation}
 \E_X\sum_{a=0}^1D_\varepsilon(\widehat F_a(X),P_a(X))
 \leq\frac{2}{c}(2U_n+b_n+o_n).
 \label{eq:app-oracle}
\end{equation}
Here and below $\E_X$ integrates a fresh covariate conditional on the fitted functions. If $U_n,b_n,o_n\to_p0$, the left side tends to zero. If, in addition, all true and fitted laws are supported on a common compact $C\subset\QK$, then, for $a=0,1$,
\begin{equation}
 \E_X\mathcal W_{1,w}(\widehat F_a(X),P_a(X))\to_p0,\qquad
 \mathcal W_{1,w}(F,G)=\inf_{\gamma\in\Gamma(F,G)}\int\|q-r\|_w\,d\gamma(q,r),
 \label{eq:app-law-consistency}
\end{equation}
where $\Gamma(F,G)$ is the set of couplings of the two outer laws.

\paragraph{Proof.}
Adding and subtracting empirical risks gives $R_\varepsilon(\widehat F)-R_\varepsilon(P)\leq2U_n+b_n+o_n$; an arbitrarily close approximate minimizer handles a nonattained infimum. Equation~\eqref{eq:app-risk-id} and $e_a\geq c$ prove the bound. On compact $C$, the space of probability laws is weakly compact and $\mathcal W_{1,w}$ metrizes weak convergence \citep[Remark 2.18]{peyre2019optimaltransport}. The bounded continuous distance kernel makes $D_\varepsilon$ continuous under this topology, and its zero set is the diagonal by Proposition A.1. For any $\delta>0$, its minimum $\kappa_\delta$ over pairs with $\mathcal W_{1,w}(F,G)\geq\delta$ is strictly positive whenever that set is nonempty. If the set is empty, all such transport distances are below $\delta$ and no minimum is needed. Otherwise, if $L_C$ is the diameter of $C$, then
\[
 \E_X\mathcal W_{1,w}(\widehat F_a,P_a)
 \leq\delta+\frac{L_C}{\kappa_\delta}\E_XD_\varepsilon(\widehat F_a,P_a).
\]
First let $n\to\infty$ and then $\delta\downarrow0$. $\square$

The three terms in \eqref{eq:app-oracle} have distinct meanings. $U_n$ controls overfitting, $b_n$ combines particle and covariate-function approximation, and $o_n$ measures failure to optimize within the declared class. For example, on compact $C$, scores lie in $[-L_C/2,L_C]$. A finite deterministic class of size $N_n$ therefore has $U_n=O_p(\sqrt{\log(2N_n)/n})$ by Hoeffding's inequality \citep[Theorem 2]{hoeffding1963probability} and a union bound over the class and both deviation signs. For an infinite class with a uniform $\delta_n$-net of the score functions of size $N_n$, approximation of the empirical and population means adds at most $2\delta_n$, giving $O_p(\delta_n+\sqrt{\log(2N_n)/n})$. Thus increasing model complexity can be allowed if this entropy grows slowly enough. These are sufficient conditions, not an entropy calculation for the implemented adaptive booster.

Particle approximation itself admits a useful quantitative check. If $Z_1,\ldots,Z_M$ are i.i.d. from a fixed $P$ and $P_M=M^{-1}\sum_m\delta_{Z_m}$, direct expansion yields
\begin{equation}
 \E D_\varepsilon(P_M,P)=M^{-1}\E d_\varepsilon(Z,Z').
 \label{eq:app-particle-rate}
\end{equation}
On compact $C$ the right side is at most $L_C/M$, so an equal-weight cloud with no larger divergence exists. This is an approximation statement at a fixed conditional law, not a convergence rate for fitted WCF. To make $b_n\to0$, the covariate-indexed model class must also approximate the conditional laws. Nonvacuous examples include a finite covariate partition on which the true laws are constant equal-weight clouds, with those clouds in the fitting class; there $b_n=0$ and bounded finite-dimensional parameter classes admit a uniform law of large numbers. More general laws require growing particle and partition capacity. Conversely, on compact $C$ the class of laws with at most $M$ atoms is weakly closed. A nonatomic conditional law therefore cannot be consistently recovered with $M$ fixed, even by perfect risk minimization.

For WCF, accepted-step descent alone does not establish $o_n\to0$, and fixed depth and a fixed number of boosting iterations do not establish $b_n\to0$. Theorem A.2 identifies exactly what must be verified for a growing-budget implementation; it is not an unconditional consistency theorem for the reported 100-step, ten-particle algorithm. The compact-support assumption is a sufficient regime for the stated transport conclusion and does not cover the Gaussian-tail simulations literally. The excess-risk inequality still holds under finite moments; extending its transport conclusion to unbounded laws requires a separate tightness and uniform-integrability argument. Neither TCDA stability nor random-forest consistency supplies these missing optimization and approximation guarantees. The tail condition is substantive: for $F_j=(1-j^{-1})\delta_{\mathbf0_K}+j^{-1}\delta_{j\mathbf1_K}$ and $G=\delta_{\mathbf0_K}$, one has $D_\varepsilon(F_j,G)=2j^{-2}(\sqrt{j^2+\varepsilon^2}-\varepsilon)\to0$ but $\mathcal W_{1,w}(F_j,G)=1$. Thus even the grid mean need not converge from vanishing energy divergence without tail control.

\paragraph{Corollary A.2a (a concrete growing shared-partition class).}
The risk conditions can be made quantitative for a comparison class of particle predictors. In addition to the compact-support conditions of Theorem A.2, suppose $X\in[0,1]^d$ and each $P_a(x)$ has a version satisfying $\mathcal W_{1,w}(P_a(x),P_a(x'))\leq L_X\|x-x'\|_2$. Partition the covariate cube into $J_n$ deterministic cubes of side length $\ell_n$, where $J_n=O(\ell_n^{-d})$, and allow an arbitrary $M_n$-particle cloud in $C$ for each arm and cell. This is a shared-partition particle class, representable by a sufficiently large axis-aligned tree. For an approximate global empirical minimizer in this class,
\begin{equation}
 \E_X\sum_aD_\varepsilon(\widehat F_a(X),P_a(X))
 =O_p\left(\ell_n+M_n^{-1}
 +\sqrt{\frac{K M_nJ_n\log n}{n}}+o_n\right).
 \label{eq:app-sieve-rate}
\end{equation}
For fixed $K,d$, the choice $\ell_n\asymp M_n^{-1}\asymp(\log n/n)^{1/(d+3)}$ makes the first three terms of order $(\log n/n)^{1/(d+3)}$. This is an upper bound on energy divergence, not a sharp minimax rate or a transport-rate bound.

\paragraph{Proof.}
In a cell with positive arm-specific probability let $G_{a,j}=\Law(Q\mid A=a,X\text{ in cell }j)$. Mixture coupling and the Lipschitz assumption give $\mathcal W_{1,w}(G_{a,j},P_a(x))\leq L_X\sqrt d\,\ell_n$ within the cell. The score is $2$-Lipschitz as a function of its forecast law in $\mathcal W_{1,w}$: the attraction term contributes one Lipschitz constant and the half-weighted pair term contributes one more. Thus the piecewise $G_{a,j}$ forecast has excess risk at most $2L_X\sqrt d\,\ell_n$. Replacing each $G_{a,j}$ by an independent empirical $M_n$-cloud increases its expected risk, averaged over the cell's observed outcomes, by at most $L_C/(2M_n)$ using \eqref{eq:app-particle-rate}. Hence some deterministic set of cell clouds achieves $b_n\leq2L_X\sqrt d\,\ell_n+L_C/(2M_n)$. Cells of zero probability can be filled arbitrarily.

For fixed $K$, a compact subset of $\mathbb R^K$ has a $\delta$-net of size at most $(C_0/\delta)^K$ for $0<\delta<1$, where $C_0$ depends on $C,w,K$. Quantizing the $2M_nJ_n$ particle locations induces a $2\delta$ uniform net of the score functions with log cardinality at most $2KM_nJ_n\log(C_0/\delta)$. Set $\delta=n^{-1}$ and use bounded-score concentration, then substitute the resulting bound for $U_n$ and the approximation bound into \eqref{eq:app-oracle}. $\square$

This construction verifies that particle count and partition complexity can grow jointly while retaining risk consistency when $o_n\to_p0$. It still requires approximate global minimization. The greedy boosting path, its shrinkage schedule, and its stopping rule would need a separate comparison with such a class before \eqref{eq:app-sieve-rate} could be asserted for the experimental algorithm. Keeping this distinction makes the available statistical result explicit without assuming that training-score descent proves optimization consistency.

\subsection{From a fitted law to TATE, TCATE, and reference effects}

\paragraph{Proposition A.3 (functional error transfer).}
If $h$ is $L_h$-Lipschitz in $\|\cdot\|_w$, then for any fitted probability laws with finite first moments,
\begin{align}
 |\widehat m_{a,h}(x)-\mu_{a,h}(x)|&\leq L_h\mathcal W_{1,w}(\widehat F_a(x),P_a(x)),\nonumber\\
 |\widehat\tau_h(x)-\tau_h(x)|&\leq L_h\sum_a\mathcal W_{1,w}(\widehat F_a(x),P_a(x)),\label{eq:app-transfer}\\
 |\E_X\widehat\tau_h(X)-\theta_h|&\leq L_h\sum_a\E_X\mathcal W_{1,w}(\widehat F_a(X),P_a(X)).\nonumber
\end{align}
The integrated inequality assumes integrability across $X$ of both true and fitted summaries; it is in particular finite whenever the right side is finite and the target has the integrability stated above. The stratum analogue divides the expectation restricted to $B$ by $p_B$. In particular $L_{h_\star}=1$, irrespective of the location of the fixed reference.

\paragraph{Proof.}
For any coupling $(Z,V)$ of the fitted and true laws, $|\E h(Z)-\E h(V)|\leq L_h\E\|Z-V\|_w$. Take the infimum over couplings, subtract arms, and integrate as needed. The reverse triangle inequality proves the reference claim. $\square$

The grid mean and grid standard deviation have Lipschitz constant at most one; for standard deviation this follows by viewing centering as an orthogonal projection in the weighted Euclidean space. Coordinate $j$ has constant $w_j^{-1/2}$. For a nonempty tail-index set $J$, the normalized weighted mean $\sum_{j\in J}w_jq_j/\sum_{j\in J}w_j$ has constant $(\sum_{j\in J}w_j)^{-1/2}$. Skewness is different: its denominator vanishes at constant vectors, and in general no global Lipschitz or even continuity claim holds there. A skewness analysis needs an explicit convention at degeneracy and its own moment/regularity conditions, or a restricted support with dispersion bounded away from zero. AIPW inference below only needs the stated scalar moment and nuisance conditions, rather than Lipschitzness of every chosen $h$.

Theorem A.2 therefore gives integrated conditional-effect consistency for Lipschitz summaries, including reference TCATE, and marginal and fixed-stratum consistency. It does not give convergence at every fixed $x$: integrated error can vanish while an error remains on a shrinking neighborhood of that point. Pointwise transport convergence gives pointwise functional convergence through \eqref{eq:app-transfer}; a uniform transport bound gives its uniform counterpart. To estimate the marginal plug-in effect using an independent evaluation sample of size $N$, average $\widehat\tau_h(X_i)$ over that sample. With bounded $h$ on $C$, its error is bounded by the population plug-in error plus $O_p(N^{-1/2})$. Foldwise evaluation gives the analogous consistency result. Root-$n$ normality of a plug-in effect does not follow from law consistency alone because its first-order regression bias remains.

The identification and conditional-target definitions specialize TCDA \citep[Theorem 4.1 and Definition 4.2]{souto_diamantis_tcda_2026}: use the outcome representation $h\circ q_K$, then estimate its conditional mean. Proposition A.3 gives the conditional-law version of its coupling stability argument \citep[Theorem 6.2 and Corollary 6.3]{souto_diamantis_tcda_2026}, with $h$ acting on the grid-vector outcome space. The finite-grid reference effect is also an outcome-level transformed contrast. It is not TCDA's nonlinear transformation of an already averaged interventional law. TCDA's silhouette-specific inference theorems concern another representation and cannot substitute for WCF law-learning or pointwise regression theory.

\subsection{Cross-fitted calibration: consistency, limiting distribution, and standard errors}

Use a fixed number $F\geq2$ of folds $I_1,\ldots,I_F$, assigned independently of the observations, with $|I_f|/n\to\rho_f\in(0,1)$. The fold count $F$ is unrelated to the quantile-grid size $K$. For $i\in I_f$, fit both nuisances outside $I_f$ and write $\widehat m_{a,h}^{(-f)}$ and $\widehat e^{(-f)}$ for their predictions. Any tuning used in the following simple cross-fitting argument is also performed outside $I_f$. Put
\begin{align}
 \widehat\phi_i(h)={}&\widehat m_{1,h}^{(-f)}(X_i)-\widehat m_{0,h}^{(-f)}(X_i)
 +\frac{A_i}{\widehat e^{(-f)}(X_i)}\{H_i-\widehat m_{1,h}^{(-f)}(X_i)\}\nonumber\\
 &-\frac{1-A_i}{1-\widehat e^{(-f)}(X_i)}\{H_i-\widehat m_{0,h}^{(-f)}(X_i)\},\qquad
 \widehat\theta_h=\frac1n\sum_i\widehat\phi_i(h).
 \label{eq:app-score}
\end{align}
For fixed candidate functions $(m_0,m_1,g)$, the corresponding score satisfies
\begin{equation}
 b_h(x):=\E\{\phi(h;m,g)\mid X=x\}-\tau_h(x)
 =(g-e)\left\{\frac{m_1-\mu_{1,h}}{g}+\frac{m_0-\mu_{0,h}}{1-g}\right\}(x).
 \label{eq:app-dr-bias}
\end{equation}
This follows by replacing $A H$ and $(1-A)H$ by their conditional means. It is the scalar specialization of TCDA's augmented remainder and product-rate bound \citep[Proposition 4.3]{souto_diamantis_tcda_2026}, and of the usual doubly robust identity \citep{bang2005doubly}. If fitted propensities lie in $[c,1-c]$, Cauchy--Schwarz gives
\[
 |\E_X b_h(X)|\leq c^{-1}\|g-e\|_2\sum_a\|m_a-\mu_{a,h}\|_2,
 \qquad \|v\|_2^2=\E_X v(X)^2.
\]
Thus cross-fitted consistency follows when this product tends to zero and the foldwise score second moments are $O_p(1)$. In particular, either nuisance side may be inconsistent if the other side is consistent and the remaining errors are bounded in $L^2$. Consistency under one correct side is weaker than the efficient central limit theorem below.

\paragraph{Theorem A.4 (scalar AIPW inference).}
Suppose the causal and i.i.d. assumptions hold, $c\leq e,\widehat e^{(-f)}\leq1-c$, $\E H^2<\infty$, and $\E[(H-\mu_{a,h}(X))^2\mid X,A=a]$ is uniformly bounded for both arms. For every fold suppose
\[
 r_{e,f}=\|\widehat e^{(-f)}-e\|_2=o_p(1),\qquad
 r_{m,f}=\sum_a\|\widehat m_{a,h}^{(-f)}-\mu_{a,h}\|_2=o_p(1),\qquad
 r_{e,f}r_{m,f}=o_p(n^{-1/2}).
\]
Let $\phi_0(O;h)$ be \eqref{eq:app-score} evaluated at $(\mu_{0,h},\mu_{1,h},e)$ and put $\psi_h(O)=\phi_0(O;h)-\theta_h$. If $0<V_h=\E\psi_h^2<\infty$, then
\begin{equation}
 \sqrt n(\widehat\theta_h-\theta_h)=n^{-1/2}\sum_i\psi_h(O_i)+o_p(1)
 \ \Longrightarrow\ N(0,V_h),\qquad
 \widehat V_h=\frac1n\sum_i\{\widehat\phi_i(h)-\widehat\theta_h\}^2\to_p V_h.
 \label{eq:app-clt}
\end{equation}
Accordingly $\sqrt{\widehat V_h/n}$ is a consistent standard error. The influence function is the efficient one in the nonparametric i.i.d. model for the fixed scalar target.

\paragraph{Proof.}
The score and cross-fitting argument are those of double/debiased machine learning for the ATE \citep[Theorem 5.1]{chernozhukov2018double}; we give a direct proof under the sufficient conditions stated here. Subtract the true score fold by fold. Equation~\eqref{eq:app-dr-bias} makes the conditional mean of this difference $o_p(n^{-1/2})$. To check its second moment, write $\Delta_f=\phi(h;\widehat m^{(-f)},\widehat e^{(-f)})-\phi_0(h)$. Expanding $\Delta_f$ separates bounded multiples of the two regression errors from terms of the form $A(H-\mu_{1,h})(1/\widehat e^{(-f)}-1/e)$ and its control-arm counterpart. The inverse-weight bounds and the conditional residual-variance bound therefore give
\[
 \E[\Delta_f^2\mid\text{training outside }I_f]
 \leq C\{r_{m,f}^2+r_{e,f}^2\}=o_p(1),
\]
where $C$ depends only on the overlap and residual-variance bounds. Conditional on that training data, the centered empirical difference on $I_f$ is $o_p(n^{-1/2})$ by Chebyshev's inequality and $|I_f|\asymp n$. Summing over the fixed number of folds does not require independence between their training samples. The ordinary i.i.d. central limit theorem applies to $\psi_h$. Conditional Markov's inequality also makes the empirical mean squared fitted-minus-true score difference $o_p(1)$; the law of large numbers for the squared true score and Cauchy--Schwarz then prove variance consistency. The efficient influence-function identification is the nonparametric ATE result in the cited theorem, applied to the observed scalar outcome $H=h(Q)$. $\square$

Under a law-transport rate $s_{a,f}=\|\mathcal W_{1,w}(\widehat F_a^{(-f)}(X),P_a(X))\|_2$, Proposition A.3 gives $r_{m,f}\leq L_h\sum_a s_{a,f}$. Hence $r_{e,f}\sum_a s_{a,f}=o_p(n^{-1/2})$ suffices for the product condition for Lipschitz $h$, alongside consistency of both nuisance sides. Theorem A.2 by itself supplies no such rate. Fixed-$M$ WCF can still yield a consistent calibrated scalar effect when the propensity is consistently estimated, but it need not satisfy the outcome-consistency condition of Theorem A.4. If the propensity is exactly known and the fitted outcome means converge in $L^2$ to fixed, possibly incorrect limits, the same proof instead uses $\phi(h;m^\dagger,e)-\theta_h$ as the influence function; normality and score-based standard errors remain available, generally without efficiency. Estimating the propensity while the outcome limit is incorrect does not inherit that conclusion without additional analysis.

For a fixed stratum, use $\widehat p_B=n^{-1}\sum_i1\{X_i\in B\}$ and
\[
 \widehat\theta_{h,B}=\frac{\sum_i1\{X_i\in B\}\widehat\phi_i(h)}{n\widehat p_B},\qquad
 \psi_{h,B}(O)=\frac{1\{X\in B\}}{p_B}\{\phi_0(O;h)-\theta_{h,B}\}.
\]
Under Theorem A.4's conditions and positive stratum variance, the ratio expansion gives root-$n$ normality with variance $\E\psi_{h,B}^2$. Its estimate is the sample mean of $[1\{X_i\in B\}\{\widehat\phi_i(h)-\widehat\theta_{h,B}\}/\widehat p_B]^2$. The same argument gives a joint finite-dimensional normal limit for finitely many prespecified summaries and strata, using their empirical influence-function covariance. It does not justify selecting favorable functionals after examining their estimated effects or simultaneous bands over all $x$.

For the archived implementation, outcome and propensity folds differ and the contrast penalty is selected using held-out risks across the outcome folds. These details are not literally the single-partition, training-only tuning scheme of Theorem A.4. One can establish an extension with suitable foldwise arguments and uniform control over the finite tuning candidates, or use common outer folds with tuning nested inside each training set; neither extension is certified here by calling predictions out of fold. Likewise, clipping at $[0.02,0.98]$ permits propensity consistency only if the true range is compatible with those bounds (or the clipping rule is changed). The theorem states sufficient conditions for valid calibration, rather than asserting that the experimental defaults meet every condition.

\subsection{Pointwise TCATE inference requires a second estimation step}

Averages within fixed strata have the preceding root-$n$ theory. A full function $x\mapsto\tau_h(x)$ requires conditional regression of the scores. Here is an explicit sufficient construction using local constant kernel regression \citep{nadaraya1964regression}, separate from the archived stratum estimator, which shows that this target also permits inference. The argument below establishes the effect of estimated scores under the stated local conditions.

\paragraph{Proposition A.5 (a local score estimator).}
Under the causal and i.i.d. assumptions above, fit the nuisances on an independent training sample and form scores $\widehat\phi_i$ on an i.i.d. evaluation sample of size $N$. Let $L:\mathbb R^d\to[0,\infty)$ be a measurable, bounded, compactly supported kernel with $\int L=1$ and $\int L^2>0$, and let $b=b_N>0$ be deterministic. At an interior point $x$, let
\[
 \widehat\tau_{h,b}(x)=\frac{\sum_{i=1}^N L((X_i-x)/b)\widehat\phi_i}{\sum_{i=1}^N L((X_i-x)/b)}.
\]
Assign any fixed value if the denominator is zero, an event whose probability tends to zero under the conditions below. Suppose $X$ has density $f_X$ continuous and positive at $x$, $\tau_h$ is locally H\"older of order $s\in(0,1]$, and $v_h(z)=\operatorname{Var}(\phi_0(O;h)\mid X=z)$ is continuous with $v_h(x)>0$. Suppose the conditional $(2+\delta)$ moments of $\phi_0$ are locally bounded for some $\delta>0$. Use clipped nuisance propensities bounded away from zero and one. Conditional on training, assume over the kernel neighborhood that
\[
 \sup_z\E[(\widehat\phi-\phi_0)^2\mid X=z,\text{training}]=o_p(1),\qquad
 \sup_z|\E[\widehat\phi-\phi_0\mid X=z,\text{training}]|=o_p((Nb^d)^{-1/2}).
\]
If $b\to0$, $Nb^d\to\infty$, and $\sqrt{Nb^d}\,b^s\to0$, then
\begin{equation}
 \sqrt{Nb^d}\{\widehat\tau_{h,b}(x)-\tau_h(x)\}
 \ \Longrightarrow\ N\left(0,\frac{v_h(x)\int L(u)^2du}{f_X(x)}\right).
 \label{eq:app-local-clt}
\end{equation}
For example, local uniform nuisance consistency, bounded conditional residual variances, and a local uniform product error $o_p((Nb^d)^{-1/2})$ imply the two score conditions through \eqref{eq:app-dr-bias}.

\paragraph{Proof.}
Write $L_i=L((X_i-x)/b)$, $D_N=(Nb^d)^{-1}\sum_iL_i$, $\xi_i=\phi_0(O_i;h)-\tau_h(X_i)$, and $\Delta_i=\widehat\phi_i-\phi_0(O_i;h)$. A change of variables gives $\E D_N\to f_X(x)$ and $\operatorname{Var}(D_N)=O((Nb^d)^{-1})$, so $D_N\to_p f_X(x)>0$. On the event $D_N>0$, the exact ratio decomposition is
\[
 \widehat\tau_{h,b}(x)-\tau_h(x)
 =D_N^{-1}\frac1{Nb^d}\sum_i L_i
 \{\xi_i+\tau_h(X_i)-\tau_h(x)+\Delta_i\}.
\]
Compact kernel support and the H\"older condition bound the conditional-mean contribution after division by $D_N$ by $O(b^s)$. Conditional on training, the $\Delta_i$ numerator has mean $o_p((Nb^d)^{-1/2})$ and centered variance $o_p((Nb^d)^{-1})$ by the two score conditions. The residual term has mean zero and
\[
 b^{-d}\E[L_i^2\xi_i^2]\longrightarrow
 f_X(x)v_h(x)\int L(u)^2\,du.
\]
Its Lyapunov ratio is $O((Nb^d)^{-\delta/2})$, using the local conditional $(2+\delta)$ moment bound, so the triangular-array central limit theorem applies. Undersmoothing removes the mean contribution, and Slutsky's theorem with $D_N\to_p f_X(x)$ gives \eqref{eq:app-local-clt}. Centering by $\tau_h(X_i)$ in this decomposition accounts for the random denominator; the raw numerator alone would contain an additional term involving $\tau_h(x)^2$. $\square$

One consistent variance construction for this proposition uses
\[
 \widehat f_b(x)=\frac{1}{Nb^d}\sum_iL((X_i-x)/b),\qquad
 \widehat v_b(x)=\frac{\sum_iL((X_i-x)/b)\{\widehat\phi_i-\widehat\tau_{h,b}(x)\}^2}{\sum_iL((X_i-x)/b)}.
\]
Under the same local moment and score conditions, $\widehat v_b(x)\to_p v_h(x)$: truncate the squared true residuals to obtain a kernel law of large numbers, use the $(2+\delta)$ moment bound to remove truncation, and use conditional Markov's inequality to show that the kernel average of $\Delta_i^2$ is $o_p(1)$. Weighted Cauchy--Schwarz controls the cross term, and $\widehat\tau_{h,b}(x)\to_p\tau_h(x)$ handles centering. Thus $\widehat v_b(x)\int L^2/\{Nb^d\widehat f_b(x)\}$ consistently estimates the asymptotic variance divided by $Nb^d$. This construction applies verbatim to the fixed-reference summary. It explains both the slower effective sample size $Nb^d$ and the extra local assumptions; global $L^2$ nuisance rates alone do not establish pointwise TCATE inference. A lower-dimensional moderator $V$ can replace $X$ in this second regression, provided the analogous density, smoothness, moment, and score conditions hold given $V$, while the nuisance models still adjust for the full $X$. No uniform confidence-band claim is made.

\subsection{Two-part laws and their causal calibration}

Let $Z=1\{Q\neq\mathbf0_K\}$, with $Z^a$ defined analogously. On the finite-grid observation space define $\pi_a(x)=\Prob(Z^a=1\mid X=x)$ and $P_a^+(x)=\Law(Q^a\mid X=x,Z^a=1)$ wherever $\pi_a(x)>0$. Then
\[
 P_a(x)=(1-\pi_a(x))\delta_{\mathbf0_K}+\pi_a(x)P_a^+(x),\qquad
 \mu_{a,h}(x)=h(\mathbf0_K)+\pi_a(x)\{\mu_{a,h}^+(x)-h(\mathbf0_K)\},
 \quad \mu_{a,h}^+=\int h\,dP_a^+.
\]
The two-part fit learns participation from the binary responses $Z_i$ and the nondegenerate particle law from units with $Z_i=1$, then assembles the probability mixture. In the archived implementation, an arm-specific logistic classifier estimates the zero-state probability (with a constant empirical-rate fallback for a single observed class), and the shared particle learner is fitted on the nondegenerate rows. The classifier defines numerical zeros by $\max_k|Q_{ik}|\leq10^{-12}$ and clips fitted zero-state probabilities to $[10^{-9},1-10^{-9}]$ before assembly. These small fixed tolerances are implementation conventions, not asymptotic guarantees for arbitrarily small participation probabilities. Both components must be refitted outside each evaluation fold for the proposed calibrated extension; the archived full-training classifiers are used for plug-in prediction. The $+$ notation does not mean every coordinate is positive. For $1-\widehat\pi_a$ to equal the total zero-atom mass of the fitted mixture, its fitted component must satisfy $\widehat P_a^+(\{\mathbf0_K\})=0$. Without this support restriction, the assembled zero mass is $1-\widehat\pi_a+\widehat\pi_a\widehat P_a^+(\{\mathbf0_K\})$; the classifier still estimates the declared structural-state probability. Training on nonzero vectors alone does not impose that restriction on particle predictions. If $\pi_a=0$, the component law is immaterial and may be assigned any convention. Learning that component precisely in a neighborhood nevertheless needs enough nondegenerate observations in that arm. The contrast $\mu_{1,h}^+-\mu_{0,h}^+$ conditions on different post-treatment populations; it is not an effect among always-participating units without additional principal-stratum assumptions.

The event $Q=\mathbf0_K$ is observable for the grid target, but is not logically equivalent to $Y=\delta_0$ for arbitrary distributions: a small tail beyond the grid can be missed. Interpreting it as an entirely degenerate underlying distribution requires that equivalence in the scientific model, or an independently observed structural-state indicator. Thresholding near-zero measurements instead defines a different state and must be declared. This qualification is separate from double robustness.

\paragraph{Proposition A.6 (mixture stability and scalar repair).}
If true and fitted component laws are supported on a set $C$ containing $\mathbf0_K$ with diameter $L_C<\infty$, then pointwise in $x$,
\begin{align}
 \mathcal W_{1,w}(\widehat P_a^{\rm 2p},P_a)
 &\leq L_C|\widehat\pi_a-\pi_a|+
 \pi_a\mathcal W_{1,w}(\widehat P_a^+,P_a^+),\label{eq:app-mixture}\\
 |\widehat m_{a,h}^{\rm 2p}-\mu_{a,h}|
 &\leq2B_h|\widehat\pi_a-\pi_a|+
 \pi_a|\widehat\mu_{a,h}^+-\mu_{a,h}^+|,\qquad \sup_{q\in C}|h(q)|\leq B_h.\nonumber
\end{align}
Here $\widehat m_{a,h}^{\rm 2p}=h(\mathbf0_K)+\widehat\pi_a(\widehat\mu_{a,h}^+-h(\mathbf0_K))$. Its cross-fitted version can replace $\widehat m_{a,h}$ in \eqref{eq:app-score}, with exactly the same double-robust identity and Theorem A.4 whenever that theorem's nuisance conditions hold.

\paragraph{Proof.}
Insert the intermediate mixture with mass $\pi_a$ and component $\widehat P_a^+$. Moving the unmatched mass between zero and this component costs at most $L_C|\widehat\pi_a-\pi_a|$; couple the two remaining positive components at common mass $\pi_a$. For the second bound expand the difference as $(\widehat\pi_a-\pi_a)(\widehat\mu_{a,h}^+-h(\mathbf0_K))+\pi_a(\widehat\mu_{a,h}^+-\mu_{a,h}^+)$. The score result follows because AIPW only requires a scalar outcome regression, regardless of its factorization. $\square$

Thus $\|\widehat m_{a,h}^{\rm 2p}-\mu_{a,h}\|_2$ is bounded by $2B_h\|\widehat\pi_a-\pi_a\|_2+\|\pi_a(\widehat\mu_{a,h}^+-\mu_{a,h}^+)\|_2$. These are the component errors to combine with the propensity rate in Theorem A.4. The law bound also shows that regions with vanishing participation need not have well-estimated component laws to have small assembled-law error. For structural-mass effects use the observed outcome $1-Z$ with nuisance $1-\widehat\pi_a$ in the same score. For reference effects, retain $h_\star(\mathbf0_K)=\|q_\star\|_w$, which is generally nonzero. No new causal identification assumption is needed merely to calibrate the assembled scalar mean. The reported two-part experiments remain plug-in experiments: these formulas establish an available calibrated extension, not evidence that it was implemented or that its coverage was tested.

\subsection{Grid resolution, learned benchmarks, and the scope of inference}

The preceding results concern exact finite-grid observations, a fixed summary and reference, and independent units. To relate them to an underlying-distribution target, suppose the underlying distributions have finite second moments and the summaries and approximation errors below are integrable. Let $\mathcal R_Kq$ be the step-quantile distribution assigning mass $w_k$ to $q_k$, and put $a_K(Y)=W_2(\mathcal R_Kq_K(Y),Y)$. If a full-distribution summary $h_\infty$ is $L$-Lipschitz in $W_2$ and $h_K(q)=h_\infty(\mathcal R_Kq)$, then
\[
 |\theta_{h_K,K}-\theta_{h_\infty}|\leq L\sum_{a=0}^1\E a_K(Y^a).
\]
For reference distance to an underlying benchmark $Y_\star$, the bound is $\sum_a\E a_K(Y^a)+2W_2(\mathcal R_Kq_\star^K,Y_\star)$. These follow by coupling each distribution to its reconstruction and applying the triangle inequality. Merely increasing $K$ does not establish these approximation rates for arbitrary grids and tails. Moreover, taking $K=K_n\to\infty$ changes the learner's dimension and requires its statistical conditions to hold along that sequence. Root-$n$ inference for a full-distribution target needs total representation bias $o(n^{-1/2})$, not just convergence of that bias to zero.

If the observed vector is an estimated $\widetilde Q$, Lipschitz summaries obey $|h(\widetilde Q)-h(Q)|\leq L_h\|\widetilde Q-Q\|_w$. Nonvanishing within-unit quantile error therefore changes the observed-outcome target; increasing the number of units does not remove it automatically. A joint measurement-error or growing-within-unit-sample analysis is needed to interpret the result as inference on latent distributions.

For a learned reference $\widehat q_\star$, the difference between population reference contrasts at $\widehat q_\star$ and a fixed $q_\star$ is at most $2\|\widehat q_\star-q_\star\|_w$. An independently learned reference can be conditioned on if the target is explicitly the effect relative to that random benchmark. For inference about a fixed limiting benchmark, $o_p(n^{-1/2})$ reference error suffices to ignore its estimation; at slower rates one must analyze the reference contribution and the remainder, which may vanish in special cases. Where $\Prob(Q^a=q_\star)=0$ for both arms, the derivative of the reference contrast in direction $v$ is
\[
 \E\frac{\langle q_\star-Q^1,v\rangle_w}{\|q_\star-Q^1\|_w}
 -\E\frac{\langle q_\star-Q^0,v\rangle_w}{\|q_\star-Q^0\|_w},\qquad
 \langle u,v\rangle_w=\sum_k w_ku_kv_k.
\]
The pointwise derivative of the norm is bounded in absolute value by $\|v\|_w$, so dominated convergence gives the displayed derivative and, in fixed dimension, a remainder $o(\|\widehat q_\star-q_\star\|_w)$. Combining it with a root-$n$ asymptotically linear reference estimate gives the reference contribution to an influence expansion. A complete estimator expansion also requires control of the score and nuisance changes when the estimated reference replaces the fixed one, and the joint covariance when the same data estimate both quantities. An atom at the reference can invalidate ordinary differentiability.

\section{Simulation Designs}\label{app:dgps}

All simulation covariates are independent $\operatorname{Unif}[-1,1]$. Given treatment $a$ and covariates $x$, the generic potential quantile vector has coordinates
\begin{equation}
 Q_k^a=m_a(x)+\xi_a+\exp\{s_a(x)+\eta_a\}\psi\{z_k;\gamma_a(x)\},
 \qquad z_k=\Phi^{-1}\{(k-1/2)/K\},\label{eq:dgp-app}
\end{equation}
where
\[
 \psi(z;\gamma)=z+\gamma\left\{\frac{z^2-1}{2}+\frac{z^3-3z}{6}\right\}.
\]
For $0\leq\gamma<1$, $\psi$ is increasing because its derivative is $1-\gamma+\gamma(z+1)^2/2$. Assignment is $A\mid X\sim\operatorname{Bernoulli}\{e(X)\}$. Write $\sigma(t)=(1+e^{-t})^{-1}$ and $C_{l,u}(t)=\min\{u,\max(l,t)\}$.

\subsection{Location, shape, and overlap family}

This family uses $d=6$, independent shocks $\xi_a\sim N(0,0.20^2)$ and $\eta_a\sim N(0,0.12^2)$, and
\begin{align*}
 b(x)&=0.55x_2+0.35x_3-0.45x_4+0.30x_5,\\
 s(x)&=0.16+0.08x_6-0.04x_2,\\
 r(x)&=0.55+0.10x_4-0.08x_2.
\end{align*}
The designs are
\begin{align*}
\mathrm{LS0}:\quad&m_a=b,\quad s_a=s,\quad \gamma_a=C_{.05,.85}(r),\\
\mathrm{LS1}:\quad&m_a=b+a(0.20+0.15x_4),\quad s_a=s,\\
&\gamma_a=C_{.05,.85}\{r-a(0.16+0.06x_4)\},\\
\mathrm{LS2}:\quad&m_a=b+a(0.10+0.05x_3),\quad s_a=s-0.12a,\\
&\gamma_a=C_{.05,.85}(r-0.08a),\\
\mathrm{LS3}:\quad&(m_a,s_a,\gamma_a)\text{ as in LS1}.
\end{align*}
LS0 and LS1 use $e(x)=C_{.10,.90}\{\sigma(1.4x_4-1.1x_5+0.4x_6)\}$. LS2 uses $C_{.05,.95}\{\sigma(1.8x_4-0.9x_5)\}$, and LS3 uses $C_{.01,.99}[\sigma\{3.5(x_4-0.9x_5)\}]$. LS0 is a placebo, LS1 changes location and shape, LS2 also changes scale, and LS3 combines LS1's outcome surfaces with weaker overlap. Conditional effects are reported along each design's active moderator: $X_4$ for LS1 and LS3, $X_3$ for LS2, and $X_1$ for the placebo LS0. LS1 and LS3 change both location and shape through $X_4$, while LS2 changes location through $X_3$ and applies a constant log-scale shift, so these strata test recovery of LS effect modification. LS0 has no treatment effect, so its $X_1$ criterion is a null-heterogeneity stability check. The reference is
\[
 q_{\star,k}^K=1.10+0.60\left[z_k+0.30
 \left\{\frac{z_k^2-1}{2}+\frac{z_k^3-3z_k}{6}\right\}\right].
\]

\subsection{Law separation and heterogeneous effects}

This family uses $d=5$, the same four moderator bins, $q_{\star,k}^K=z_k$, and
\[
 f(x)=0.6\sin(\pi x_1)+0.4x_2x_3,\qquad
 e_m(x)=C_{.10,.90}[\sigma\{0.8(x_1+0.5x_2)\}].
\]
The six consecutive paper labels replace the nonconsecutive legacy labels D0, D2, and D5 to D8 in the archived experiment files. S1 (legacy D0) sets $m_a=f+0.8ax_1$, $s_a=0.20x_4+0.25ax_2$, $\gamma_a=0$, and both shocks to zero. S2 (legacy D2) is a stochastic null with $m_a=f$, $s_a=0.20x_4$, $\gamma_a=0$, and shock standard deviations $(0.35,0.20)$.

S3 (legacy D5) has equal mean quantile curves but different laws. It sets $m_a=f$ and $\gamma_a=0$; the control arm uses $s_0=0.20x_4$ and shock standard deviations $(0.40,0.45)$, while the treated arm uses $s_1=0.20x_4+0.45^2/2$ and $(0.15,0)$. The log-scale adjustment equates expected scales. S4 (legacy D6) sets $m_a=f+0.7ax_1$, $s_a=0.20x_4$, $\gamma_a=0$, $\eta_a\sim N(0,0.15^2)$, and
\[
 \xi_a\sim\tfrac12N(-1.5,0.25^2)+\tfrac12N(1.5,0.25^2),
\]
which creates a multimodal across-unit law. S5 (legacy D7) changes shape through $m_a=f$, $s_a=0.20x_4$, $\gamma_a=0.20+0.10x_3+a(0.30+0.10x_1)$, and shock standard deviations $(0.35,0.20)$. S1, S4, and S5 have effects that vary with $X_1$, while S3's spread change is carried by $X_4$. S1 to S5 use $e_m$.

S6 (legacy D8) combines a weak heterogeneous effect with stronger confounding:
\begin{align*}
 m_a(x)&=1.2\sin(\pi x_1)+0.9x_2x_3+a(0.12x_1+0.06),\\
 s_a(x)&=0.20x_4+0.05ax_2,\qquad\gamma_a(x)=0,\\
 e(x)&=C_{.05,.95}[\sigma\{2.5(x_1+0.7x_2-0.5x_3)\}],
\end{align*}
with shock standard deviations $(0.35,0.20)$. Its effects vary with $X_1$ and $X_2$.

\subsection{Structural-zero family}

The Z family uses $d=6$. With probability $1-\pi_a(x)$ the unit outcome is $\mathbf0_K$; otherwise equation~\eqref{eq:dgp-app} holds with
\begin{align*}
 m_a^+(x)&=4.80+0.35x_2-0.25x_4+a(\Delta+\beta x_2),\\
 s_a^+(x)&=0.10+0.06x_6-0.03x_2,\\
 \gamma_a^+(x)&=C_{.05,.85}(0.50+0.06x_4-a\kappa),
\end{align*}
and shock standard deviations $(0.15,0.10)$. Z0 and Z1 use $(\Delta,\beta,\kappa)=(0,0,0)$, whereas Z2 and Z3 use $(0.28,0.08,0.10)$. Z0 is a placebo and Z2 changes only the nondegenerate component; both use $\pi_a(x)=C_{.05,.95}\{\sigma(1.2x_4-0.8x_2)\}$. Z1 changes participation only, adding $a(0.90+0.20x_4)$ to this index. Z0 to Z2 use LS1's assignment mechanism. Z3 changes both components and uses
\[
 \pi_a(x)=C_{.01,.99}[\sigma\{3(x_4-0.9x_5)+a(0.90+0.60x_4)\}],
 \qquad e(x)=\pi_0(x).
\]
The reference is the same as for the LS family. The reported strata use $X_4$ for Z1 and Z3, where the participation contrast varies, and $X_2$ for Z2, where the positive-component effect varies; Z0 retains $X_1$ as a stability check. The superscript $+$ labels the nondegenerate branch, not strictly positive scalar support.

\section{Complete Simulation Results}\label{app:simulation-results}

Tables~\ref{tab:full_results_ls} to~\ref{tab:full_results_z} report the complete design families. Law is the mean MMD$^2$ criterion defined in Section~\ref{sec:sim-design}. TATE and TCATE are Monte Carlo RMSEs pooled over the grid mean, grid standard deviation, grid skewness, and upper-half mean; TCATE additionally pools four strata of the design's active moderator (Appendix~\ref{app:dgps}). Because the four functionals use different units, their aggregate is a compact descriptive summary on the stated simulation scale. Ref. TATE and Ref. TCATE are Monte Carlo RMSEs for the reference-distance functional.

\input{tables/full_results_ls.tex}
\input{tables/full_results_s.tex}
\input{tables/full_results_z.tex}

\subsection{Reconciling law and functional accuracy}

\begin{figure}[p]
\centering
\includegraphics[width=0.98\linewidth]{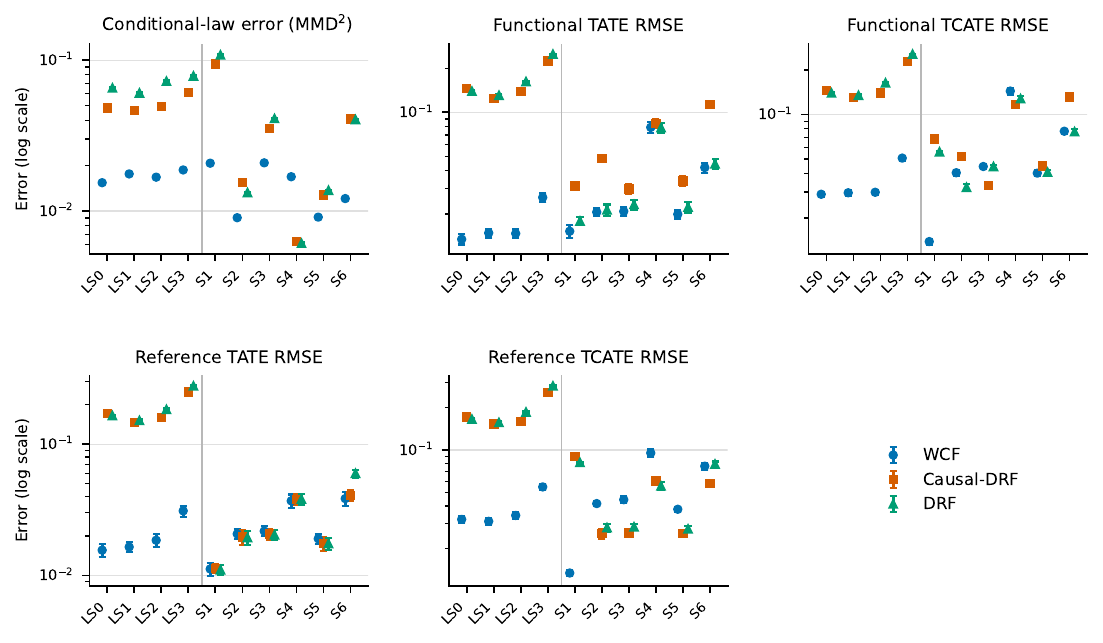}
\caption{All ordinary designs at $n=1000$ and fifty paired Monte Carlo replications. Bars are one Monte Carlo SE for each method separately. The vertical divider separates LS from S designs. Nonoverlap makes several LS and law-level rankings clear, and the narrower intervals sharpen most comparisons, but overlap remains for several S functional criteria. Because the displayed bars are not SEs of paired method differences, they should not be used as a formal paired test.}
\label{fig:all-dgp-summary}
\end{figure}

\begin{figure}[p]
\centering
\includegraphics[width=0.94\linewidth]{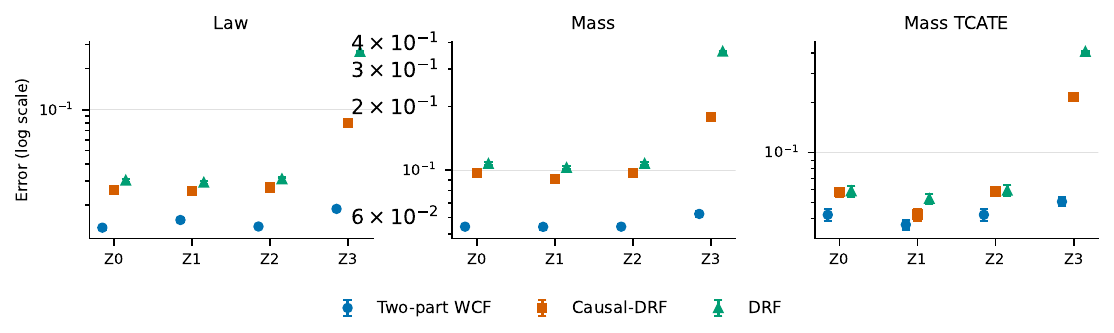}
\caption{All structural-zero designs at $n=1000$ and fifty paired Monte Carlo replications. Bars are one method-specific Monte Carlo SE. Two-part WCF is clearly separated on law and mass error; among the Z0 to Z2 mass-TCATE comparisons, Z1 remains within paired uncertainty.}
\label{fig:all-dgp-zero-summary}
\end{figure}

Figures~\ref{fig:all-dgp-summary} and~\ref{fig:all-dgp-zero-summary} display every $n=1000$ design. The apparent reversal after LS is not a general failure to estimate heterogeneous effects. S1 has strong effect modification through $X_1$ and favors WCF on the law, TATE, TCATE, and reference-TCATE criteria; its reference-TATE comparison is within paired uncertainty. In S5, WCF has the lowest law error and is best or essentially tied on the four-functional TATE and TCATE aggregates, but is less accurate for reference TCATE and statistically tied on reference TATE. S2 is a stochastic null, S3 changes the outer-law spread while preserving the mean quantile curve, S4 deliberately violates the small-particle model through separated multimodality, and S6 combines weak effects with strong confounding. These designs probe distinct failure mechanisms rather than forming a monotone scale of heterogeneity.

The law and reference criteria also emphasize different discrepancies. Gaussian-kernel MMD$^2$ is a global comparison of probability laws, whereas $h_{\star,K}(q)=d_{W,K}(q,q_\star^K)$ is a nonlinear, distance-sensitive functional. Lower MMD$^2$ does not provide a finite-sample ordering for its expectation. With $M=10$, particle discretization can be especially consequential for outer spread or multimodality, while shared partitions and arm-contrast shrinkage can favor smooth common structure but attenuate a weak, complex contrast. The current evidence does not separate these mechanisms because particle-budget sensitivity was conducted only in the LS family.

There is a second, procedural distinction. For WCF, the reported functional predictions are cross-fitted AIPW marginal or stratum estimates constructed from out-of-fold particle means. DRF and Causal-DRF are evaluated by applying the same declared functionals directly to their fitted conditional laws and averaging the resulting plug-in contrasts. Both procedures target the stated marginal or stratum estimands, but their finite-sample errors combine different law-estimation and calibration components. Consequently, the scalar columns compare the implemented end-to-end procedures; they do not identify whether a ranking arises from the law learner, nonlinear functional extraction, or AIPW variance. The fifty-replication run behind these tables measures this distinction directly: every fitted law is evaluated both by a common plug-in transformation and by a common cross-fitted AIPW layer, and the tables retain the end-to-end convention for continuity with the sensitivity archive.

The LS family is intended as a regular-regime benchmark with smooth Gaussian location, scale, and shape variation. The S family is a mechanism-based stress-test suite: S3 represents latent spread changes, S4 tests separated subpopulations, S5 tests heterogeneous shape, and S6 tests weak signal under confounding. Such mechanisms can occur in applications, but the parameter values are not calibrated claims about their empirical frequency. Accordingly, the paper's central empirical claim concerns conditional-law accuracy across both regular and stress regimes, not uniform superiority for every scalar treatment-effect summary.

Fifty paired replications sharpen every interval in Figures~\ref{fig:all-dgp-summary} and~\ref{fig:all-dgp-zero-summary} and confirm the largest separations, including the LS law and functional advantages, WCF's law advantage in most S designs, S4's multimodal law reversal, and the two-part law and mass improvements. Paired 95\% intervals resolve several formerly close rankings, including S1's law, TATE, TCATE, and reference-TCATE advantages and the Z0 and Z2 mass-TCATE comparisons, while the S1, S3, S4, and S5 reference-TATE comparisons and the Z1 mass-TCATE comparison remain within paired uncertainty and are described as ties. The S3 and S5 reference-TCATE deficits for WCF remain significant. Marginal one-SE bars in the figures do not exploit the paired seeds.

\clearpage

\section{Sensitivity Analyses}\label{app:sensitivity}

\subsection{Quantile-grid and particle resolution}

The sensitivity study varies $K\in\{5,25,49\}$ at $M=10$ and $M\in\{5,10,25\}$ at $K=25$ in LS0 to LS3 at $n=1000$. Table~\ref{tab:sensitivity-km} reports the native-grid law error, the mean-quantile error, and the end-to-end reference-TCATE and reference-TATE errors. Moving from $K=25$ to $K=49$ at $M=10$ changes native law error by 0.8 to 2.8 percent and the end-to-end reference-TCATE error by at most 3.6 percent; the paired standard errors in Table~\ref{tab:sensitivity-resolution} are of the same order, so none of these differences is statistically resolved. Increasing $M$ from 10 to 25 at $K=25$ changes native law error by at most 3.3 percent, and the largest paired change in end-to-end reference-TCATE error is a 3.8 percent reduction in LS3 with a paired standard error of 3.1 percentage points. The factorial grid in Table~\ref{tab:sensitivity-factorial} varies by at most about 7 percent across the nine $(K,M)$ combinations for LS1 and LS3, with the largest gaps at $K=5$. The study therefore does not identify a resolution that is uniformly better on the native protocol, and the primary specification $(K,M)=(25,10)$ is retained. Because the native target changes with $K$, these comparisons mix quadrature resolution with tail coverage and should not be read as a pure fit-quality ranking.

\input{tables/wcf_sensitivity_km.tex}

\input{tables/wcf_sensitivity_factorial.tex}

\input{tables/wcf_sensitivity_resolution.tex}

\begin{figure}[!ht]
\centering
\includegraphics[width=0.96\linewidth]{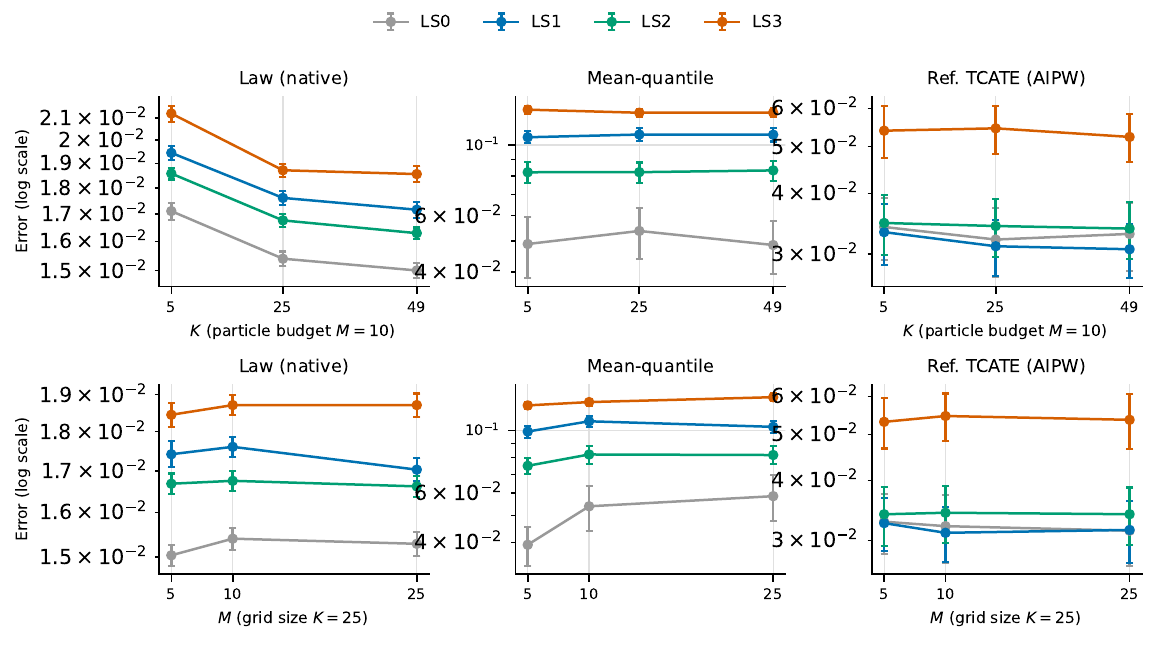}
\caption{Quantile-grid and particle-budget sensitivity of WCF at $n=1000$, one line per design (LS0 to LS3). The top row varies $K$ at $M=10$ and the bottom row varies $M$ at $K=25$. Panels report native-grid law error (Monte Carlo means), mean-quantile error, and end-to-end reference-TCATE error (Monte Carlo RMSEs). Because the native target changes with $K$, the top row is a stability check rather than a ranking of grid resolutions. Bars are Monte Carlo standard errors over fifty paired replications; lower is better.}
\label{fig:sensitivity-resolution}
\end{figure}

\subsection{Assignment and propensity estimation}

A symmetric family holds the potential-outcome law fixed while comparing constant, linear, nonlinear, and prognosis-dependent propensity functions. At $n=1000$, WCF's end-to-end reference-TCATE errors are $0.0306$, $0.0391$, $0.0420$, and $0.0361$ under constant, linear, nonlinear, and prognosis-dependent assignment. The corresponding Causal-DRF values are $0.0354$, $0.0406$, $0.0813$, and $0.0506$, and DRF records $0.0329$, $0.0364$, $0.0725$, and $0.0459$. WCF has the lowest error in all four mechanisms at $n=500$ and in three of four at $n=1000$, where DRF is slightly more accurate under linear assignment. The aligned and irrelevant assignment pair differs by $+0.0034$ (SE $0.0022$) in end-to-end reference-TCATE error at $n=1000$, a small gap that is not decisive at conventional levels.

\begin{figure}[!ht]
\centering
\includegraphics[width=0.92\linewidth]{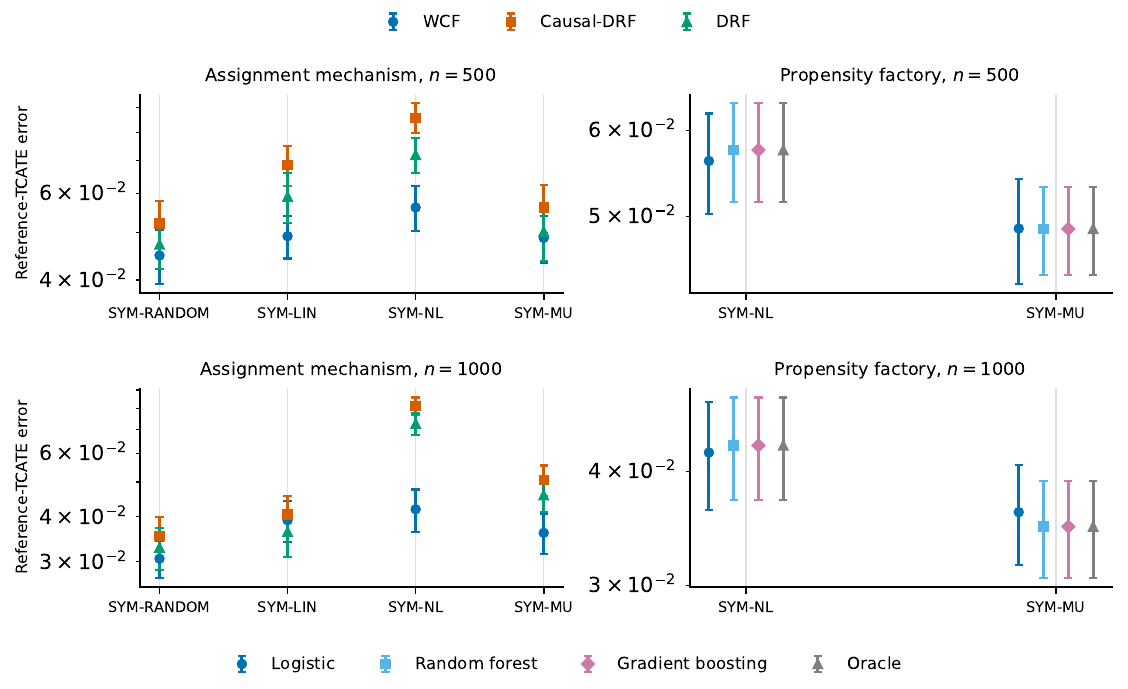}
\caption{Assignment and propensity sensitivity of the end-to-end reference-TCATE error at $n=500$ and $n=1000$. Left column: constant (SYM-RANDOM), linear (SYM-LIN), and nonlinear (SYM-NL, SYM-MU) assignment mechanisms under the default logistic propensity. Right column: the SYM-NL and SYM-MU designs under the default logistic propensity and under random-forest, gradient-boosting, and oracle propensity factories; the factories nearly coincide because the confirmatory AIPW layer is fixed across columns. Bars are Monte Carlo standard errors over fifty paired replications; lower is better.}
\label{fig:sensitivity-assignment}
\end{figure}

Under the confirmatory protocol the tabulated errors are produced by a common cross-fitted AIPW layer whose propensity model is fixed across columns, so replacing logistic regression with a random-forest, gradient-boosting, or oracle factory changes the end-to-end reference-TCATE errors by only a few percent (Table~\ref{tab:sensitivity-propensity}). The factory's effect appears instead in the estimator's internal calibration diagnostic: the influence-function standard error of the reference estimate is $0.079$ for the gradient-boosting factory against $0.021$ to $0.022$ for the alternatives, confirming the archive's diagnosis of one poorly calibrated factory while showing that the common calibration layer is insensitive to it. This comparison diagnoses one factory; it does not imply that flexible propensity estimation is intrinsically harmful or prove double robustness at fixed budgets. Figure~\ref{fig:sensitivity-assignment} displays both comparisons.

\input{tables/wcf_sensitivity_assignment.tex}

\input{tables/wcf_sensitivity_propensity.tex}

\subsection{Null companions}

\begin{figure}[!ht]
\centering
\includegraphics[width=0.92\linewidth]{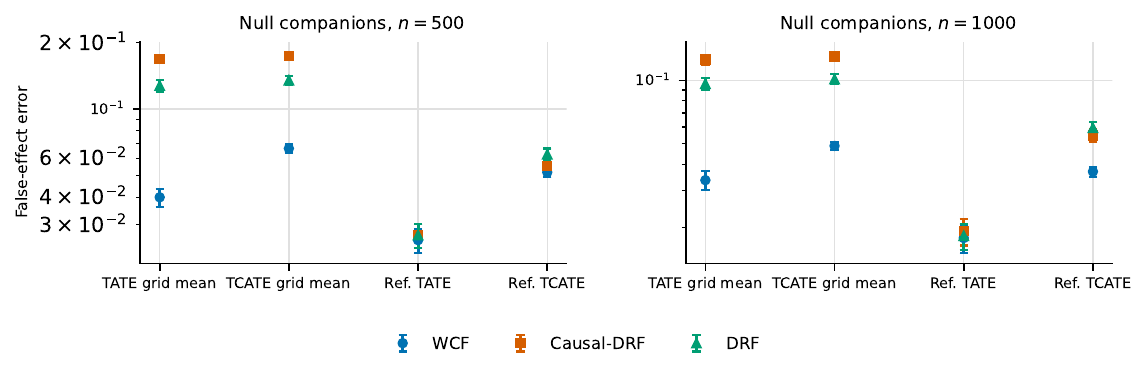}
\caption{False-effect error of the six null companion designs at $n=500$ and $n=1000$, pooled over designs and fifty paired replications. Entries are end-to-end Monte Carlo RMSEs; every treatment effect is exactly zero, so each value is a false-effect error. Bars are Monte Carlo standard errors; lower is better.}
\label{fig:sensitivity-nulls}
\end{figure}

Six companion designs set every treatment effect exactly to zero while retaining the corresponding prognostic and assignment structures. Table~\ref{tab:sensitivity-placebo} reports pooled false-effect error at $n=500$ and $n=1000$. WCF has the lowest error on both grid-mean functionals by a wide margin, and at $n=1000$ its pooled reference-TCATE error is $0.0369$ against $0.0546$ for Causal-DRF and $0.0595$ for DRF; the reference-TATE errors are close across methods. The comparison checks whether apparent functional heterogeneity is induced by regularization or assignment rather than by a true treatment contrast. Figure~\ref{fig:sensitivity-nulls} displays the pooled false-effect errors.

\input{tables/wcf_sensitivity_placebo.tex}

\subsection{Sample-size sensitivity}

\begin{figure}[!ht]
\centering
\includegraphics[width=0.96\linewidth]{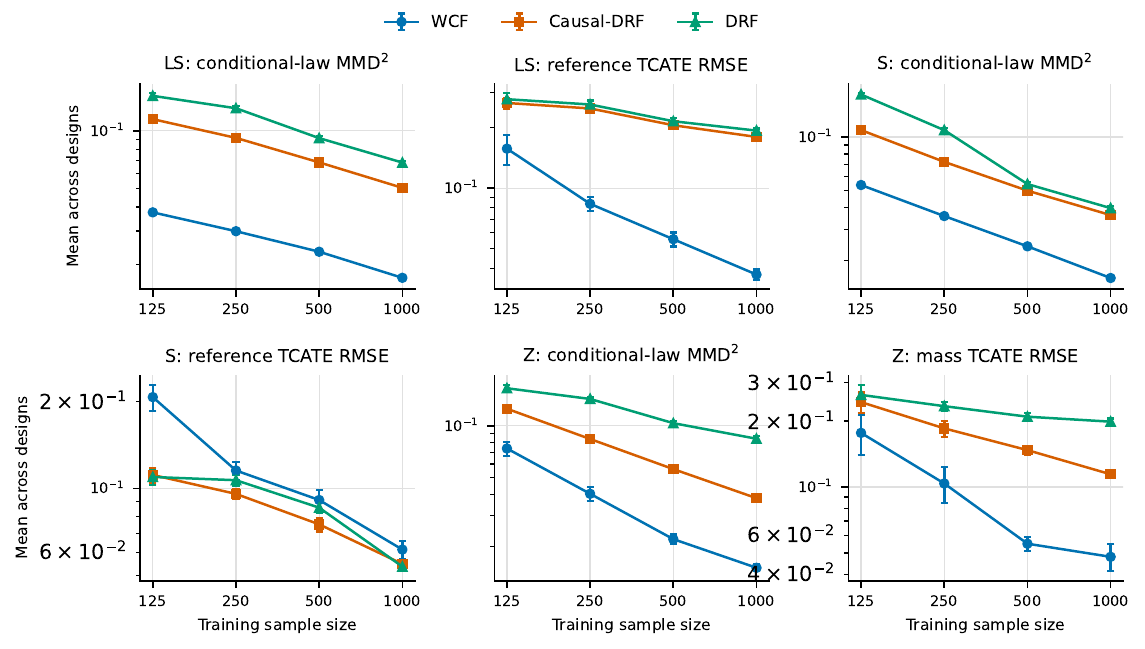}
\caption{Sample-size sensitivity averaged over designs within each family. Law panels report mean MMD$^2$; reference and mass-TCATE panels report Monte Carlo RMSE. The LS and S panels use ordinary WCF, whereas Z uses two-part WCF. Lines are descriptive family averages and bars are Monte Carlo standard errors over the ten paired replications; complete design-level results appear in Tables~\ref{tab:sample-size-ls} to~\ref{tab:sample-size-z}.}
\label{fig:sample-size-sensitivity}
\end{figure}

The sample-size study crosses all 14 designs with $n\in\{125,250,500,1000\}$, the three family-appropriate methods, and ten paired replications, for 1,680 completed cells. Ordinary WCF is used for LS0 to LS3 and S1 to S6; two-part WCF is used for Z0 to Z3. Every cell retains $K=25$, $M=10$, and an independent test sample of 1000 units. Figure~\ref{fig:sample-size-sensitivity} averages the errors over designs within each family to display the principal trajectories. Tables~\ref{tab:sample-size-ls} to~\ref{tab:sample-size-z} report every design, sample size, and method.

WCF's family-average law error is lowest at every sample size in all three families. It also has the lowest individual-design law error in all 16 LS comparisons, 20 of 24 S comparisons, and all 16 Z comparisons. The gain is not invariant to the target. Across S1 to S6, WCF's average reference-TCATE RMSE is $0.1769$ at $n=125$ and $0.0534$ at $n=1000$, compared with $0.0965$ and $0.0482$ for DRF. WCF leads on this target only for S1; the largest small-sample gaps occur in the multimodal S4 and weak-effect, strong-confounding S6 designs. Thus a small particle law can estimate a broad law discrepancy well while remaining inefficient for a particular convex functional.

For structural mass, the two-part model is most valuable when both participation and the positive component change. In Z3 it has the lowest law, mass, and mass-TCATE error at every $n$. In Z0 to Z2, Causal-DRF is slightly better for several mass or reference contrasts at $n=125$ and $n=250$, although two-part WCF regains the lead in most comparisons by $n=500$. This is consistent with the extra first-stage classification burden being material when only about 125 to 250 unit-level distributions are observed.

\input{tables/sample_size_ls.tex}

\input{tables/sample_size_s.tex}

\input{tables/sample_size_z.tex}

\section{Applied-Study Details}\label{app:applications}

\input{applied_star_appendix.tex}

\input{applied_minwage_appendix.tex}
\input{applied_cash_appendix.tex}

%% file: tables/full_results_ls.tex
\begin{table*}[t]
\centering
\scriptsize
\setlength{\tabcolsep}{4pt}
\caption{Location, shape, and overlap designs at $n=1000$. Law entries are Monte Carlo means; functional-effect entries are Monte Carlo RMSEs. Parentheses give Monte Carlo SEs over fifty replications; lower is better.}
\label{tab:full_results_ls}
\begin{tabular}{llrrrrr}
\toprule
Design & Method & Law & TATE & TCATE & Ref. TATE & Ref. TCATE \\
\midrule
LS0 & WCF & 0.0154(0.0002) & 0.0134(0.0011) & 0.0289(0.0012) & 0.0156(0.0017) & 0.0321(0.0020) \\
LS0 & Causal-DRF & 0.0483(0.0006) & 0.1451(0.0027) & 0.1452(0.0027) & 0.1714(0.0036) & 0.1715(0.0035) \\
LS0 & DRF & 0.0654(0.0007) & 0.1396(0.0024) & 0.1396(0.0024) & 0.1653(0.0031) & 0.1653(0.0031) \\
\addlinespace
LS1 & WCF & 0.0176(0.0003) & 0.0148(0.0010) & 0.0296(0.0012) & 0.0165(0.0014) & 0.0311(0.0016) \\
LS1 & Causal-DRF & 0.0466(0.0006) & 0.1250(0.0028) & 0.1302(0.0028) & 0.1457(0.0036) & 0.1526(0.0037) \\
LS1 & DRF & 0.0605(0.0007) & 0.1308(0.0022) & 0.1347(0.0023) & 0.1515(0.0029) & 0.1566(0.0030) \\
\addlinespace
LS2 & WCF & 0.0168(0.0002) & 0.0147(0.0011) & 0.0299(0.0011) & 0.0186(0.0020) & 0.0343(0.0018) \\
LS2 & Causal-DRF & 0.0494(0.0006) & 0.1394(0.0027) & 0.1401(0.0027) & 0.1591(0.0036) & 0.1598(0.0036) \\
LS2 & DRF & 0.0725(0.0008) & 0.1626(0.0026) & 0.1636(0.0026) & 0.1843(0.0032) & 0.1855(0.0033) \\
\addlinespace
LS3 & WCF & 0.0187(0.0003) & 0.0260(0.0019) & 0.0508(0.0019) & 0.0311(0.0030) & 0.0546(0.0029) \\
LS3 & Causal-DRF & 0.0608(0.0007) & 0.2248(0.0026) & 0.2277(0.0025) & 0.2502(0.0032) & 0.2551(0.0032) \\
LS3 & DRF & 0.0785(0.0008) & 0.2507(0.0020) & 0.2553(0.0020) & 0.2782(0.0026) & 0.2846(0.0026) \\
\addlinespace
\bottomrule
\end{tabular}
\end{table*}

%% file: tables/full_results_s.tex
\begin{table*}[t]
\centering
\scriptsize
\setlength{\tabcolsep}{4pt}
\caption{Law-separation and heterogeneous-effect designs at $n=1000$. Law entries are Monte Carlo means; functional-effect entries are Monte Carlo RMSEs. Parentheses give Monte Carlo SEs over fifty replications; lower is better.}
\label{tab:full_results_s}
\begin{tabular}{llrrrrr}
\toprule
Design & Method & Law & TATE & TCATE & Ref. TATE & Ref. TCATE \\
\midrule
S1 & WCF & 0.0207(0.0006) & 0.0152(0.0015) & 0.0139(0.0006) & 0.0112(0.0013) & 0.0134(0.0007) \\
S1 & Causal-DRF & 0.0942(0.0020) & 0.0312(0.0013) & 0.0683(0.0009) & 0.0113(0.0009) & 0.0895(0.0011) \\
S1 & DRF & 0.1084(0.0020) & 0.0179(0.0012) & 0.0560(0.0011) & 0.0110(0.0010) & 0.0814(0.0012) \\
\addlinespace
S2 & WCF & 0.0090(0.0001) & 0.0206(0.0013) & 0.0405(0.0017) & 0.0207(0.0019) & 0.0416(0.0017) \\
S2 & Causal-DRF & 0.0154(0.0002) & 0.0483(0.0028) & 0.0520(0.0026) & 0.0197(0.0025) & 0.0254(0.0020) \\
S2 & DRF & 0.0132(0.0001) & 0.0214(0.0019) & 0.0322(0.0017) & 0.0194(0.0023) & 0.0281(0.0018) \\
\addlinespace
S3 & WCF & 0.0209(0.0003) & 0.0209(0.0015) & 0.0446(0.0017) & 0.0218(0.0019) & 0.0444(0.0025) \\
S3 & Causal-DRF & 0.0354(0.0003) & 0.0298(0.0022) & 0.0332(0.0020) & 0.0207(0.0020) & 0.0258(0.0018) \\
S3 & DRF & 0.0410(0.0004) & 0.0233(0.0018) & 0.0445(0.0011) & 0.0204(0.0018) & 0.0284(0.0014) \\
\addlinespace
S4 & WCF & 0.0169(0.0003) & 0.0788(0.0070) & 0.1439(0.0074) & 0.0370(0.0044) & 0.0951(0.0059) \\
S4 & Causal-DRF & 0.0063(0.0001) & 0.0841(0.0064) & 0.1168(0.0056) & 0.0380(0.0038) & 0.0601(0.0040) \\
S4 & DRF & 0.0061(0.0001) & 0.0784(0.0064) & 0.1280(0.0047) & 0.0381(0.0038) & 0.0554(0.0038) \\
\addlinespace
S5 & WCF & 0.0091(0.0001) & 0.0200(0.0013) & 0.0402(0.0019) & 0.0191(0.0017) & 0.0379(0.0016) \\
S5 & Causal-DRF & 0.0128(0.0001) & 0.0338(0.0025) & 0.0449(0.0020) & 0.0174(0.0020) & 0.0255(0.0014) \\
S5 & DRF & 0.0137(0.0001) & 0.0223(0.0018) & 0.0408(0.0014) & 0.0175(0.0019) & 0.0274(0.0014) \\
\addlinespace
S6 & WCF & 0.0121(0.0002) & 0.0418(0.0034) & 0.0771(0.0032) & 0.0385(0.0046) & 0.0767(0.0041) \\
S6 & Causal-DRF & 0.0408(0.0005) & 0.1135(0.0055) & 0.1317(0.0045) & 0.0408(0.0039) & 0.0580(0.0028) \\
S6 & DRF & 0.0403(0.0004) & 0.0442(0.0035) & 0.0765(0.0033) & 0.0599(0.0042) & 0.0793(0.0035) \\
\addlinespace
\bottomrule
\end{tabular}
\end{table*}

%% file: tables/full_results_z.tex
\begin{table*}[t]
\centering
\scriptsize
\caption{Structural-zero designs at $n=1000$. Mass is the error in the degenerate-component probability, and mass TCATE is its moderator-stratum contrast error. Entries are Monte Carlo mean (SE) over fifty replications; lower is better.}
\label{tab:full_results_z}
\begin{tabular}{llrrr}
\toprule
Design & Method & Law & Mass & Mass TCATE \\
\midrule
Z0 & Two-part WCF & 0.0136(0.0004) & 0.0542(0.0016) & 0.0421(0.0033) \\
Z0 & Causal-DRF & 0.0258(0.0007) & 0.0969(0.0015) & 0.0577(0.0037) \\
Z0 & DRF & 0.0303(0.0008) & 0.1072(0.0018) & 0.0585(0.0043) \\
\addlinespace
Z1 & Two-part WCF & 0.0155(0.0004) & 0.0541(0.0015) & 0.0366(0.0024) \\
Z1 & Causal-DRF & 0.0254(0.0006) & 0.0910(0.0013) & 0.0422(0.0037) \\
Z1 & DRF & 0.0292(0.0007) & 0.1026(0.0015) & 0.0526(0.0038) \\
\addlinespace
Z2 & Two-part WCF & 0.0139(0.0004) & 0.0542(0.0016) & 0.0421(0.0033) \\
Z2 & Causal-DRF & 0.0269(0.0007) & 0.0970(0.0015) & 0.0580(0.0037) \\
Z2 & DRF & 0.0310(0.0008) & 0.1074(0.0018) & 0.0590(0.0043) \\
\addlinespace
Z3 & Two-part WCF & 0.0187(0.0005) & 0.0622(0.0017) & 0.0505(0.0030) \\
Z3 & Causal-DRF & 0.0796(0.0014) & 0.1782(0.0021) & 0.2163(0.0039) \\
Z3 & DRF & 0.2671(0.0044) & 0.3624(0.0031) & 0.4076(0.0057) \\
\addlinespace
\bottomrule
\end{tabular}
\end{table*}

%% file: tables/wcf_sensitivity_km.tex
\begin{table}[!ht]
\centering
\scriptsize
\setlength{\tabcolsep}{3pt}
\caption{Sensitivity of WCF to the quantile grid size $K$ and the particle budget $M$ at $n=1000$ for designs LS0 to LS3 under the confirmatory protocol. Panel A varies $K$ at $M=10$; Panel B varies $M$ at $K=25$. Law entries are Monte Carlo means and functional entries are Monte Carlo RMSEs over fifty paired replications, with Monte Carlo standard errors in parentheses; lower is better. The native-grid target changes with $K$, so Panel A is a stability check rather than a ranking of grid resolutions. WCF's functional columns use its cross-fitted AIPW layer; the plug-in column is shown for comparison.}
\label{tab:sensitivity-km}
\begin{tabular}{llrrrrr}
\toprule
\multicolumn{7}{l}{\textit{Panel A: $K$ sensitivity at $M=10$}} \\
DGP & $K$ & Law & Mean-quantile & Ref. TCATE plug-in & Ref. TCATE AIPW & Ref. TATE AIPW \\
\midrule
LS0 & 5 & 0.0171(0.0003) & 0.0489(0.0106) & 0.0197(0.0058) & 0.0341(0.0050) & 0.0158(0.0056) \\
LS0 & 25 & 0.0154(0.0002) & 0.0538(0.0098) & 0.0205(0.0062) & 0.0321(0.0052) & 0.0156(0.0062) \\
LS0 & 49 & 0.0150(0.0002) & 0.0486(0.0091) & 0.0194(0.0058) & 0.0330(0.0053) & 0.0158(0.0056) \\
LS1 & 5 & 0.0194(0.0003) & 0.1059(0.0046) & 0.0655(0.0052) & 0.0333(0.0048) & 0.0176(0.0060) \\
LS1 & 25 & 0.0176(0.0003) & 0.1079(0.0051) & 0.0648(0.0062) & 0.0311(0.0041) & 0.0165(0.0050) \\
LS1 & 49 & 0.0172(0.0003) & 0.1078(0.0052) & 0.0636(0.0054) & 0.0307(0.0039) & 0.0158(0.0054) \\
LS2 & 5 & 0.0186(0.0002) & 0.0822(0.0061) & 0.0538(0.0056) & 0.0348(0.0049) & 0.0186(0.0067) \\
LS2 & 25 & 0.0168(0.0002) & 0.0822(0.0059) & 0.0520(0.0057) & 0.0343(0.0047) & 0.0186(0.0070) \\
LS2 & 49 & 0.0163(0.0002) & 0.0833(0.0059) & 0.0501(0.0058) & 0.0338(0.0046) & 0.0198(0.0063) \\
LS3 & 5 & 0.0212(0.0004) & 0.1292(0.0039) & 0.0980(0.0042) & 0.0540(0.0066) & 0.0308(0.0088) \\
LS3 & 25 & 0.0187(0.0003) & 0.1264(0.0037) & 0.0920(0.0037) & 0.0546(0.0063) & 0.0311(0.0088) \\
LS3 & 49 & 0.0186(0.0003) & 0.1265(0.0041) & 0.0943(0.0044) & 0.0524(0.0059) & 0.0302(0.0085) \\
\midrule
\multicolumn{7}{l}{\textit{Panel B: $M$ sensitivity at $K=25$}} \\
DGP & $M$ & Law & Mean-quantile & Ref. TCATE plug-in & Ref. TCATE AIPW & Ref. TATE AIPW \\
\midrule
LS0 & 5 & 0.0150(0.0002) & 0.0392(0.0062) & 0.0182(0.0060) & 0.0328(0.0047) & 0.0160(0.0058) \\
LS0 & 10 & 0.0154(0.0002) & 0.0538(0.0098) & 0.0205(0.0062) & 0.0321(0.0052) & 0.0156(0.0062) \\
LS0 & 25 & 0.0153(0.0003) & 0.0584(0.0109) & 0.0200(0.0059) & 0.0314(0.0049) & 0.0143(0.0043) \\
LS1 & 5 & 0.0174(0.0003) & 0.0993(0.0049) & 0.0676(0.0064) & 0.0326(0.0041) & 0.0173(0.0057) \\
LS1 & 10 & 0.0176(0.0003) & 0.1079(0.0051) & 0.0648(0.0062) & 0.0311(0.0041) & 0.0165(0.0050) \\
LS1 & 25 & 0.0170(0.0003) & 0.1032(0.0050) & 0.0661(0.0054) & 0.0315(0.0046) & 0.0166(0.0058) \\
LS2 & 5 & 0.0167(0.0002) & 0.0749(0.0047) & 0.0500(0.0057) & 0.0340(0.0048) & 0.0190(0.0071) \\
LS2 & 10 & 0.0168(0.0002) & 0.0822(0.0059) & 0.0520(0.0057) & 0.0343(0.0047) & 0.0186(0.0070) \\
LS2 & 25 & 0.0166(0.0003) & 0.0819(0.0060) & 0.0501(0.0055) & 0.0340(0.0047) & 0.0193(0.0072) \\
LS3 & 5 & 0.0185(0.0003) & 0.1230(0.0039) & 0.0918(0.0037) & 0.0531(0.0064) & 0.0296(0.0082) \\
LS3 & 10 & 0.0187(0.0003) & 0.1264(0.0037) & 0.0920(0.0037) & 0.0546(0.0063) & 0.0311(0.0088) \\
LS3 & 25 & 0.0187(0.0003) & 0.1315(0.0041) & 0.0942(0.0046) & 0.0536(0.0071) & 0.0298(0.0087) \\
\bottomrule
\end{tabular}
\end{table}

%% file: tables/wcf_sensitivity_factorial.tex
\begin{table}[!ht]
\centering
\scriptsize
\setlength{\tabcolsep}{3pt}
\caption{Interaction grid for the quantile grid size $K$ and the particle budget $M$ on LS1 and LS3 at $n=1000$ over fifty paired replications. Entries are end-to-end reference-TCATE Monte Carlo RMSEs with Monte Carlo standard errors in parentheses; lower is better, rows are $K$, and columns are $M$.}
\label{tab:sensitivity-factorial}
\begin{tabular}{lrrr}
\toprule
$K$ & $M=5$ & $M=10$ & $M=25$ \\
\midrule
\multicolumn{4}{l}{\textit{Panel A: LS1}} \\
\midrule
5 & 0.0309(0.0042) & 0.0333(0.0048) & 0.0333(0.0047) \\
25 & 0.0326(0.0041) & 0.0311(0.0041) & 0.0315(0.0046) \\
49 & 0.0312(0.0044) & 0.0307(0.0039) & 0.0315(0.0042) \\
\multicolumn{4}{l}{\textit{Panel B: LS3}} \\
\midrule
5 & 0.0562(0.0065) & 0.0540(0.0066) & 0.0563(0.0067) \\
25 & 0.0531(0.0064) & 0.0546(0.0063) & 0.0536(0.0071) \\
49 & 0.0531(0.0066) & 0.0524(0.0059) & 0.0541(0.0061) \\
\bottomrule
\end{tabular}
\end{table}

%% file: tables/wcf_sensitivity_resolution.tex
\begin{table}[!ht]
\centering
\scriptsize
\setlength{\tabcolsep}{3pt}
\caption{Seed-paired resolution differences at $n=1000$ for designs LS0 to LS3, computed as $K=49$ minus $K=25$ at $M=10$ over fifty paired replications. Entries are percentage changes, so a negative value means the larger grid has the smaller error; the value in parentheses is the Monte Carlo standard error of the paired percentage difference in percentage points. Law is a mean, the other columns are errors. The native-grid target changes with $K$, so these differences do not separate quadrature resolution from tail coverage.}
\label{tab:sensitivity-resolution}
\begin{tabular}{lrrrrr}
\toprule
DGP & Law & Mean-quantile & Ref. TCATE plug-in & Ref. TCATE AIPW & Ref. TATE AIPW \\
\midrule
LS0 & -2.6(1.5) & -9.1(13.5) & -4.0(8.1) & 2.6(3.2) & 6.2(6.9) \\
LS1 & -2.6(1.7) & -0.2(4.4) & -0.7(4.8) & -1.0(3.4) & -6.6(5.5) \\
LS2 & -2.8(1.3) & 1.3(5.9) & -4.2(3.2) & -1.1(2.8) & 13.3(7.2) \\
LS3 & -0.8(1.4) & -0.1(2.8) & 2.0(2.7) & -3.6(2.9) & -1.8(5.9) \\
\bottomrule
\end{tabular}
\end{table}

%% file: tables/wcf_sensitivity_assignment.tex
\begin{table}[!ht]
\centering
\scriptsize
\setlength{\tabcolsep}{3pt}
\caption{Symmetric and nonlinear assignment designs at $n=500$ and $n=1000$ over fifty paired replications. Entries are end-to-end reference-TCATE Monte Carlo RMSEs with Monte Carlo standard errors in parentheses; lower is better. WCF uses its cross-fitted AIPW layer and the forest baselines use plug-in contrasts. SYM-RANDOM is a constant-propensity calibration control, SYM-LIN uses a linear assignment index, and SYM-NL and SYM-MU use nonlinear assignment with the same symmetric outcome laws.}
\label{tab:sensitivity-assignment}
\begin{tabular}{lrrr}
\toprule
DGP & WCF & Causal-DRF & DRF \\
\midrule
\multicolumn{4}{l}{\textit{Panel A: $n=500$}} \\
\midrule
SYM-RANDOM & 0.0449(0.0056) & 0.0522(0.0058) & 0.0472(0.0051) \\
SYM-LIN & 0.0491(0.0049) & 0.0686(0.0064) & 0.0592(0.0068) \\
SYM-NL & 0.0562(0.0059) & 0.0856(0.0060) & 0.0720(0.0060) \\
SYM-MU & 0.0487(0.0054) & 0.0562(0.0062) & 0.0502(0.0064) \\
\multicolumn{4}{l}{\textit{Panel B: $n=1000$}} \\
\midrule
SYM-RANDOM & 0.0306(0.0036) & 0.0354(0.0046) & 0.0329(0.0044) \\
SYM-LIN & 0.0391(0.0050) & 0.0406(0.0051) & 0.0364(0.0055) \\
SYM-NL & 0.0420(0.0057) & 0.0813(0.0044) & 0.0725(0.0049) \\
SYM-MU & 0.0361(0.0045) & 0.0506(0.0050) & 0.0459(0.0047) \\
\bottomrule
\end{tabular}
\end{table}

%% file: tables/wcf_sensitivity_propensity.tex
\begin{table}[!ht]
\centering
\scriptsize
\setlength{\tabcolsep}{3pt}
\caption{Propensity-model sensitivity on the SYM-NL and SYM-MU designs at $n=500$ and $n=1000$ over fifty paired replications. Entries are end-to-end reference-TCATE Monte Carlo RMSEs with Monte Carlo standard errors in parentheses; lower is better. The columns replace the default logistic propensity with a random-forest factory, a gradient-boosting factory, and the oracle propensity, which is a diagnostic and not a feasible competitor. The outcome learner, grid, and particle budget are held fixed.}
\label{tab:sensitivity-propensity}
\begin{tabular}{lrrrr}
\toprule
DGP & Logistic & Random forest & Gradient boosting & Oracle \\
\midrule
\multicolumn{5}{l}{\textit{Panel A: $n=500$}} \\
\midrule
SYM-NL & 0.0562(0.0059) & 0.0575(0.0060) & 0.0575(0.0060) & 0.0575(0.0060) \\
SYM-MU & 0.0487(0.0054) & 0.0487(0.0045) & 0.0487(0.0045) & 0.0487(0.0045) \\
\multicolumn{5}{l}{\textit{Panel B: $n=1000$}} \\
\midrule
SYM-NL & 0.0420(0.0057) & 0.0427(0.0055) & 0.0427(0.0055) & 0.0427(0.0055) \\
SYM-MU & 0.0361(0.0045) & 0.0348(0.0042) & 0.0348(0.0042) & 0.0348(0.0042) \\
\bottomrule
\end{tabular}
\vspace{2pt}
{\footnotesize \textit{Note:} the confirmatory protocol reports the common cross-fitted AIPW view, whose propensity layer is fixed across columns, so the factory choice has little effect on the tabulated errors. The estimator's internal reference-estimate influence-function SE, a calibration diagnostic, is logistic 0.021, random forest 0.022, gradient boosting 0.079, and oracle 0.022; the gradient-boosting factory is roughly three to four times less stable than the alternatives, consistent with the archive's miscalibration diagnosis.}
\end{table}

%% file: tables/wcf_sensitivity_placebo.tex
\begin{table}[!ht]
\centering
\scriptsize
\setlength{\tabcolsep}{3pt}
\caption{Null companion designs with exactly zero treatment effects, so every entry is a false-effect error. Entries are end-to-end Monte Carlo RMSEs pooled over the six null companion designs and fifty paired replications, with Monte Carlo standard errors in parentheses; lower is better. WCF uses its cross-fitted AIPW layer and the forest baselines use plug-in contrasts. The first two rows use the grid-mean functional, and the last two rows use the reference-distance marginal and conditional targets.}
\label{tab:sensitivity-placebo}
\begin{tabular}{lrrr}
\toprule
Target & WCF & Causal-DRF & DRF \\
\midrule
\multicolumn{4}{l}{\textit{Panel A: $n=500$}} \\
\midrule
TATE grid mean & 0.0400(0.0038) & 0.1684(0.0070) & 0.1274(0.0074) \\
TCATE grid mean & 0.0664(0.0028) & 0.1739(0.0068) & 0.1348(0.0065) \\
Ref. TATE & 0.0256(0.0033) & 0.0269(0.0033) & 0.0269(0.0033) \\
Ref. TCATE & 0.0518(0.0025) & 0.0554(0.0037) & 0.0624(0.0039) \\
\multicolumn{4}{l}{\textit{Panel B: $n=1000$}} \\
\midrule
TATE grid mean & 0.0336(0.0034) & 0.1254(0.0067) & 0.0963(0.0067) \\
TCATE grid mean & 0.0488(0.0024) & 0.1300(0.0062) & 0.1013(0.0056) \\
Ref. TATE & 0.0178(0.0027) & 0.0192(0.0028) & 0.0182(0.0026) \\
Ref. TCATE & 0.0369(0.0020) & 0.0546(0.0039) & 0.0595(0.0039) \\
\bottomrule
\end{tabular}
\end{table}

%% file: tables/sample_size_ls.tex
\begingroup
\tiny
\setlength{\tabcolsep}{1.5pt}
\begin{longtable}{lllrrrrr}
\caption{Sample-size sensitivity for the LS family. Law is mean MMD$^2$; other entries are Monte Carlo RMSE. Parentheses are Monte Carlo SEs over ten replications.}\label{tab:sample-size-ls}\\
\toprule
Design & $n$ & Method & Law & TATE & TCATE & Ref. TATE & Ref. TCATE \\
\midrule
\endfirsthead
\toprule
Design & $n$ & Method & Law & TATE & TCATE & Ref. TATE & Ref. TCATE \\
\midrule
\endhead
LS0 & 125 & WCF & 0.0300(0.0019) & 0.0616(0.0070) & 0.1264(0.0186) & 0.0699(0.0111) & 0.1486(0.0301) \\
LS0 & 125 & Causal-DRF & 0.1136(0.0035) & 0.2094(0.0150) & 0.2095(0.0149) & 0.2497(0.0166) & 0.2498(0.0165) \\
LS0 & 125 & DRF & 0.1520(0.0044) & 0.2241(0.0203) & 0.2241(0.0203) & 0.2658(0.0223) & 0.2658(0.0223) \\
\addlinespace
LS0 & 250 & WCF & 0.0255(0.0009) & 0.0369(0.0056) & 0.0686(0.0064) & 0.0399(0.0078) & 0.0715(0.0090) \\
LS0 & 250 & Causal-DRF & 0.0899(0.0026) & 0.2067(0.0089) & 0.2070(0.0089) & 0.2388(0.0114) & 0.2392(0.0114) \\
LS0 & 250 & DRF & 0.1330(0.0031) & 0.2180(0.0102) & 0.2180(0.0102) & 0.2518(0.0119) & 0.2518(0.0119) \\
\addlinespace
LS0 & 500 & WCF & 0.0208(0.0010) & 0.0163(0.0025) & 0.0423(0.0037) & 0.0237(0.0035) & 0.0471(0.0056) \\
LS0 & 500 & Causal-DRF & 0.0651(0.0018) & 0.1692(0.0076) & 0.1692(0.0076) & 0.1895(0.0099) & 0.1896(0.0099) \\
LS0 & 500 & DRF & 0.0905(0.0021) & 0.1670(0.0070) & 0.1670(0.0070) & 0.1881(0.0093) & 0.1882(0.0093) \\
\addlinespace
LS0 & 1000 & WCF & 0.0153(0.0005) & 0.0144(0.0012) & 0.0273(0.0024) & 0.0163(0.0018) & 0.0296(0.0026) \\
LS0 & 1000 & Causal-DRF & 0.0467(0.0010) & 0.1387(0.0061) & 0.1388(0.0061) & 0.1587(0.0086) & 0.1589(0.0086) \\
LS0 & 1000 & DRF & 0.0641(0.0013) & 0.1357(0.0049) & 0.1357(0.0048) & 0.1555(0.0069) & 0.1556(0.0069) \\
\addlinespace
\midrule
LS2 & 125 & WCF & 0.0372(0.0022) & 0.0561(0.0142) & 0.1213(0.0096) & 0.0734(0.0178) & 0.1349(0.0206) \\
LS2 & 125 & Causal-DRF & 0.1205(0.0038) & 0.2256(0.0160) & 0.2258(0.0160) & 0.2623(0.0189) & 0.2624(0.0189) \\
LS2 & 125 & DRF & 0.1601(0.0047) & 0.2333(0.0188) & 0.2333(0.0188) & 0.2713(0.0218) & 0.2713(0.0217) \\
\addlinespace
LS2 & 250 & WCF & 0.0283(0.0008) & 0.0280(0.0036) & 0.0669(0.0069) & 0.0299(0.0044) & 0.0745(0.0091) \\
LS2 & 250 & Causal-DRF & 0.0933(0.0024) & 0.2165(0.0115) & 0.2167(0.0115) & 0.2398(0.0144) & 0.2400(0.0144) \\
LS2 & 250 & DRF & 0.1382(0.0032) & 0.2287(0.0125) & 0.2287(0.0125) & 0.2539(0.0149) & 0.2540(0.0149) \\
\addlinespace
LS2 & 500 & WCF & 0.0229(0.0010) & 0.0143(0.0013) & 0.0416(0.0023) & 0.0218(0.0028) & 0.0463(0.0036) \\
LS2 & 500 & Causal-DRF & 0.0683(0.0016) & 0.1724(0.0077) & 0.1724(0.0077) & 0.1883(0.0105) & 0.1884(0.0104) \\
LS2 & 500 & DRF & 0.0978(0.0023) & 0.1848(0.0086) & 0.1848(0.0086) & 0.2048(0.0122) & 0.2048(0.0122) \\
\addlinespace
LS2 & 1000 & WCF & 0.0172(0.0006) & 0.0171(0.0028) & 0.0285(0.0021) & 0.0191(0.0039) & 0.0327(0.0036) \\
LS2 & 1000 & Causal-DRF & 0.0488(0.0010) & 0.1433(0.0052) & 0.1435(0.0053) & 0.1575(0.0071) & 0.1577(0.0071) \\
LS2 & 1000 & DRF & 0.0724(0.0016) & 0.1643(0.0042) & 0.1644(0.0042) & 0.1802(0.0051) & 0.1802(0.0051) \\
\addlinespace
\midrule
LS1 & 125 & WCF & 0.0373(0.0015) & 0.0651(0.0079) & 0.1432(0.0244) & 0.0677(0.0097) & 0.1523(0.0324) \\
LS1 & 125 & Causal-DRF & 0.1082(0.0034) & 0.1858(0.0143) & 0.1861(0.0143) & 0.2181(0.0154) & 0.2185(0.0153) \\
LS1 & 125 & DRF & 0.1431(0.0043) & 0.1986(0.0193) & 0.1986(0.0193) & 0.2313(0.0206) & 0.2314(0.0206) \\
\addlinespace
LS1 & 250 & WCF & 0.0304(0.0010) & 0.0382(0.0073) & 0.0765(0.0082) & 0.0412(0.0082) & 0.0833(0.0102) \\
LS1 & 250 & Causal-DRF & 0.0865(0.0026) & 0.1823(0.0086) & 0.1828(0.0085) & 0.2064(0.0114) & 0.2071(0.0113) \\
LS1 & 250 & DRF & 0.1226(0.0033) & 0.1906(0.0096) & 0.1907(0.0096) & 0.2157(0.0112) & 0.2158(0.0111) \\
\addlinespace
LS1 & 500 & WCF & 0.0228(0.0009) & 0.0229(0.0033) & 0.0487(0.0026) & 0.0279(0.0067) & 0.0545(0.0071) \\
LS1 & 500 & Causal-DRF & 0.0627(0.0023) & 0.1466(0.0073) & 0.1468(0.0073) & 0.1590(0.0088) & 0.1593(0.0088) \\
LS1 & 500 & DRF & 0.0821(0.0022) & 0.1498(0.0069) & 0.1499(0.0068) & 0.1637(0.0090) & 0.1638(0.0090) \\
\addlinespace
LS1 & 1000 & WCF & 0.0169(0.0005) & 0.0155(0.0017) & 0.0298(0.0031) & 0.0139(0.0026) & 0.0318(0.0028) \\
LS1 & 1000 & Causal-DRF & 0.0455(0.0009) & 0.1207(0.0069) & 0.1209(0.0069) & 0.1351(0.0091) & 0.1355(0.0091) \\
LS1 & 1000 & DRF & 0.0591(0.0012) & 0.1262(0.0042) & 0.1262(0.0042) & 0.1404(0.0060) & 0.1405(0.0060) \\
\addlinespace
\midrule
LS3 & 125 & WCF & 0.0455(0.0020) & 0.0938(0.0133) & 0.1715(0.0449) & 0.0929(0.0119) & 0.1886(0.0658) \\
LS3 & 125 & Causal-DRF & 0.1178(0.0040) & 0.2879(0.0190) & 0.2881(0.0190) & 0.3205(0.0196) & 0.3208(0.0196) \\
LS3 & 125 & DRF & 0.1538(0.0051) & 0.2975(0.0193) & 0.2975(0.0193) & 0.3309(0.0206) & 0.3310(0.0206) \\
\addlinespace
LS3 & 250 & WCF & 0.0353(0.0012) & 0.0450(0.0070) & 0.0956(0.0110) & 0.0453(0.0150) & 0.1019(0.0135) \\
LS3 & 250 & Causal-DRF & 0.0975(0.0030) & 0.2720(0.0112) & 0.2721(0.0112) & 0.3013(0.0141) & 0.3015(0.0140) \\
LS3 & 250 & DRF & 0.1307(0.0035) & 0.2817(0.0128) & 0.2818(0.0128) & 0.3126(0.0153) & 0.3127(0.0153) \\
\addlinespace
LS3 & 500 & WCF & 0.0269(0.0010) & 0.0484(0.0052) & 0.0706(0.0050) & 0.0383(0.0057) & 0.0715(0.0071) \\
LS3 & 500 & Causal-DRF & 0.0773(0.0019) & 0.2465(0.0076) & 0.2466(0.0076) & 0.2689(0.0100) & 0.2692(0.0099) \\
LS3 & 500 & DRF & 0.0958(0.0029) & 0.2619(0.0074) & 0.2620(0.0073) & 0.2850(0.0095) & 0.2852(0.0095) \\
\addlinespace
LS3 & 1000 & WCF & 0.0188(0.0007) & 0.0242(0.0040) & 0.0470(0.0046) & 0.0217(0.0041) & 0.0508(0.0057) \\
LS3 & 1000 & Causal-DRF & 0.0603(0.0013) & 0.2248(0.0063) & 0.2249(0.0063) & 0.2469(0.0073) & 0.2471(0.0073) \\
LS3 & 1000 & DRF & 0.0773(0.0014) & 0.2477(0.0043) & 0.2477(0.0043) & 0.2705(0.0052) & 0.2706(0.0052) \\
\addlinespace
\midrule
\bottomrule
\end{longtable}
\endgroup

%% file: tables/sample_size_s.tex
\begingroup
\tiny
\setlength{\tabcolsep}{1.5pt}
\begin{longtable}{lllrrrrr}
\caption{Sample-size sensitivity for the S family. Law is mean MMD$^2$; other entries are Monte Carlo RMSE. Parentheses are Monte Carlo SEs over ten replications.}\label{tab:sample-size-s}\\
\toprule
Design & $n$ & Method & Law & TATE & TCATE & Ref. TATE & Ref. TCATE \\
\midrule
\endfirsthead
\toprule
Design & $n$ & Method & Law & TATE & TCATE & Ref. TATE & Ref. TCATE \\
\midrule
\endhead
S1 & 125 & WCF & 0.0877(0.0043) & 0.0372(0.0087) & 0.0620(0.0065) & 0.0182(0.0059) & 0.0702(0.0079) \\
S1 & 125 & Causal-DRF & 0.2911(0.0121) & 0.1899(0.0159) & 0.2750(0.0111) & 0.0420(0.0078) & 0.1804(0.0026) \\
S1 & 125 & DRF & 0.5065(0.0079) & 0.2448(0.0228) & 0.4023(0.0138) & 0.0419(0.0078) & 0.1925(0.0018) \\
\addlinespace
S1 & 250 & WCF & 0.0573(0.0033) & 0.0147(0.0047) & 0.0325(0.0035) & 0.0228(0.0043) & 0.0363(0.0036) \\
S1 & 250 & Causal-DRF & 0.1930(0.0068) & 0.0948(0.0091) & 0.1532(0.0078) & 0.0329(0.0090) & 0.1499(0.0019) \\
S1 & 250 & DRF & 0.3209(0.0095) & 0.1113(0.0123) & 0.2980(0.0386) & 0.0376(0.0095) & 0.1873(0.0028) \\
\addlinespace
S1 & 500 & WCF & 0.0341(0.0018) & 0.0072(0.0007) & 0.0207(0.0016) & 0.0098(0.0029) & 0.0197(0.0014) \\
S1 & 500 & Causal-DRF & 0.1386(0.0049) & 0.0622(0.0046) & 0.1027(0.0030) & 0.0222(0.0041) & 0.1183(0.0036) \\
S1 & 500 & DRF & 0.1652(0.0054) & 0.0399(0.0052) & 0.1049(0.0044) & 0.0227(0.0059) & 0.1434(0.0051) \\
\addlinespace
S1 & 1000 & WCF & 0.0234(0.0015) & 0.0061(0.0006) & 0.0122(0.0009) & 0.0049(0.0011) & 0.0110(0.0013) \\
S1 & 1000 & Causal-DRF & 0.1033(0.0036) & 0.0302(0.0023) & 0.0678(0.0014) & 0.0097(0.0012) & 0.0889(0.0023) \\
S1 & 1000 & DRF & 0.1191(0.0041) & 0.0169(0.0012) & 0.0544(0.0018) & 0.0069(0.0021) & 0.0802(0.0023) \\
\addlinespace
\midrule
S2 & 125 & WCF & 0.0256(0.0011) & 0.0474(0.0057) & 0.1250(0.0077) & 0.0500(0.0097) & 0.1168(0.0185) \\
S2 & 125 & Causal-DRF & 0.0444(0.0015) & 0.1662(0.0184) & 0.1671(0.0184) & 0.0480(0.0093) & 0.0506(0.0086) \\
S2 & 125 & DRF & 0.0647(0.0013) & 0.1825(0.0197) & 0.1825(0.0197) & 0.0481(0.0107) & 0.0481(0.0107) \\
\addlinespace
S2 & 250 & WCF & 0.0192(0.0008) & 0.0440(0.0066) & 0.0866(0.0080) & 0.0312(0.0074) & 0.0680(0.0073) \\
S2 & 250 & Causal-DRF & 0.0306(0.0009) & 0.0889(0.0119) & 0.0903(0.0116) & 0.0334(0.0083) & 0.0387(0.0065) \\
S2 & 250 & DRF & 0.0418(0.0008) & 0.0812(0.0108) & 0.1099(0.0107) & 0.0325(0.0067) & 0.0342(0.0062) \\
\addlinespace
S2 & 500 & WCF & 0.0135(0.0005) & 0.0329(0.0077) & 0.0803(0.0059) & 0.0354(0.0073) & 0.0686(0.0066) \\
S2 & 500 & Causal-DRF & 0.0222(0.0007) & 0.0681(0.0106) & 0.0734(0.0100) & 0.0352(0.0109) & 0.0411(0.0088) \\
S2 & 500 & DRF & 0.0189(0.0005) & 0.0357(0.0086) & 0.0602(0.0120) & 0.0322(0.0101) & 0.0431(0.0069) \\
\addlinespace
S2 & 1000 & WCF & 0.0095(0.0003) & 0.0203(0.0028) & 0.0396(0.0029) & 0.0206(0.0091) & 0.0423(0.0029) \\
S2 & 1000 & Causal-DRF & 0.0154(0.0003) & 0.0420(0.0061) & 0.0463(0.0055) & 0.0215(0.0111) & 0.0264(0.0075) \\
S2 & 1000 & DRF & 0.0138(0.0003) & 0.0188(0.0035) & 0.0302(0.0032) & 0.0211(0.0104) & 0.0273(0.0068) \\
\addlinespace
\midrule
S3 & 125 & WCF & 0.0576(0.0039) & 0.0598(0.0108) & 0.1426(0.0123) & 0.0686(0.0189) & 0.1138(0.0164) \\
S3 & 125 & Causal-DRF & 0.1086(0.0035) & 0.1627(0.0154) & 0.1645(0.0152) & 0.0592(0.0109) & 0.0605(0.0108) \\
S3 & 125 & DRF & 0.1605(0.0026) & 0.1860(0.0172) & 0.1860(0.0172) & 0.0597(0.0116) & 0.0598(0.0116) \\
\addlinespace
S3 & 250 & WCF & 0.0355(0.0026) & 0.0465(0.0042) & 0.0818(0.0055) & 0.0335(0.0080) & 0.0738(0.0078) \\
S3 & 250 & Causal-DRF & 0.0703(0.0019) & 0.0739(0.0096) & 0.0793(0.0084) & 0.0472(0.0112) & 0.0541(0.0093) \\
S3 & 250 & DRF & 0.1069(0.0073) & 0.0758(0.0109) & 0.1130(0.0166) & 0.0507(0.0114) & 0.0525(0.0108) \\
\addlinespace
S3 & 500 & WCF & 0.0312(0.0021) & 0.0367(0.0068) & 0.0822(0.0064) & 0.0350(0.0089) & 0.0757(0.0079) \\
S3 & 500 & Causal-DRF & 0.0515(0.0019) & 0.0540(0.0099) & 0.0643(0.0077) & 0.0339(0.0089) & 0.0471(0.0089) \\
S3 & 500 & DRF & 0.0553(0.0014) & 0.0389(0.0077) & 0.0772(0.0115) & 0.0315(0.0098) & 0.0428(0.0091) \\
\addlinespace
S3 & 1000 & WCF & 0.0224(0.0007) & 0.0221(0.0047) & 0.0412(0.0024) & 0.0227(0.0043) & 0.0459(0.0060) \\
S3 & 1000 & Causal-DRF & 0.0374(0.0009) & 0.0279(0.0055) & 0.0383(0.0036) & 0.0177(0.0032) & 0.0251(0.0023) \\
S3 & 1000 & DRF & 0.0431(0.0011) & 0.0195(0.0037) & 0.0381(0.0039) & 0.0188(0.0035) & 0.0250(0.0026) \\
\addlinespace
\midrule
S6 & 125 & WCF & 0.0455(0.0035) & 0.1610(0.0381) & 0.2932(0.0773) & 0.1468(0.0470) & 0.3065(0.0581) \\
S6 & 125 & Causal-DRF & 0.1420(0.0052) & 0.4775(0.0298) & 0.4797(0.0303) & 0.1187(0.0303) & 0.1266(0.0315) \\
S6 & 125 & DRF & 0.2050(0.0044) & 0.5374(0.0324) & 0.5401(0.0324) & 0.1111(0.0300) & 0.1225(0.0270) \\
\addlinespace
S6 & 250 & WCF & 0.0290(0.0011) & 0.0630(0.0130) & 0.1318(0.0192) & 0.0666(0.0148) & 0.1429(0.0227) \\
S6 & 250 & Causal-DRF & 0.0950(0.0043) & 0.3327(0.0291) & 0.3382(0.0284) & 0.1022(0.0180) & 0.1103(0.0154) \\
S6 & 250 & DRF & 0.1165(0.0033) & 0.2916(0.0219) & 0.3304(0.0200) & 0.1085(0.0190) & 0.1178(0.0180) \\
\addlinespace
S6 & 500 & WCF & 0.0184(0.0007) & 0.0613(0.0095) & 0.1174(0.0089) & 0.0625(0.0130) & 0.1326(0.0157) \\
S6 & 500 & Causal-DRF & 0.0579(0.0019) & 0.1852(0.0168) & 0.1961(0.0149) & 0.0747(0.0096) & 0.0870(0.0103) \\
S6 & 500 & DRF & 0.0551(0.0016) & 0.0929(0.0140) & 0.1291(0.0094) & 0.0981(0.0134) & 0.1078(0.0124) \\
\addlinespace
S6 & 1000 & WCF & 0.0131(0.0003) & 0.0351(0.0085) & 0.0850(0.0107) & 0.0535(0.0216) & 0.0923(0.0169) \\
S6 & 1000 & Causal-DRF & 0.0416(0.0007) & 0.0975(0.0082) & 0.1180(0.0062) & 0.0420(0.0131) & 0.0628(0.0081) \\
S6 & 1000 & DRF & 0.0416(0.0008) & 0.0339(0.0040) & 0.0662(0.0051) & 0.0560(0.0164) & 0.0764(0.0106) \\
\addlinespace
\midrule
S4 & 125 & WCF & 0.0796(0.0039) & 0.5150(0.1202) & 0.6438(0.1020) & 0.1156(0.0162) & 0.3462(0.0662) \\
S4 & 125 & Causal-DRF & 0.0274(0.0034) & 0.4004(0.0862) & 0.4494(0.0799) & 0.0995(0.0243) & 0.1315(0.0245) \\
S4 & 125 & DRF & 0.0404(0.0033) & 0.3966(0.0859) & 0.4865(0.0716) & 0.1084(0.0237) & 0.1084(0.0237) \\
\addlinespace
S4 & 250 & WCF & 0.0529(0.0023) & 0.0720(0.0105) & 0.3604(0.0438) & 0.1408(0.0175) & 0.2104(0.0159) \\
S4 & 250 & Causal-DRF & 0.0164(0.0017) & 0.1139(0.0153) & 0.1965(0.0210) & 0.1147(0.0179) & 0.1204(0.0169) \\
S4 & 250 & DRF & 0.0261(0.0018) & 0.1337(0.0212) & 0.2560(0.0232) & 0.1184(0.0192) & 0.1195(0.0189) \\
\addlinespace
S4 & 500 & WCF & 0.0332(0.0019) & 0.1058(0.0375) & 0.2052(0.0203) & 0.0625(0.0116) & 0.1340(0.0166) \\
S4 & 500 & Causal-DRF & 0.0092(0.0007) & 0.1436(0.0261) & 0.2016(0.0173) & 0.0694(0.0104) & 0.0827(0.0098) \\
S4 & 500 & DRF & 0.0108(0.0009) & 0.1364(0.0288) & 0.1794(0.0228) & 0.0684(0.0098) & 0.0818(0.0087) \\
\addlinespace
S4 & 1000 & WCF & 0.0180(0.0008) & 0.0857(0.0165) & 0.1433(0.0190) & 0.0267(0.0071) & 0.0943(0.0100) \\
S4 & 1000 & Causal-DRF & 0.0064(0.0004) & 0.0788(0.0100) & 0.1208(0.0121) & 0.0345(0.0093) & 0.0643(0.0105) \\
S4 & 1000 & DRF & 0.0059(0.0003) & 0.0707(0.0089) & 0.1185(0.0076) & 0.0285(0.0075) & 0.0540(0.0072) \\
\addlinespace
\midrule
S5 & 125 & WCF & 0.0249(0.0012) & 0.0550(0.0071) & 0.1355(0.0100) & 0.0324(0.0040) & 0.1080(0.0162) \\
S5 & 125 & Causal-DRF & 0.0425(0.0015) & 0.1624(0.0176) & 0.1734(0.0163) & 0.0433(0.0082) & 0.0469(0.0072) \\
S5 & 125 & DRF & 0.0652(0.0013) & 0.1832(0.0195) & 0.1981(0.0180) & 0.0438(0.0098) & 0.0476(0.0088) \\
\addlinespace
S5 & 250 & WCF & 0.0204(0.0008) & 0.0428(0.0084) & 0.0871(0.0070) & 0.0236(0.0055) & 0.0607(0.0053) \\
S5 & 250 & Causal-DRF & 0.0285(0.0007) & 0.0833(0.0105) & 0.0943(0.0093) & 0.0291(0.0067) & 0.0370(0.0047) \\
S5 & 250 & DRF & 0.0432(0.0008) & 0.0824(0.0109) & 0.1233(0.0098) & 0.0291(0.0056) & 0.0346(0.0045) \\
\addlinespace
S5 & 500 & WCF & 0.0141(0.0004) & 0.0337(0.0061) & 0.0791(0.0056) & 0.0308(0.0054) & 0.0593(0.0050) \\
S5 & 500 & Causal-DRF & 0.0189(0.0006) & 0.0589(0.0101) & 0.0735(0.0085) & 0.0306(0.0094) & 0.0387(0.0069) \\
S5 & 500 & DRF & 0.0198(0.0006) & 0.0374(0.0081) & 0.0702(0.0105) & 0.0282(0.0092) & 0.0392(0.0059) \\
\addlinespace
S5 & 1000 & WCF & 0.0095(0.0003) & 0.0187(0.0024) & 0.0391(0.0033) & 0.0180(0.0059) & 0.0346(0.0029) \\
S5 & 1000 & Causal-DRF & 0.0129(0.0003) & 0.0283(0.0049) & 0.0403(0.0036) & 0.0184(0.0077) & 0.0259(0.0044) \\
S5 & 1000 & DRF & 0.0142(0.0003) & 0.0202(0.0035) & 0.0393(0.0027) & 0.0186(0.0072) & 0.0263(0.0045) \\
\addlinespace
\midrule
\bottomrule
\end{longtable}
\endgroup

%% file: tables/sample_size_z.tex
\begingroup
\tiny
\setlength{\tabcolsep}{1.5pt}
\begin{longtable}{lllrrr}
\caption{Sample-size sensitivity for the Z family. Law is mean MMD$^2$; other entries are Monte Carlo RMSE. Parentheses are Monte Carlo SEs over ten replications.}\label{tab:sample-size-z}\\
\toprule
Design & $n$ & Method & Law & Mass & Mass TCATE \\
\midrule
\endfirsthead
\toprule
Design & $n$ & Method & Law & Mass & Mass TCATE \\
\midrule
\endhead
Z1 & 125 & Two-part WCF & 0.0549(0.0047) & 0.1465(0.0076) & 0.1187(0.0204) \\
Z1 & 125 & Causal-DRF & 0.0732(0.0028) & 0.1659(0.0031) & 0.1125(0.0188) \\
Z1 & 125 & DRF & 0.1000(0.0041) & 0.1958(0.0043) & 0.1218(0.0259) \\
\addlinespace
Z1 & 250 & Two-part WCF & 0.0380(0.0041) & 0.1150(0.0090) & 0.1050(0.0214) \\
Z1 & 250 & Causal-DRF & 0.0556(0.0040) & 0.1422(0.0071) & 0.1043(0.0308) \\
Z1 & 250 & DRF & 0.0875(0.0031) & 0.1831(0.0038) & 0.1193(0.0245) \\
\addlinespace
Z1 & 500 & Two-part WCF & 0.0226(0.0014) & 0.0764(0.0037) & 0.0603(0.0065) \\
Z1 & 500 & Causal-DRF & 0.0366(0.0025) & 0.1113(0.0056) & 0.0676(0.0127) \\
Z1 & 500 & DRF & 0.0502(0.0031) & 0.1365(0.0057) & 0.0756(0.0142) \\
\addlinespace
Z1 & 1000 & Two-part WCF & 0.0154(0.0011) & 0.0550(0.0039) & 0.0401(0.0046) \\
Z1 & 1000 & Causal-DRF & 0.0247(0.0011) & 0.0905(0.0027) & 0.0473(0.0083) \\
Z1 & 1000 & DRF & 0.0291(0.0016) & 0.1036(0.0036) & 0.0535(0.0101) \\
\addlinespace
\midrule
Z2 & 125 & Two-part WCF & 0.0554(0.0074) & 0.1444(0.0136) & 0.1616(0.0507) \\
Z2 & 125 & Causal-DRF & 0.0767(0.0056) & 0.1771(0.0077) & 0.1557(0.0468) \\
Z2 & 125 & DRF & 0.1007(0.0080) & 0.2045(0.0096) & 0.1597(0.0539) \\
\addlinespace
Z2 & 250 & Two-part WCF & 0.0365(0.0045) & 0.1161(0.0095) & 0.1049(0.0225) \\
Z2 & 250 & Causal-DRF & 0.0533(0.0042) & 0.1445(0.0073) & 0.1005(0.0305) \\
Z2 & 250 & DRF & 0.0838(0.0030) & 0.1844(0.0043) & 0.1114(0.0262) \\
\addlinespace
Z2 & 500 & Two-part WCF & 0.0179(0.0011) & 0.0707(0.0039) & 0.0474(0.0046) \\
Z2 & 500 & Causal-DRF & 0.0371(0.0023) & 0.1180(0.0044) & 0.0857(0.0092) \\
Z2 & 500 & DRF & 0.0500(0.0016) & 0.1397(0.0030) & 0.0869(0.0090) \\
\addlinespace
Z2 & 1000 & Two-part WCF & 0.0136(0.0007) & 0.0556(0.0031) & 0.0508(0.0118) \\
Z2 & 1000 & Causal-DRF & 0.0260(0.0015) & 0.0958(0.0029) & 0.0617(0.0045) \\
Z2 & 1000 & DRF & 0.0304(0.0013) & 0.1073(0.0030) & 0.0613(0.0080) \\
\addlinespace
\midrule
Z3 & 125 & Two-part WCF & 0.1345(0.0146) & 0.2405(0.0157) & 0.2420(0.0392) \\
Z3 & 125 & Causal-DRF & 0.2778(0.0125) & 0.3620(0.0077) & 0.4200(0.0289) \\
Z3 & 125 & DRF & 0.3611(0.0150) & 0.4153(0.0086) & 0.4598(0.0314) \\
\addlinespace
Z3 & 250 & Two-part WCF & 0.0506(0.0038) & 0.1332(0.0072) & 0.1007(0.0134) \\
Z3 & 250 & Causal-DRF & 0.1742(0.0108) & 0.2793(0.0090) & 0.3259(0.0146) \\
Z3 & 250 & DRF & 0.3192(0.0069) & 0.3929(0.0046) & 0.4240(0.0138) \\
\addlinespace
Z3 & 500 & Two-part WCF & 0.0301(0.0036) & 0.0950(0.0096) & 0.0637(0.0120) \\
Z3 & 500 & Causal-DRF & 0.1156(0.0069) & 0.2212(0.0080) & 0.2606(0.0137) \\
Z3 & 500 & DRF & 0.2653(0.0112) & 0.3620(0.0072) & 0.3933(0.0149) \\
\addlinespace
Z3 & 1000 & Two-part WCF & 0.0176(0.0010) & 0.0623(0.0036) & 0.0497(0.0083) \\
Z3 & 1000 & Causal-DRF & 0.0766(0.0028) & 0.1737(0.0039) & 0.2061(0.0084) \\
Z3 & 1000 & DRF & 0.2472(0.0099) & 0.3505(0.0080) & 0.3843(0.0128) \\
\addlinespace
\midrule
Z0 & 125 & Two-part WCF & 0.0504(0.0074) & 0.1444(0.0136) & 0.1616(0.0507) \\
Z0 & 125 & Causal-DRF & 0.0747(0.0057) & 0.1770(0.0080) & 0.1579(0.0478) \\
Z0 & 125 & DRF & 0.0984(0.0080) & 0.2045(0.0096) & 0.1597(0.0539) \\
\addlinespace
Z0 & 250 & Two-part WCF & 0.0360(0.0047) & 0.1161(0.0095) & 0.1049(0.0225) \\
Z0 & 250 & Causal-DRF & 0.0520(0.0042) & 0.1451(0.0072) & 0.1003(0.0304) \\
Z0 & 250 & DRF & 0.0818(0.0031) & 0.1844(0.0043) & 0.1113(0.0262) \\
\addlinespace
Z0 & 500 & Two-part WCF & 0.0173(0.0011) & 0.0707(0.0039) & 0.0474(0.0046) \\
Z0 & 500 & Causal-DRF & 0.0349(0.0021) & 0.1166(0.0043) & 0.0842(0.0091) \\
Z0 & 500 & DRF & 0.0486(0.0016) & 0.1396(0.0031) & 0.0864(0.0089) \\
\addlinespace
Z0 & 1000 & Two-part WCF & 0.0133(0.0007) & 0.0556(0.0031) & 0.0508(0.0118) \\
Z0 & 1000 & Causal-DRF & 0.0246(0.0013) & 0.0951(0.0026) & 0.0617(0.0048) \\
Z0 & 1000 & DRF & 0.0296(0.0013) & 0.1071(0.0029) & 0.0609(0.0079) \\
\addlinespace
\midrule
\bottomrule
\end{longtable}
\endgroup

%% file: applied_star_appendix.tex
The main text reports a complete applied evaluation on the Project STAR class-size experiment. This appendix presents two further applications of the same estimator: state-year minimum-wage increases and hourly wage distributions, and a village-level cash-transfer experiment in Kenya. Table~\ref{tab:applied-overview} compares the three designs by unit of analysis, sample size, outcome, treatment definition, benchmark, and overlap, where the final column reports the share of fitted propensities outside the overlap band $[0.05,0.95]$. All three applications use the same fixed estimator configuration and report the same five functionals, namely the mean, standard deviation, skewness, upper-half mean, and reference distance to the stated benchmark, together with the tenth and ninetieth percentiles for every design. The summaries below state plainly what each design can support: the cash-transfer application rests on randomized assignment, and the minimum-wage application is an observational design whose calibrated contrasts are descriptive where the propensity support is thin. Additionally, gor the minimum-wage panel, estimates are reported after restricting the sample to the overlap band, because that restriction is the natural diagnostic when positivity is questionable. Where the design diagnostics fail, the estimates are reported as descriptive associations rather than causal effects.

\input{tables/applied_overview.tex}

%% file: tables/applied_overview.tex
\begin{table*}[htbp]
\centering
\caption{Design summary of the three applied studies. Each outcome is represented by $K=25$ midpoint quantiles, and the final column reports the share of fitted propensities outside the overlap band $[0.05,0.95]$.}
\label{tab:applied-overview}
\small
\setlength{\tabcolsep}{4pt}
\begin{tabular}{@{}p{2.0cm}p{1.9cm}p{1.8cm}p{2.4cm}p{2.6cm}p{2.3cm}c@{}}
\toprule
Study & Unit & $n$ (treated/\allowbreak control) & Outcome ($K{=}25$) & Treatment & Benchmark & \shortstack{Outside\\ $[0.05,0.95]$} \\
\midrule
Project STAR & classroom-grade & 1{,}333 (522/811) & within-grade $z$ math scores & small class vs.\ regular/aide & pooled control classrooms & 0\% \\
State minimum wage & state-year & 867 (117/750) & hourly wages, 2016 USD & real rise $>\$0.25$, own-rate & Nordic (scaled); US 2016 & 12.3\% \\
Kenya cash transfers & village & 653 (328/325) & per-capita nondurable consumption & village randomization & pooled control villages & 0\% \\
\bottomrule
\end{tabular}
\end{table*}

%% file: applied_minwage_appendix.tex
\subsection{State-year wage distributions and minimum-wage policy}\label{app:applied-minwage}

The wage analysis uses the NBER extracts of the Current Population Survey (CPS) Merged Outgoing Rotation Groups for 1979 to 2024 \citep{nbermorg}. The primary window is 2000 to 2016, which gives 867 state-year units (51 units per year over 17 years); 117 of these units are treated under the pre-registered event definition. Hourly wages are expressed in 2016 dollars, weighted by the MORG earnings weight, and restricted to respondents aged 16 to 64 who are not self-employed. Each state-year distribution is represented by $K=25$ quantiles of weighted hourly wages. Minimum-wage policy is taken from a historical state minimum-wage panel \citep{vaghul2016historical}. The event definition follows the recent minimum-wage literature in focusing on state-specific policy changes rather than comparisons of levels \citep{cengiz2019effect,lee1999wage}: the pre-registered indicator equals one when the state's real effective minimum wage rises by more than $0.25$, the state's own nominal rate rises, and the increase is not merely the pass-through of a federal change. Two alternatives broaden the definition: one also counts federal increases, and the other classifies a state-year as treated whenever the state rate lies above the federal floor.

\begin{figure}[t]
\centering
\includegraphics[width=\linewidth]{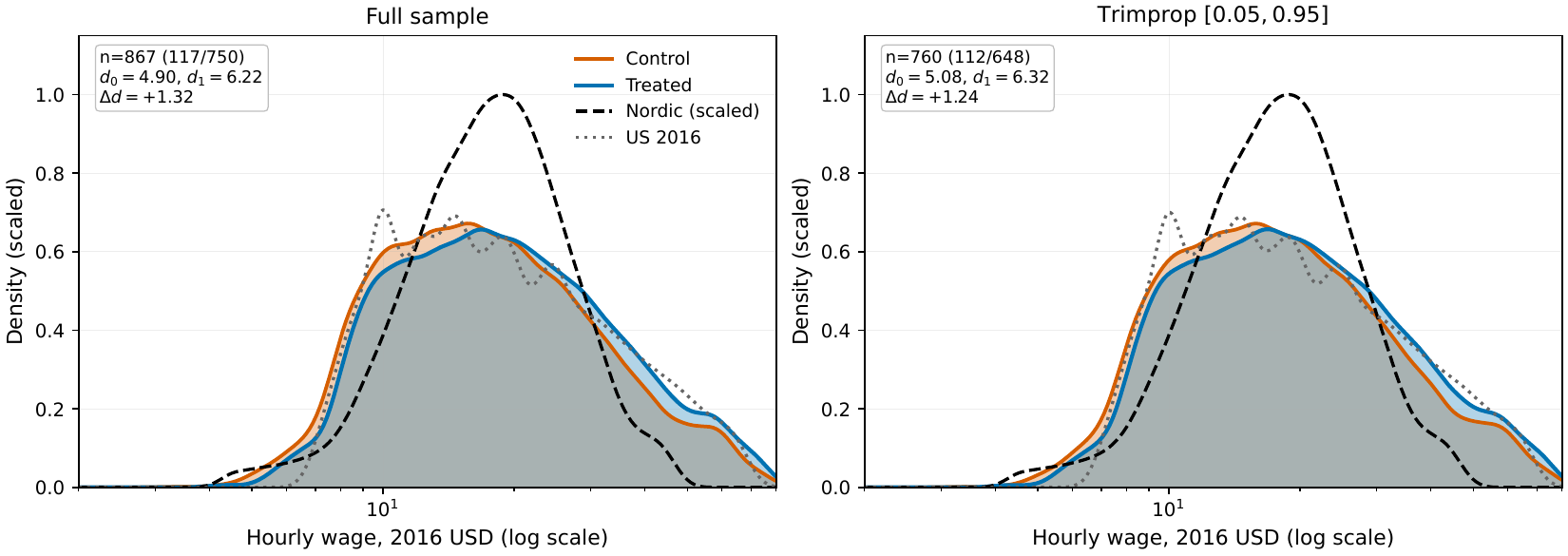}
\caption{Minimum-wage arm densities for the full sample and for the sample retained by the standard overlap band $[0.05,0.95]$. Each panel shows mixture densities of the unit laws, constructed as in Figure~\ref{fig:applied-star-densities}, with the Nordic benchmark (dashed) and the US 2016 national distribution (dotted) on a log scale; $d_0$ and $d_1$ are the mean unit-level distances to the Nordic benchmark and $\Delta d=d_1-d_0$. Restricting to the band leaves the mixture shapes and the benchmark distances close to their full-sample values, and the wage distributions remain far apart on the dollar scale.}
\label{fig:applied-minwage-densities}
\end{figure}

Two external references place the estimated movements on a common scale. The first is the Nordic income distribution, built from Eurostat table ilc\_di01 \citep{eurostat2026ilc}: the national distributions of Denmark, Finland, Norway, and Sweden are combined with population weights into one quantile vector, which is then rescaled to the analysis units by a single documented scalar, the ratio of the median hourly wage in the 2016 US analysis sample, $17.6922$ in 2016 dollars, to the population-weighted Nordic median equivalized disposable income, $26583.71$ in purchasing-power-standard units for income year 2024, equal to $6.6553\times10^{-4}$. Rescaling by one scalar preserves the shape of the Nordic distribution while placing it on the same dollar scale as the state-year wage distributions. The second is the US 2016 national distribution, the weighted quantiles of hourly wages in the 2016 CPS analysis sample, which needs no rescaling and serves as a same-country reference. Figure~\ref{fig:applied-minwage-densities} draws both references against the fitted arm mixtures.

\begin{figure}[ht]
\centering
\includegraphics[width=\linewidth]{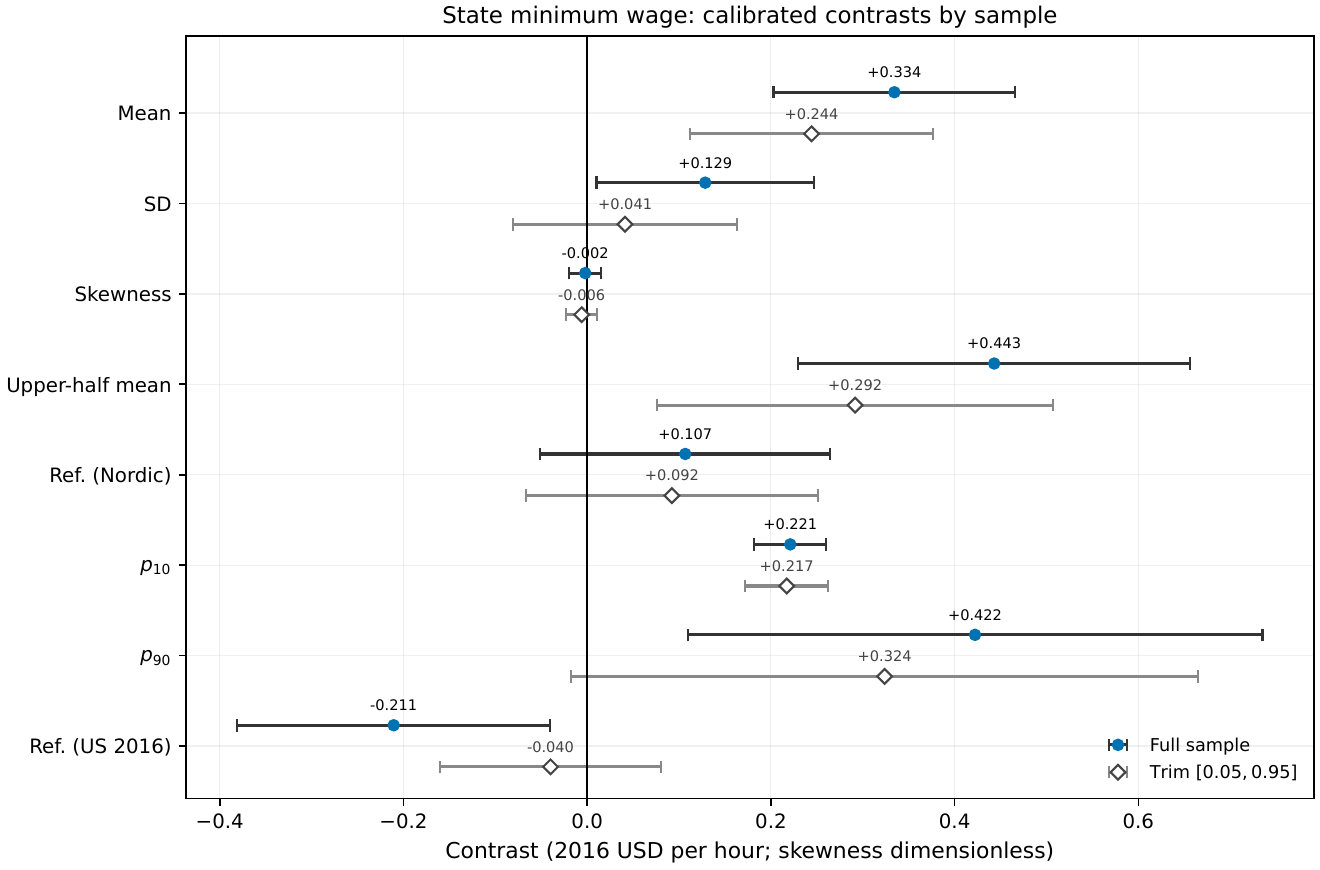}
\caption{State minimum-wage calibrated contrasts by sample, with influence-function standard errors from the first fitting replicate. Diamonds report the sample retained by the standard overlap band $[0.05,0.95]$; the two reference rows use the Nordic benchmark and the US 2016 distribution. Contrasts are in 2016 USD per hour except skewness, which is dimensionless.}
\label{fig:applied-minwage-estimates}
\end{figure}

Figure~\ref{fig:applied-minwage-estimates} reports the primary marginal contrasts. The calibrated mean contrast is $+0.334$, the dispersion contrast $+0.129$, the skewness contrast $-0.002$, and the upper-half mean $+0.443$; the tenth- and ninetieth-percentile contrasts are $+0.221$ and $+0.422$. Relative to the Nordic benchmark the reference contrast is $+0.107$; relative to the US 2016 distribution it is $-0.211$. Both reference contrasts follow the convention of Section~\ref{sec:targets}, where a negative value means the treated law moves closer to the benchmark: the calibrated wage distributions move away from the Nordic benchmark (SE $0.158$) and toward the US 2016 distribution (SE $0.170$). Neither is distinguishable from zero. It is worth remembering that the fitted propensities span 0.020 to 0.408, $12.3$ percent lie below $0.05$ (none above $0.95$), $1.0$ percent sit at the $0.02$ lower clip, and treated state-years have a mean propensity of $0.181$ against $0.128$ for control state-years. Under this limited support at the bottom of the distribution the augmented score can rely heavily on observations that the outcome model fits poorly, so the calibrated contrasts are descriptive summaries rather than stable causal estimates.

Because the full-sample propensities include a nontrivial mass below $0.05$, the model is re-estimated on the sample retained by the standard overlap band $[0.05,0.95]$ (diamonds in Figure~\ref{fig:applied-minwage-estimates}; right panel of Figure~\ref{fig:applied-minwage-densities}). Dropping units whose full-sample fitted propensity lies below $0.05$ (no unit lies above $0.95$) retains $760$ of $867$ units ($112$ treated, $648$ control). The model is then refit on the retained sample, so the reported propensities are not constrained to the interval. The calibrated contrasts shrink but keep their signs: the mean falls from $+0.334$ to $+0.244$, the dispersion from $+0.129$ to $+0.041$, the upper-half mean from $+0.443$ to $+0.292$, and the Nordic reference from $+0.107$ to $+0.092$, while the tenth-percentile contrast stays positive at $+0.217$. The Nordic reference contrast remains unstable across fitting replicates, indicating a null effect, and the US 2016 reference moves from $-0.211$ to $-0.040$, moving it to a null effect. One therefore can read the lower-tail, mean and upper-half mean contrasts as a descriptive association that survives the placebo checks and the overlap restriction, and we do not read the reference contrasts as established effects.

Two limitations bound these results. No state-clustered inference is available, so the reported standard errors do not account for dependence across years within a state. The binary cross-sectional treatment does not identify a causal effect on its own, because treated and untreated state-years differ in many ways that a cross-section cannot absorb; the estimates are therefore associations under the stated event definition.

%% file: applied_cash_appendix.tex
\subsection{Cash transfers and village consumption distributions}\label{app:applied-egger}

Cash transfers can change both the level and the shape of household consumption, and the analyzed design randomized a transfer program across $653$ villages, with $328$ assigned to treatment and $325$ to control \citep{egger2022general,egger2024replication}. Assignment was at the village level, but the two-stage saturation structure links villages within the same local economy, so untreated villages can be affected by transfers made in nearby treated villages and the stable unit treatment value assumption could be questionable. Nonetheless, for this applied study, it is assumed that this spillover effect is virtually zero. Because the program operated at village scale, transfers could also change local prices and demand, so a comparison of treated and control villages captures both the direct transfer and any general-equilibrium response. The outcome is the village distribution of per-capita nondurable consumption, represented by $K=25$ weighted quantiles computed with household sampling weights; $8{,}226$ households contribute. Each village quantile is estimated from few households (median $12.0$, mean $12.597$, range $8$ to $25$), so quantile sampling error is material, especially in the tails where the effective sample is smallest. With roughly a dozen households per village, the lower and upper quantiles are supported by only one or two order statistics, which makes the tail coordinates the noisiest part of each village vector.

\begin{figure}[t]
\centering
\includegraphics[width=\linewidth]{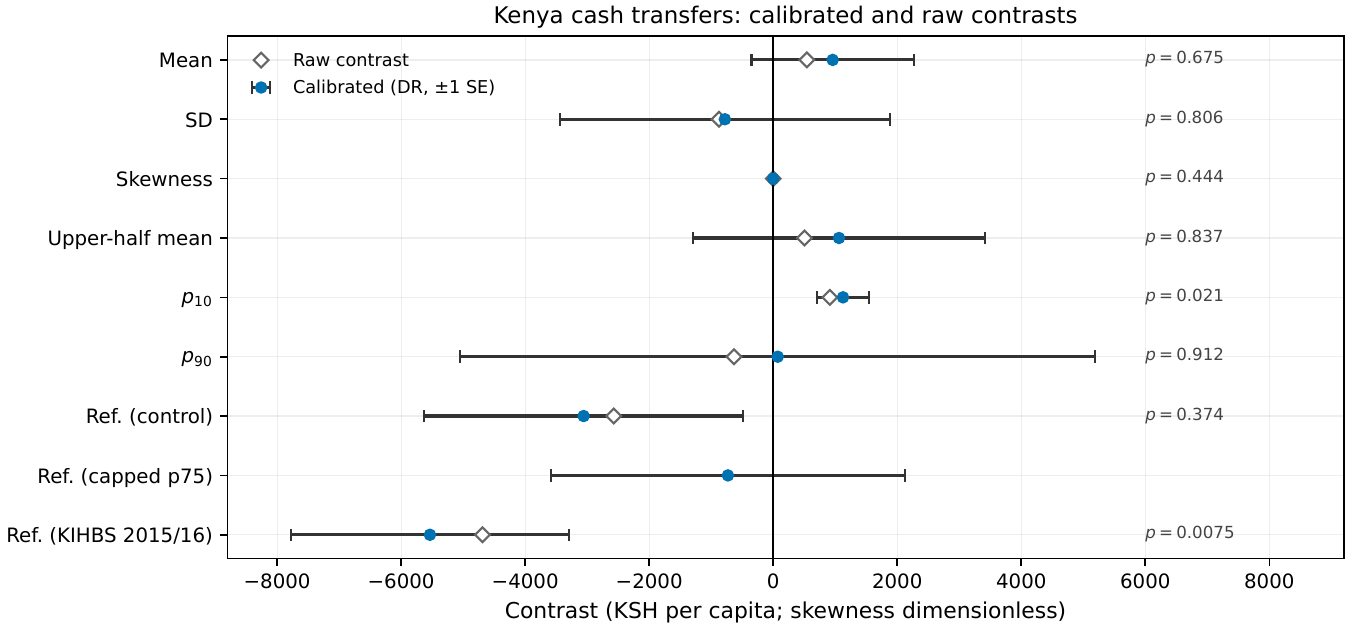}
\caption{Kenya cash-transfer calibrated contrasts with influence-function standard errors (bars) and unadjusted raw contrasts (open diamonds). The right-hand column reports the two-sided permutation $p$-value for each raw contrast. Estimates are in KSH per capita except skewness, which is dimensionless; the last two rows are the reference contrasts to the pooled control benchmark capped at its $75$th percentile (calibrated layer only) and to the external KIHBS 2015/16 benchmark, the latter fitted as a separate variant.}
\label{fig:applied-egger-estimates}
\end{figure}

Three benchmarks are considered. The first is the pooled distribution of control villages. The second caps the benchmark at the control $75$th percentile, $43{,}827$ Kenyan shillings per capita, so that the reference vector does not inherit the most extreme control quantiles. The third is external and normative: the national Kenyan per-capita consumption distribution from the KIHBS 2015/16 survey \citep{knbs_kihbs_2016}, as summarized by the World Bank Poverty and Inequality Platform \citep{worldbank_pip}. This study rebuilds it on the $25$-point grid from the published mean, median, and decile shares under a monotone Lorenz curve constrained to match the published median, and convert it to nominal shillings with the study's own conversion factor of $46.49$ Kenyan shillings per 2017-PPP dollar, which is constant across households in the replication package. The reconstruction reproduces the published mean to $0.01$ percent, the median exactly, and the Gini coefficient to one percent. Because the national distribution is much richer than either arm, this benchmark represents an external target rather than a description of the study population. The reported contrasts are averages over fit seeds of the calibrated marginal estimates for seven declared functionals: the mean, standard deviation, skewness, upper-half mean, reference distance, and the tenth and ninetieth percentiles. Figure~\ref{fig:applied-egger-estimates} reports them with influence-function standard errors, alongside the design-based raw contrasts and their two-sided permutation $p$-values from 10{,}000 permutations; the external benchmark is fitted as a separate variant because the reference vector enters the estimator's calibration.

The calibrated mean contrast is $+958.5$ (influence-function SE $1308.6$), the standard-deviation contrast is $-779.5$ (SE $2660.0$), the skewness contrast is $+0.060$ (SE $0.0787$), and the upper-half mean contrast is $+1060.4$ (SE $2353.7$). The tenth-percentile contrast is $+1125.5$ (SE $418.7$) and the ninetieth-percentile contrast is $+72.5$ (SE $5121.7$). For the reference distance, the pooled control-village benchmark gives $-3057.5$ (SE $2573.6$), the benchmark capped at the control $75$th percentile gives $-730.6$ (SE $2856.6$), and the external national benchmark gives $-5535.6$ (SE $2235.0$, across-seed standard deviation $132.3$). Only the tenth-percentile and external-reference contrasts are large relative to their standard errors. The design-based contrasts, which use the randomized assignment directly (open diamonds in Figure~\ref{fig:applied-egger-estimates}), are smaller for the mean ($+543.5$, permutation $p=0.675$) and indistinguishable from zero for the pooled control reference ($-2572.1$, $p=0.374$), while the lower-tail result ($+913.9$, $p=0.021$) and the external-reference result ($-4687.4$, $p=0.0075$) are the two contrasts whose permutation nulls are rejected at the conventional level; the design-based ninetieth-percentile contrast is $-632.5$ ($p=0.912$). No multiplicity adjustment is applied, so these two results are nominal. Across the figure, the calibrated and design-based contrasts agree in the lower tail and for the external benchmark, and diverge in the extreme upper tail, where neither is precise enough to support a sign.

\begin{figure}[t]
\centering
\includegraphics[width=\linewidth]{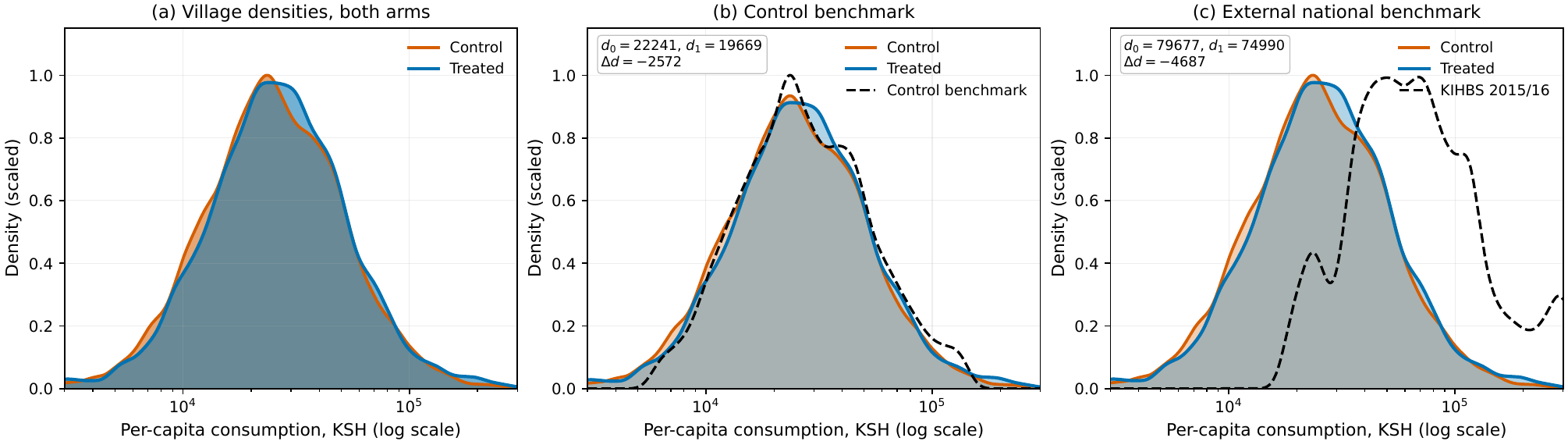}
\caption{Kenya cash-transfer village densities. Each arm curve is a Gaussian kernel density estimate of the equal-unit mixture of the village laws, formed from the pooled village quantile coordinates; benchmark curves apply the same estimator to their quantile vectors interpolated to a fine level grid. The bandwidth is Silverman's rule applied to the pooled arm coordinates and is common to all curves in a panel. Densities are scaled to a common maximum in each panel and are unadjusted for covariates. Panel (a) overlays the two arms, panel (b) fades the control density and adds the pooled control-village benchmark, and panel (c) repeats panel (b) with the external national KIHBS 2015/16 benchmark. The control benchmark overlaps both arms, while the national benchmark lies to the right with a long right tail; the raw reference contrast is negative for both and is larger for the national benchmark.}
\label{fig:applied-egger-densities}
\end{figure}

Figure~\ref{fig:applied-egger-densities} shows the two arm densities together with the two distributional benchmarks on the consumption scale. The pooled control benchmark overlaps the two arms closely, which is why its raw reference contrast is small relative to the spread of the laws, and the lower tail is where the treated density is visibly shifted right. The national benchmark lies to the right of both arms and carries a long right tail, so distances to it are dominated by the upper tail; the treated density is closer to it than the control density, and the permutation test rejects a zero raw contrast at the one percent level. Taken together, the evidence from this design is a lower-tail consumption gain rather than a mean effect. For the pooled control benchmark the reference-distance estimate has the sign of movement toward the control distribution, but its standard error is as large as the estimate and the design-based contrast is indistinguishable from zero, so an effect on the reference distance is not statistically established. With the external national benchmark, by contrast, the reference contrast is negative and nominally distinguishable from zero in both the calibrated and design-based layers, which is consistent with transfers moving village consumption distributions toward the national distribution. The saturation reversal cautions against reading the reference contrast as a single homogeneous quantity, since it depends on the local intensity of treatment, which the pooled estimate does not capture, and the external benchmark is a reconstruction from grouped survey data rather than a direct measurement of the study population.

%% file: references.bib
@techreport{word1990state,
  title={The State of Tennessee’s Student/Teacher Achievement Ratio ({STAR}) Project: Technical Report (1985--1990)},
  author={Word, Elizabeth and Johnston, John and Bain, Helen P and Fulton, B DeWayne and Zaharias, Jayne B and Achilles, Charles M and Lintz, Martha N and Folger, John and Breda, Carolyn},
  institution={Tennessee State Department of Education},
  address={Nashville, TN},
  year={1990},
  url={https://eric.ed.gov/?id=ED328356}
}

@article{souto_diamantis_tcda_2026,
  title   = {A Mathematical Framework for Topological Causal Data Analysis},
  author  = {Souto, Hugo Gobato and Diamantis, Ioannis},
  journal = {arXiv preprint arXiv:2607.28161},
  year    = {2026},
  url     = {https://arxiv.org/abs/2607.28161},
  eprint  = {2607.28161},
  archivePrefix = {arXiv},
  primaryClass = {stat.ME}
}

@inproceedings{naf2026causaldrf,
  author    = {N{\"a}f, Jeffrey and Park, Junhyung and Susmann, Herbert},
  title     = {Causal-{DRF}: Conditional Kernel Treatment Effect Estimation using Distributional Random Forest},
  booktitle = {Proceedings of the 29th International Conference on Artificial Intelligence and Statistics (AISTATS)},
  series    = {Proceedings of Machine Learning Research},
  publisher = {PMLR},
  volume    = {300},
  pages     = {1000--1008},
  year      = {2026},
  url       = {https://proceedings.mlr.press/v300/naf26a.html}
}

@article{DRF-paper,
  author  = {{\'C}evid, Domagoj and Michel, Loris and N{\"a}f, Jeffrey and B{\"u}hlmann, Peter and Meinshausen, Nicolai},
  title   = {Distributional Random Forests: Heterogeneity Adjustment and Multivariate Distributional Regression},
  journal = {Journal of Machine Learning Research},
  year    = {2022},
  volume  = {23},
  number  = {333},
  pages   = {1--79},
  url     = {https://jmlr.org/papers/v23/21-0585.html}
}

@article{naf2023confidence,
  author  = {N{\"a}f, Jeffrey and Emmenegger, Corinne and B{\"u}hlmann, Peter and Meinshausen, Nicolai},
  title   = {Confidence and Uncertainty Assessment for Distributional Random Forests},
  journal = {Journal of Machine Learning Research},
  year    = {2023},
  volume  = {24},
  number  = {366},
  pages   = {1--77},
  url     = {https://jmlr.org/papers/v24/23-0185.html}
}

@article{wager2018estimation,
  title     = {Estimation and inference of heterogeneous treatment effects using random forests},
  author    = {Wager, Stefan and Athey, Susan},
  journal   = {Journal of the American Statistical Association},
  volume    = {113},
  number    = {523},
  pages     = {1228--1242},
  year      = {2018},
  publisher = {Taylor \& Francis},
  doi       = {10.1080/01621459.2017.1319839}
}

@article{athey2019generalized,
  author    = {Athey, Susan and Tibshirani, Julie and Wager, Stefan},
  title     = {Generalized random forests},
  volume    = {47},
  journal   = {The Annals of Statistics},
  number    = {2},
  publisher = {Institute of Mathematical Statistics},
  pages     = {1148--1178},
  year      = {2019},
  doi       = {10.1214/18-AOS1709}
}

@article{chernozhukov2018double,
  author  = {Chernozhukov, Victor and Chetverikov, Denis and Demirer, Mert and Duflo, Esther and Hansen, Christian and Newey, Whitney and Robins, James},
  title   = {Double/debiased machine learning for treatment and structural parameters},
  journal = {The Econometrics Journal},
  volume  = {21},
  number  = {1},
  pages   = {C1--C68},
  year    = {2018},
  doi     = {10.1111/ectj.12097}
}

@article{gretton2012kernel,
  title     = {A kernel two-sample test},
  author    = {Gretton, Arthur and Borgwardt, Karsten M. and Rasch, Malte J. and Sch{\"o}lkopf, Bernhard and Smola, Alexander},
  journal   = {Journal of Machine Learning Research},
  volume    = {13},
  number    = {25},
  pages     = {723--773},
  year      = {2012},
  url       = {https://jmlr.org/papers/v13/gretton12a.html}
}

@inproceedings{CATEGeneralization,
  title     = {Conditional Distributional Treatment Effect with Kernel Conditional Mean Embeddings and {U}-Statistic Regression},
  author    = {Park, Junhyung and Shalit, Uri and Sch{\"o}lkopf, Bernhard and Muandet, Krikamol},
  booktitle = {Proceedings of the 38th International Conference on Machine Learning (ICML)},
  volume    = {139},
  pages     = {8401--8412},
  year      = {2021},
  url       = {https://proceedings.mlr.press/v139/park21c.html}
}

@InProceedings{GRFDistributionalCausalEffects,
  title     = {Robust and Agnostic Learning of Conditional Distributional Treatment Effects},
  author    = {Kallus, Nathan and Oprescu, Miruna},
  booktitle = {Proceedings of The 26th International Conference on Artificial Intelligence and Statistics},
  pages     = {6037--6060},
  year      = {2023},
  volume    = {206},
  url       = {https://proceedings.mlr.press/v206/kallus23a.html}
}

@inproceedings{martineztaboada2023efficient,
  author    = {Martinez Taboada, Diego and Ramdas, Aaditya and Kennedy, Edward},
  booktitle = {Advances in Neural Information Processing Systems},
  pages     = {59924--59952},
  title     = {An Efficient Doubly-Robust Test for the Kernel Treatment Effect},
  volume    = {36},
  year      = {2023},
  doi       = {10.52202/075280-2619},
  url       = {https://papers.nips.cc/paper_files/paper/2023/hash/bccdd196d798a51a4961989984a9ed4a-Abstract-Conference.html}
}

@article{Counterfactualmeanembeddings,
  author  = {Muandet, Krikamol and Kanagawa, Motonobu and Saengkyongam, Sorawit and Marukatat, Sanparith},
  title   = {Counterfactual Mean Embeddings},
  journal = {Journal of Machine Learning Research},
  year    = {2021},
  volume  = {22},
  number  = {162},
  pages   = {1--71},
  url     = {https://www.jmlr.org/papers/v22/20-185.html}
}

@article{jain2026conditional,
  author  = {Jain, Saksham and Luedtke, Alex},
  title   = {Conditional Distributional Treatment Effects: Doubly Robust Estimation and Testing},
  journal = {arXiv preprint arXiv:2603.16829},
  year    = {2026},
  url     = {https://arxiv.org/abs/2603.16829},
  eprint  = {2603.16829},
  archivePrefix = {arXiv},
  primaryClass = {stat.ML}
}

@article{Chernozhukov2013,
  title   = {Inference on counterfactual distributions},
  author  = {Chernozhukov, Victor and Fern{\'a}ndez-Val, Iv{\'a}n and Melly, Blaise},
  journal = {Econometrica},
  volume  = {81},
  number  = {6},
  pages   = {2205--2268},
  year    = {2013},
  doi     = {10.3982/ECTA10582}
}

@article{rosenbaum1983propensity,
  author  = {Rosenbaum, Paul R. and Rubin, Donald B.},
  title   = {The Central Role of the Propensity Score in Observational Studies for Causal Effects},
  journal = {Biometrika},
  volume  = {70},
  number  = {1},
  pages   = {41--55},
  year    = {1983},
  doi     = {10.1093/biomet/70.1.41}
}

@article{bang2005doubly,
  author  = {Bang, Heejung and Robins, James M.},
  title   = {Doubly Robust Estimation in Missing Data and Causal Inference Models},
  journal = {Biometrics},
  volume  = {61},
  number  = {4},
  pages   = {962--973},
  year    = {2005},
  doi     = {10.1111/j.1541-0420.2005.00377.x}
}

@article{gneiting2007proper,
  author  = {Gneiting, Tilmann and Raftery, Adrian E.},
  title   = {Strictly Proper Scoring Rules, Prediction, and Estimation},
  journal = {Journal of the American Statistical Association},
  volume  = {102},
  number  = {477},
  pages   = {359--378},
  year    = {2007},
  doi     = {10.1198/016214506000001437}
}

@article{hoeffding1963probability,
  author  = {Hoeffding, Wassily},
  title   = {Probability Inequalities for Sums of Bounded Random Variables},
  journal = {Journal of the American Statistical Association},
  volume  = {58},
  number  = {301},
  pages   = {13--30},
  year    = {1963},
  doi     = {10.1080/01621459.1963.10500830},
  url     = {https://www.cs.rpi.edu/academics/courses/spring06/random/hoefding.pdf}
}

@article{nadaraya1964regression,
  author  = {Nadaraya, E. A.},
  title   = {On Estimating Regression},
  journal = {Theory of Probability and Its Applications},
  volume  = {9},
  number  = {1},
  pages   = {141--142},
  year    = {1964},
  doi     = {10.1137/1109020},
  url     = {https://www.mathnet.ru/eng/tvp356}
}

@article{busing2022monotone,
  author  = {Busing, Frank M. T. A.},
  title   = {Monotone Regression: A Simple and Fast {O(n)} {PAVA} Implementation},
  journal = {Journal of Statistical Software, Code Snippets},
  volume  = {102},
  number  = {1},
  pages   = {1--25},
  year    = {2022},
  doi     = {10.18637/jss.v102.c01},
  url     = {https://www.jstatsoft.org/article/view/v102c01}
}

@article{peyre2019optimaltransport,
  author  = {Peyr{\'e}, Gabriel and Cuturi, Marco},
  title   = {Computational Optimal Transport},
  journal = {Foundations and Trends in Machine Learning},
  volume  = {11},
  number  = {5--6},
  pages   = {355--607},
  year    = {2019},
  doi     = {10.1561/2200000073},
  url     = {https://www.nowpublishers.com/article/Details/MAL-073}
}

@article{duan1983medical,
  author  = {Duan, Naihua and Manning, Willard G. and Morris, Carl N. and Newhouse, Joseph P.},
  title   = {A Comparison of Alternative Models for the Demand for Medical Care},
  journal = {Journal of Business and Economic Statistics},
  volume  = {1},
  number  = {2},
  pages   = {115--126},
  year    = {1983},
  doi     = {10.1080/07350015.1983.10509330}
}

@article{mullahy1986modified,
  author  = {Mullahy, John},
  title   = {Specification and Testing of Some Modified Count Data Models},
  journal = {Journal of Econometrics},
  volume  = {33},
  number  = {3},
  pages   = {341--365},
  year    = {1986},
  doi     = {10.1016/0304-4076(86)90002-3}
}

@article{hahn2020bayesian,
  title   = {Bayesian regression tree models for causal inference: Regularization, confounding, and heterogeneous effects},
  author  = {Hahn, P. Richard and Murray, Jared S. and Carvalho, Carlos M.},
  journal = {Bayesian Analysis},
  volume  = {15},
  number  = {3},
  pages   = {965--1056},
  year    = {2020},
  doi     = {10.1214/19-BA1195}
}

@article{krueger1999experimental,
  title   = {Experimental Estimates of Education Production Functions},
  author  = {Krueger, Alan B.},
  journal = {The Quarterly Journal of Economics},
  volume  = {114},
  number  = {2},
  pages   = {497--532},
  year    = {1999},
  doi     = {10.1162/003355399556052},
  url     = {https://doi.org/10.1162/003355399556052}
}

@article{chetty2011kindergarten,
  title   = {How Does Your Kindergarten Classroom Affect Your Earnings? Evidence from Project {STAR}},
  author  = {Chetty, Raj and Friedman, John N. and Hilger, Nathaniel and Saez, Emmanuel and Schanzenbach, Diane Whitmore and Yagan, Danny},
  journal = {The Quarterly Journal of Economics},
  volume  = {126},
  number  = {4},
  pages   = {1593--1660},
  year    = {2011},
  doi     = {10.1093/qje/qjr041},
  url     = {https://doi.org/10.1093/qje/qjr041}
}

@article{egger2022general,
  author  = {Egger, Dennis and Haushofer, Johannes and Miguel, Edward and Niehaus, Paul and Walker, Michael},
  title   = {General Equilibrium Effects of Cash Transfers: Experimental Evidence from Kenya},
  journal = {Econometrica},
  volume  = {90},
  number  = {6},
  pages   = {2603--2643},
  year    = {2022},
  doi     = {10.3982/ECTA17945},
  url     = {https://doi.org/10.3982/ECTA17945}
}

@misc{egger2024replication,
  author       = {Egger, Dennis and Haushofer, Johannes and Miguel, Edward and Niehaus, Paul and Walker, Michael},
  title        = {Replication Package for: ``General Equilibrium Effects of Cash Transfers: Experimental Evidence from Kenya''},
  year         = {2026},
  howpublished = {Zenodo, v1},
  doi          = {10.5281/zenodo.16548593},
  url          = {https://doi.org/10.5281/zenodo.16548593}
}

@article{cengiz2019effect,
  author  = {Cengiz, Doruk and Dube, Arindrajit and Lindner, Attila and Zipperer, Ben},
  title   = {The Effect of Minimum Wages on Low-Wage Jobs},
  journal = {The Quarterly Journal of Economics},
  volume  = {134},
  number  = {3},
  pages   = {1405--1454},
  year    = {2019},
  doi     = {10.1093/qje/qjz014},
  url     = {https://doi.org/10.1093/qje/qjz014}
}

@article{lee1999wage,
  author  = {Lee, David S.},
  title   = {Wage Inequality in the United States during the 1980s: Rising Dispersion or Falling Minimum Wage?},
  journal = {The Quarterly Journal of Economics},
  volume  = {114},
  number  = {3},
  pages   = {977--1023},
  year    = {1999},
  doi     = {10.1162/003355399556197},
  url     = {https://doi.org/10.1162/003355399556197}
}

@techreport{vaghul2016historical,
  author      = {Vaghul, Kavya and Zipperer, Ben},
  title       = {Historical State and Sub-State Minimum Wage Data},
  institution = {Washington Center for Equitable Growth},
  type        = {Working Paper},
  month       = {September},
  year        = {2016},
  url         = {https://equitablegrowth.org/working-papers/historical-state-and-sub-state-minimum-wage-data}
}

@misc{nbermorg,
  author       = {{National Bureau of Economic Research}},
  title        = {Current Population Survey ({CPS}) Merged Outgoing Rotation Group Earnings Data},
  year         = {2025},
  howpublished = {NBER Public Use Data Archive},
  url          = {https://www.nber.org/research/data/current-population-survey-cps-merged-outgoing-rotation-group-earnings-data}
}

@misc{eurostat2026ilc,
  author       = {{Eurostat}},
  title        = {Distribution of Income by Quantiles},
  year         = {2026},
  howpublished = {Eurostat online data code ilc\_di01},
  url          = {https://ec.europa.eu/eurostat/databrowser/view/ilc_di01/default/table}
}

@misc{worldbank_pip,
  author       = {{World Bank}},
  title        = {Poverty and Inequality Platform ({PIP})},
  year         = {2026},
  howpublished = {World Bank, Washington, DC},
  url          = {https://pip.worldbank.org}
}

@misc{knbs_kihbs_2016,
  author       = {{Kenya National Bureau of Statistics}},
  title        = {Kenya Integrated Household Budget Survey 2015/16},
  year         = {2018},
  howpublished = {KNBS, Nairobi},
  url          = {https://www.knbs.or.ke/reports/kenya-integrated-household-budget-survey-2015-2016/}
}
